\documentclass[11pt]{article}

\usepackage[margin=1in]{geometry}
\usepackage{amsmath,amssymb,amsfonts}
\usepackage{graphicx}
\usepackage{booktabs}
\usepackage{pifont}
\usepackage{hyperref}
\usepackage{cleveref}
\usepackage{natbib}
\usepackage{xcolor}
\usepackage{algorithm}
\usepackage{algorithmic}
\usepackage{caption}
\usepackage{subcaption}
\usepackage{microtype}

\newcommand{\R}{\mathbb{R}}
\newcommand{\norm}[1]{\left\|#1\right\|}

\newcommand{\defect}{\mathcal{D}}

\title{Low-Dimensional and Transversely Curved Optimization Dynamics in Grokking}

\author{%
  Yongzhong Xu
  \thanks{abbyxu@gmail.com ; code at https://github.com/skydancerosel/grokking-integrability}
}

\date{}

\begin{document}
\maketitle

\begin{abstract}

Grokking---the abrupt onset of generalization long after memorization---challenges standard accounts of how neural networks learn, yet the geometric mechanisms driving this transition remain poorly understood.
We propose that grokking corresponds to prolonged confinement on a low-dimensional subspace in weight space, during which transverse curvature barriers accumulate until the trajectory escapes into the generalizing solution.
We identify this subspace---the \emph{execution manifold}---in the weight space of transformers trained on modular arithmetic.
Using PCA on attention weight trajectories and commutator defect analysis across six binary operations mod~97, we show that weight evolution during grokking is essentially one-dimensional: a single principal component captures 68--83\% of variance across 36 experimental conditions.
We then measure loss-landscape curvature via commutator defects---the non-commutativity of successive gradient steps---and project these onto the learned submanifold.
The commutator vectors are predominantly orthogonal to the execution manifold, even relative to random baselines (exec/random ratio $\approx 2$--$3\times$), indicating that the subspace is \emph{empirically invariant} under the optimization dynamics: curvature does not deflect the trajectory out of its learned subspace.
Yet curvature explodes orthogonally: grokking operations show $10$--$1000\times$ higher commutator defect than non-grokking controls, concentrated entirely in the normal bundle of the execution manifold.
The onset of this curvature growth consistently \emph{precedes} the generalization transition by 600--1600 training steps at $\eta = 10^{-3}$ (sign test $p = 2^{-12} < 0.001$), though non-grokking operations also exhibit moderate curvature growth ($30$--$50\times$) without generalizing, so onset is necessary but not sufficient.
Across a $300\times$ learning-rate sweep, the lead time obeys a power law $\Delta t \propto t_{\mathrm{grok}}^{\,\alpha}$ with $\alpha = 1.27 \pm 0.03$ ($R^2 = 0.97$, $n = 43$): because $\alpha > 1$, the predictive window grows super-linearly with grokking timescale---at $\eta = 3\!\times\!10^{-5}$, defect onset occurs within the first 5\% of training, providing a 95\% advance warning window.
All findings replicate across this learning-rate range, a qualitatively different slow regime ($\mathrm{lr}=5\!\times\!10^{-5}$, $\mathrm{wd}=0.1$, 3 layers), and three random seeds, though alignment dynamics differ quantitatively between regimes.
Causal intervention experiments establish that orthogonal gradient flow is necessary but not sufficient for grokking: suppressing it prevents generalization with a monotonic dose--response across four operations, while artificially boosting curvature defects has no effect.
Weight-SVD spectral analysis reveals the mechanism: grokking is preceded by a near-degeneracy of the leading singular values of the attention matrices ($\sigma_1 \approx \sigma_2$), creating an orientation instability that drives the commutator transition.
Generalization coincides with the breaking of this spectral symmetry as one mode dominates.
This spectral structure is confirmed to be basis-independent: three independent decompositions (weight SVD, displacement SVD, gradient SVD) all exhibit a consistent sign flip in the commutator--subspace alignment at grokking.

\end{abstract}

\section{Introduction}
\label{sec:intro}

Grokking---the phenomenon where neural networks trained on small algorithmic datasets first memorize the training set and then, long after achieving perfect training accuracy, suddenly generalize to the test set---was first reported by \citet{power2022grokking} in modular arithmetic tasks.
The phenomenon has attracted significant attention because it challenges the conventional understanding that generalization and memorization are tightly coupled in optimization dynamics.

Prior work has characterized grokking through the lens of representation learning \citep{nanda2023grokking}, weight decay as implicit regularization \citep{liu2022omnigrok}, circuit formation \citep{zhong2024clock}, and phase transitions in loss landscapes.
Recently, \citet{xu2026lowdim} showed that the weight-space trajectory during grokking lies on a low-dimensional execution manifold, with PCA revealing that a single principal component captures the majority of trajectory variance.
However, a fundamental geometric question remains open: \emph{is this low-dimensional manifold invariant under the optimization dynamics---does curvature deflect the trajectory out of its learned subspace---and does its geometry predict when generalization will occur?}

Building on \citet{xu2026lowdim}, we address this question by studying the differential geometry of the parameter-space trajectory during grokking.
Our approach combines PCA eigenanalysis with commutator defect analysis---probing the curvature structure of the loss landscape relative to the learned submanifold:
\begin{enumerate}
    \item \textbf{PCA eigenanalysis} of attention weight trajectories, revealing the intrinsic dimensionality of the learned submanifold;
    \item \textbf{Commutator defect analysis}, measuring loss-landscape curvature and its relationship to the learned submanifold.
\end{enumerate}

The commutator defect quantifies the non-commutativity of successive gradient steps: given two mini-batches $A$ and $B$, the defect measures how much the final parameter vector depends on the order of gradient updates.
In a flat region of the loss landscape, gradient steps commute; in a curved region, they do not.
By projecting these commutator vectors onto the PCA submanifold, we can determine whether the learned subspace is flat or curved.

\paragraph{Key contributions.}
Our work makes six main contributions, spanning observation, prediction, and causal testing:
\begin{enumerate}
    \item \textbf{Rank-1 manifold}: The weight-space trajectory during grokking lies on a rank-1 submanifold (68--83\% of variance in PC1).
    \item \textbf{Invariant submanifold}: This submanifold exhibits strong empirical invariance under the optimization dynamics: commutator defects are predominantly orthogonal to it ($\rho \approx 1.000$ within numerical precision across 36 conditions, with exec/random projection ratio $\approx 2$--$3\times$), meaning curvature does not deflect the trajectory out of its learned subspace.
    \item \textbf{Temporal ordering}: Curvature explodes orthogonally during grokking ($10$--$1000\times$ increase) and its onset consistently \emph{precedes} generalization by 600--1600 steps. Non-grokking operations show moderate curvature growth ($30$--$50\times$) without generalization, so onset is necessary but not sufficient.
    \item \textbf{Scaling law}: The lead time between defect onset and generalization obeys a power law $\Delta t \propto t_{\mathrm{grok}}^{\,\alpha}$ with $\alpha = 1.27 \pm 0.03$ ($R^2 = 0.97$) across a $300\times$ learning-rate sweep: the predictive window improves from 24\% at $\eta = 3\!\times\!10^{-3}$ to 95\% at $\eta = 3\!\times\!10^{-5}$.
    \item \textbf{Causal interventions}: Suppressing orthogonal gradient flow prevents grokking (necessary) while boosting curvature defects has no effect (not sufficient), establishing an asymmetric causal relationship.
    \item \textbf{Robustness}: All results replicate across the full learning-rate range, a $200\times$ timescale difference between regimes, four operations, and three seeds.
    \item \textbf{Spectral mechanism}: Weight-SVD analysis reveals that the commutator transition is driven by a transient near-degeneracy of the leading singular values ($\sigma_1 \approx \sigma_2$) of the attention matrices, followed by symmetry breaking as one mode dominates and the operators align.
\end{enumerate}

\noindent
Our central thesis is that grokking reflects prolonged confinement of optimization dynamics within a low-dimensional subspace of weight space, during which curvature accumulates in transverse directions; generalization emerges as the trajectory exits this metastable regime.
The remainder of the paper develops this picture quantitatively and probes its causal structure.

\paragraph{Paper outline.}
We proceed as follows.
\Cref{sec:setup} describes the experimental setup.
\Cref{sec:methods} describes the geometric tools: PCA eigenanalysis, commutator defect, manifold projection, and trajectory--curvature alignment.
\Cref{sec:results} presents results in three stages: geometric structure (\Cref{sec:rank1}--\Cref{sec:curvature}), predictive power and robustness (\Cref{sec:prediction}--\Cref{sec:lr_sweep}), and causal interventions (\Cref{sec:interventions}).
\Cref{sec:spectral} identifies the spectral mechanism underlying the commutator transition: a symmetry-breaking event in the singular value spectrum of the attention matrices.
\Cref{sec:discussion} discusses implications and connections to broader themes.

\section{Experimental Setup}
\label{sec:setup}

\subsection{Model and Training}

We use a Transformer encoder following the canonical grokking setup of \citet{power2022grokking}.
The model processes two integer tokens $a, b \in \{0, \ldots, p-1\}$ (with $p = 97$) and predicts $f(a,b) \bmod p$ for a binary operation $f$.

\paragraph{Architecture.}
The model consists of:
\begin{itemize}
    \item A token embedding $\mathrm{Emb}: \{0,\ldots,96\} \to \R^{128}$ plus a learnable positional embedding $\mathbf{P} \in \R^{2 \times 128}$;
    \item A 2-layer Transformer encoder with pre-norm (LayerNorm before attention and FFN), $d_\text{model} = 128$, 4 attention heads, $d_\text{ff} = 256$, GELU activation, no dropout;
    \item A final LayerNorm followed by a linear head $\R^{128} \to \R^{97}$ applied to the first token position.
\end{itemize}
The total parameter count is approximately 290k.

\paragraph{Training.}
We train with AdamW ($\beta_1 = 0.9$, $\beta_2 = 0.98$) at learning rate $10^{-3}$ with weight decay $\lambda = 1.0$ (or $\lambda = 0.0$ for non-grokking controls), batch size 512, gradient clipping at 1.0, and a 50/50 train/test split.
Training runs for up to 200k steps with early stopping when test accuracy exceeds 98\% for 3 consecutive evaluations.
We say a model has \textbf{grokked} if test accuracy reaches 98\% and remains there for 3 consecutive evaluations (early stopping).
For runs that have grokked, we define the \textbf{grok step} $t_\text{grok}$ as the first step at which test accuracy reaches 90\%---a lower threshold that captures the onset of generalization rather than its plateau, and is used throughout for lead-time analysis.
(No non-grokking operation exceeds 78\% test accuracy in any condition, so this threshold does not conflate the two groups.)

\paragraph{Operations.}
We test six binary operations mod~97, four of which exhibit grokking under these hyperparameters and two that do not (\Cref{tab:operations}).

\begin{table}[ht]
\centering
\caption{Operations tested. Grok step is the mean step at which test accuracy reaches 90\%, averaged over 3 seeds.}
\label{tab:operations}
\begin{tabular}{@{}llcc@{}}
\toprule
Operation & Formula & Groks? & Grok step \\
\midrule
add & $(a+b) \bmod 97$ & Yes & $\sim$2900 \\
sub & $(a-b) \bmod 97$ & Yes & $\sim$3400 \\
mul & $(a \times b) \bmod 97$ & Yes & $\sim$2600 \\
x2\_y2 & $(a^2+b^2) \bmod 97$ & Yes & $\sim$1900 \\
x2\_xy\_y2 & $(a^2+ab+b^2) \bmod 97$ & No & --- \\
x3\_xy & $(a^3+ab) \bmod 97$ & No & --- \\
\bottomrule
\end{tabular}
\end{table}

\subsection{Hyperparameter Regimes}

To test regime invariance, we additionally run a \textbf{slow regime} with qualitatively different hyperparameters: $\mathrm{lr} = 5 \times 10^{-5}$, $\lambda = 0.1$, 3 Transformer layers, and $\beta_2 = 0.999$.
In this regime, grokking occurs at $\sim$570k steps (vs.\ $\sim$3k in the fast regime), providing a $200\times$ difference in training timescale.

\subsection{Attention Weight Logging}

During training, we log the four attention weight matrices---$W_Q$, $W_K$, $W_V$ (extracted from the fused \texttt{in\_proj\_weight}) and $W_O$ (\texttt{out\_proj.weight})---every 100 steps.
Each matrix is $128 \times 128$ (or $128 \times 32$ per head), giving a trajectory of snapshots for subsequent PCA analysis.

\section{Methods}
\label{sec:methods}

\subsection{PCA Eigenanalysis of Weight Trajectories}
\label{sec:pca}

For each attention weight matrix $W \in \R^{d \times d}$, we collect $T$ training snapshots $\{W_t\}_{t=1}^T$ and compute PCA on the flattened trajectory of weight \emph{changes} from initialization:
\begin{equation}
    X = \begin{bmatrix} \mathrm{vec}(W_1 - W_0) \\ \vdots \\ \mathrm{vec}(W_T - W_0) \end{bmatrix} \in \R^{T \times d^2},
\end{equation}
after centering columns.
We compute the SVD $X = U \Sigma V^\top$ and define the explained variance ratio of the $k$-th principal component as $\sigma_k^2 / \sum_i \sigma_i^2$.
The quantity PC1\% $= 100 \times \sigma_1^2 / \sum_i \sigma_i^2$ measures the fraction of trajectory variance captured by a single direction.

\paragraph{Execution manifold.}
We define the \textbf{execution subspace} (henceforth, \emph{execution manifold}) as the low-dimensional subspace spanned by the top-$K$ PCA directions of the weight trajectory.
Concretely, let $V_K = [v_1, \ldots, v_K]$ denote the top-$K$ right singular vectors of $X$.
The execution manifold is the affine subspace $\mathcal{M} = \{W_0 + V_K \alpha : \alpha \in \R^K\}$ through the initial weights $W_0$.
When PC1\% is high ($>70\%$), this manifold is effectively rank-1: the weight trajectory is confined to a one-dimensional curve in parameter space.

\subsection{Commutator Defect}
\label{sec:commutator}

\emph{Intuition.}
If the loss landscape is locally flat, the order in which we apply two gradient updates does not matter: updating with batch $A$ then $B$ gives the same result as $B$ then $A$.
In a curved region, the order matters---just as walking east then north on a sphere leads to a different point than north then east.
The commutator defect quantifies this order-dependence, providing a local probe of loss-landscape curvature that requires no Hessian computation.

\emph{Formal construction.}
The commutator defect measures loss-landscape curvature by quantifying the non-commutativity of gradient steps from two independent mini-batches.
Given the current parameters $\theta_0$ and two mini-batches $A, B$:
\begin{align}
    \theta_{AB} &= \theta_0 - \eta\, g_A(\theta_0) - \eta\, g_B(\theta_0 - \eta\, g_A(\theta_0)) \\
    \theta_{BA} &= \theta_0 - \eta\, g_B(\theta_0) - \eta\, g_A(\theta_0 - \eta\, g_B(\theta_0))
\end{align}
where $g_A(\theta) = \nabla_\theta \mathcal{L}_A(\theta)$ is the gradient of the cross-entropy loss on mini-batch $A$ at parameters $\theta$, and $\eta = 10^{-3}$ is a fixed step size.
The (scale-normalized) commutator defect is:
\begin{equation}
\label{eq:defect}
    \defect = \frac{\norm{\theta_{AB} - \theta_{BA}}}{\norm{\eta\, g_A} \cdot \norm{\eta\, g_B}}.
\end{equation}
We justify this via first-order Taylor expansion: to leading order in $\eta$, $\theta_{AB} - \theta_{BA} \approx \eta^2 (\nabla g_B \cdot g_A - \nabla g_A \cdot g_B)$, which is the Lie bracket of the gradient vector fields.
$\defect$ is thus proportional to the Lie bracket of stochastic gradient vector fields, and serves as a proxy for local nonlinearity of the loss landscape: if the landscape is locally flat, gradient steps commute and $\defect = 0$.

We compute $K = 9$ independent samples of $\defect$ at each measurement point and report the median, providing a robust estimate.

\subsection{Projection onto the PCA Manifold}
\label{sec:projection}

To determine whether loss-landscape curvature lives inside or outside the learned submanifold, we construct an orthonormal basis $B \in \R^{P \times K}$ for the PCA subspace embedded in full parameter space ($P \approx 290$k).

For each Transformer layer and each attention weight matrix $\{W_Q, W_K, W_V, W_O\}$:
\begin{enumerate}
    \item Compute the top-2 PCA directions from the weight trajectory (each a vector in $\R^{d^2}$);
    \item Embed each direction into the full parameter space at the correct offset;
    \item Stack all embedded directions and orthonormalize via QR decomposition.
\end{enumerate}

Given a commutator vector $\delta = \theta_{AB} - \theta_{BA}$, we decompose it as:
\begin{equation}
    \delta = \underbrace{B\, B^\top \delta}_{\delta_\parallel\;\text{(projected)}} + \underbrace{(\delta - B\, B^\top \delta)}_{\delta_\perp\;\text{(residual)}}.
\end{equation}
The \textbf{invariance measure} is the residual fraction:
\begin{equation}
\label{eq:invariance}
    \rho = \frac{\norm{\delta_\perp}}{\norm{\delta}}.
\end{equation}
If $\rho \approx 1$, the commutator is orthogonal to the PCA subspace, indicating that loss-landscape curvature is confined to the normal bundle and the execution manifold is empirically approximately invariant under the observed optimization dynamics---curvature does not deflect the trajectory out of its learned subspace.
If $\rho \approx 0$, curvature lies within the learned subspace and the submanifold is not invariant.

\paragraph{Transverse decoupling.}
We say the optimization dynamics are \textbf{transversely decoupled} on the execution manifold $\mathcal{M}$ if commutator defect vectors are confined to its normal bundle ($\rho \approx 1$).
In this regime, curvature-induced perturbations act almost entirely orthogonally to $\mathcal{M}$, so that the observed trajectory remains confined to the learned subspace over training.
This notion is purely empirical and should not be confused with integrability in the Hamiltonian sense.

Throughout, we use ``invariant'' in an empirical sense: invariance is assessed via near-orthogonality of commutator defects to the learned PCA subspace over finite training windows, rather than in the strict dynamical-systems sense of exact invariance under a continuous flow.
Although our analysis uses differential-geometric language, all quantities are computed for discrete-time stochastic optimization; geometric terms are used descriptively rather than in a strict continuous-time sense.

\subsection{Random Subspace Control}
\label{sec:random_control}

To verify that the near-zero projection fraction $1 - \rho$ reflects genuine geometric structure rather than a trivial dimensionality artifact---any $K$-dimensional subspace of $\R^P$ captures $\sim\!\sqrt{K/P}$ of a random vector---we compare the PCA-basis projection against a random baseline.
For each commutator vector $\delta$, we compute the projection fraction onto $N_\text{rand} = 5$ random $K$-dimensional orthonormal bases (generated via QR decomposition of Gaussian random matrices) and average the results.
The ratio $\text{proj}_{\text{exec}} / \text{proj}_{\text{rand}}$ is the key diagnostic: values significantly above $1.0$ confirm that the PCA subspace captures more commutator energy than expected by chance.
Because absolute projection magnitudes vanish in high-dimensional spaces ($\text{proj}/\text{full} \sim \sqrt{K/P} \ll 1$ for any $K$-dimensional subspace of $\R^P$), only normalized comparisons to random subspaces are geometrically meaningful; accordingly, we report $\rho$ alongside the exec/random ratio throughout.

\subsection{Converse Analysis: Trajectory Alignment with Curvature}
\label{sec:converse}

As a converse test, we ask whether the weight trajectory \emph{avoids} high-curvature directions.
At each checkpoint, we compute the mean absolute cosine similarity between the trajectory step $\Delta\theta_t = \theta_t - \theta_{t-1}$ and $K = 12$ commutator vectors $\{\delta_k\}$:
\begin{equation}
    \bar{c}_t = \frac{1}{K} \sum_{k=1}^K \frac{|\Delta\theta_t \cdot \delta_k|}{\norm{\Delta\theta_t}\,\norm{\delta_k}}.
\end{equation}
For comparison, the expected absolute cosine between random vectors in $\R^P$ is $\sqrt{2/(\pi P)} \approx 1.5 \times 10^{-3}$ for $P = 290$k.
If $\bar{c}_t$ is near this baseline, the trajectory is not aligned with curvature directions.

\section{Results}
\label{sec:results}

We present our findings in three stages.
First, we establish the geometric structure of grokking: rank-1 weight trajectories on an invariant submanifold with orthogonal curvature explosion (\Cref{sec:rank1}--\Cref{sec:curvature}).
Second, we characterize the temporal relationship between curvature growth and generalization, and demonstrate robustness across regimes and learning rates (\Cref{sec:prediction}--\Cref{sec:lr_sweep}).
Third, we test causality through targeted interventions (\Cref{sec:interventions}).

\subsection{Weight Evolution is Rank-1}
\label{sec:rank1}

\begin{figure}[t]
    \centering
    \begin{subfigure}[t]{0.48\textwidth}
        \centering
        \includegraphics[width=\textwidth]{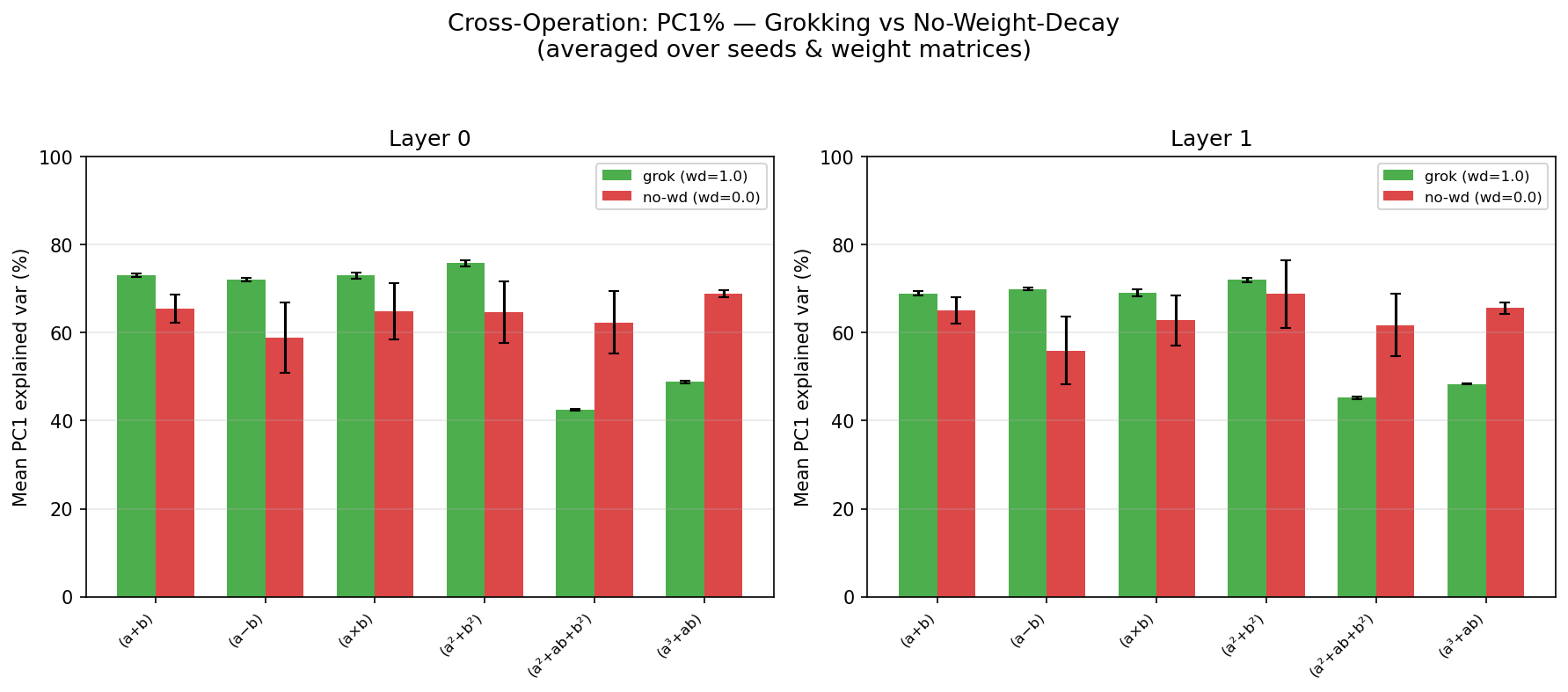}
        \caption{PC1\% for grokking (wd=1.0) vs.\ no-wd (wd=0.0) across operations. Grokking operations consistently show high PC1\%.}
        \label{fig:pca_grok_vs_nowd}
    \end{subfigure}
    \hfill
    \begin{subfigure}[t]{0.48\textwidth}
        \centering
        \includegraphics[width=\textwidth]{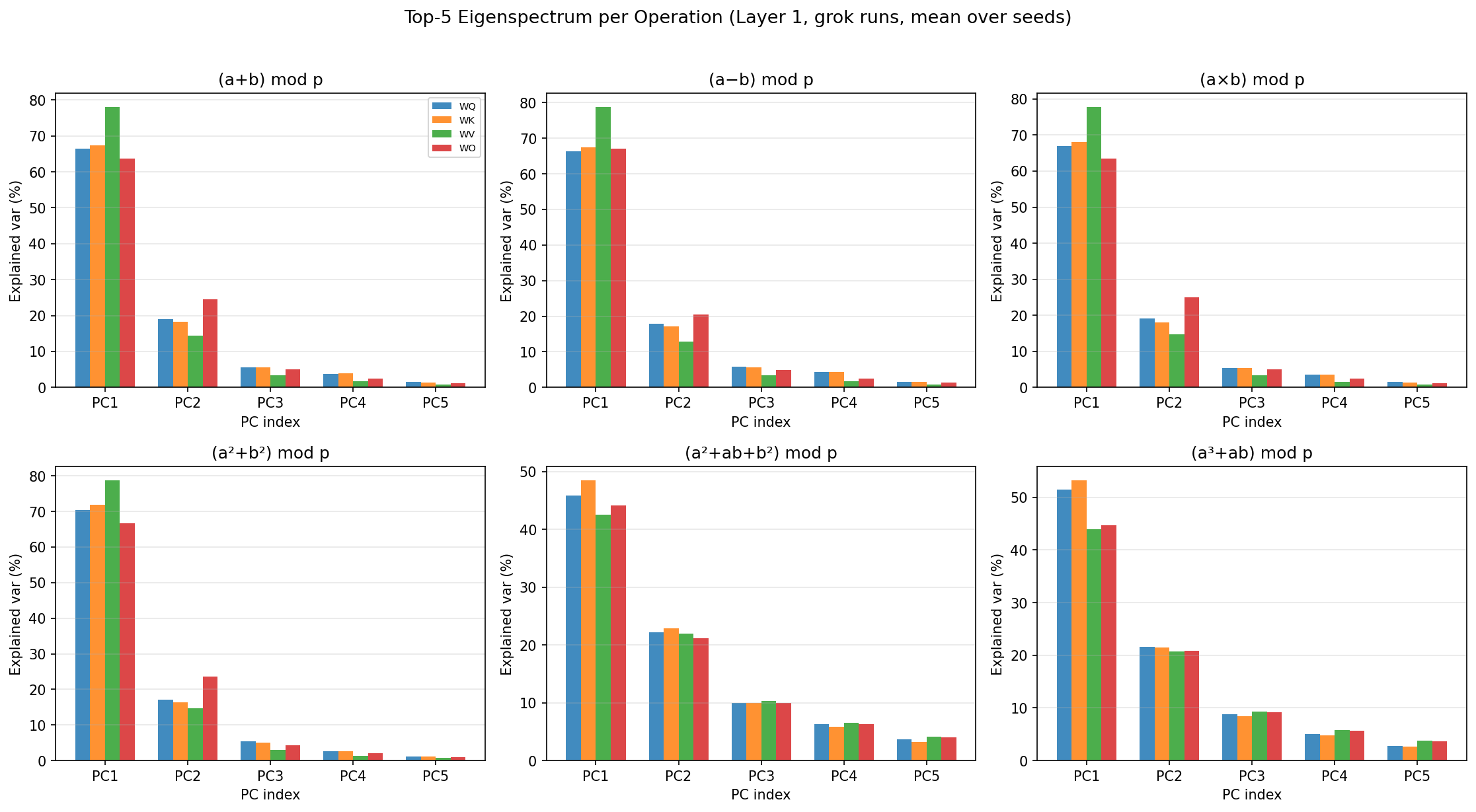}
        \caption{Top-5 eigenspectrum per operation. The first eigenvalue dominates across all operations.}
        \label{fig:eigenspectrum}
    \end{subfigure}
    \caption{Weight trajectories during grokking are rank-1. \textbf{(a)} PC1\% across operations: grokking runs (wd=1.0) show 68--83\% variance in a single component. \textbf{(b)} Eigenspectrum showing dominant first eigenvalue.}
    \label{fig:pca_overview}
\end{figure}

PCA on attention weight trajectories reveals that the first principal component captures 68--83\% of trajectory variance across all grokking conditions (\Cref{fig:pca_overview}).
Weight evolution during grokking is essentially one-dimensional.

\begin{figure}[t]
    \centering
    \begin{subfigure}[t]{0.48\textwidth}
        \centering
        \includegraphics[width=\textwidth]{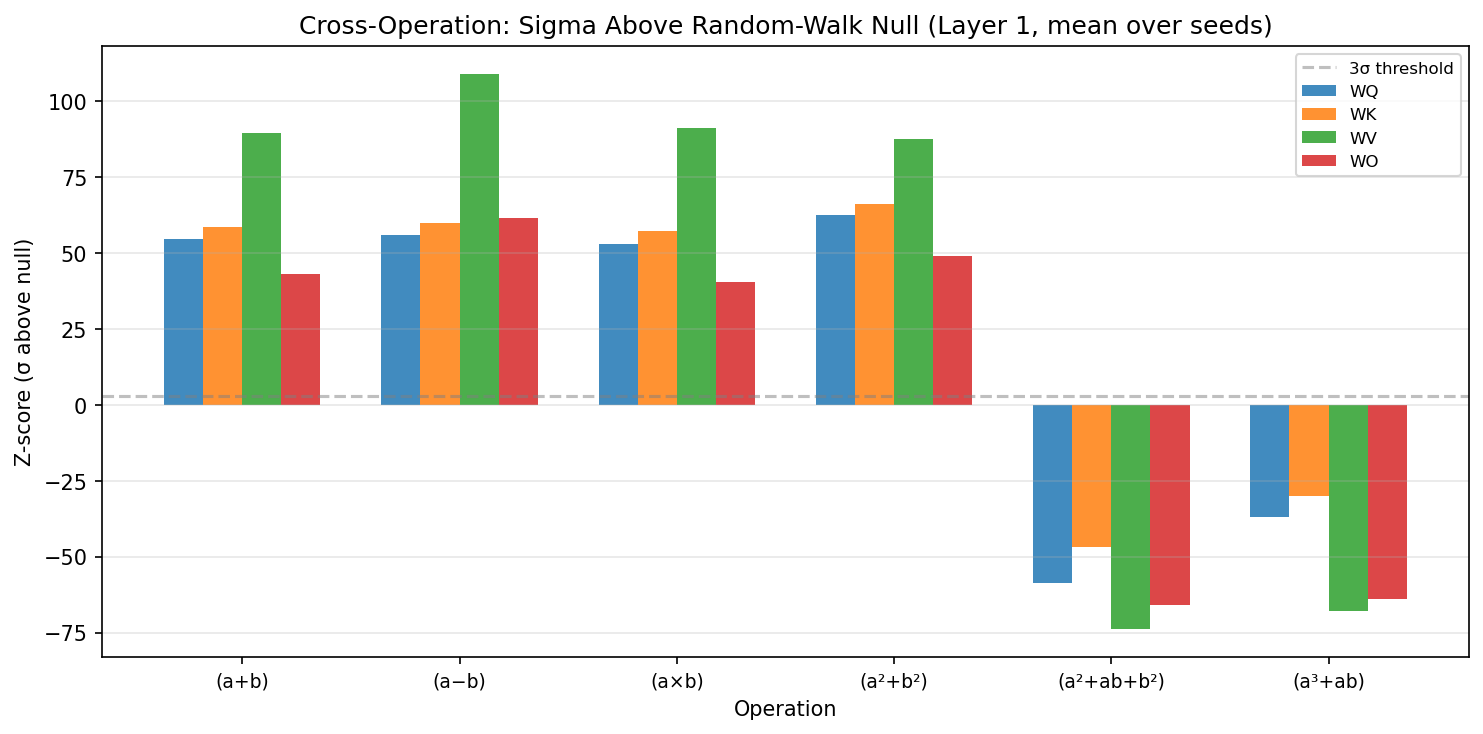}
        \caption{Z-scores vs.\ random-walk null model. All operations exceed the null by $>5\sigma$.}
        \label{fig:null_model}
    \end{subfigure}
    \hfill
    \begin{subfigure}[t]{0.48\textwidth}
        \centering
        \includegraphics[width=\textwidth]{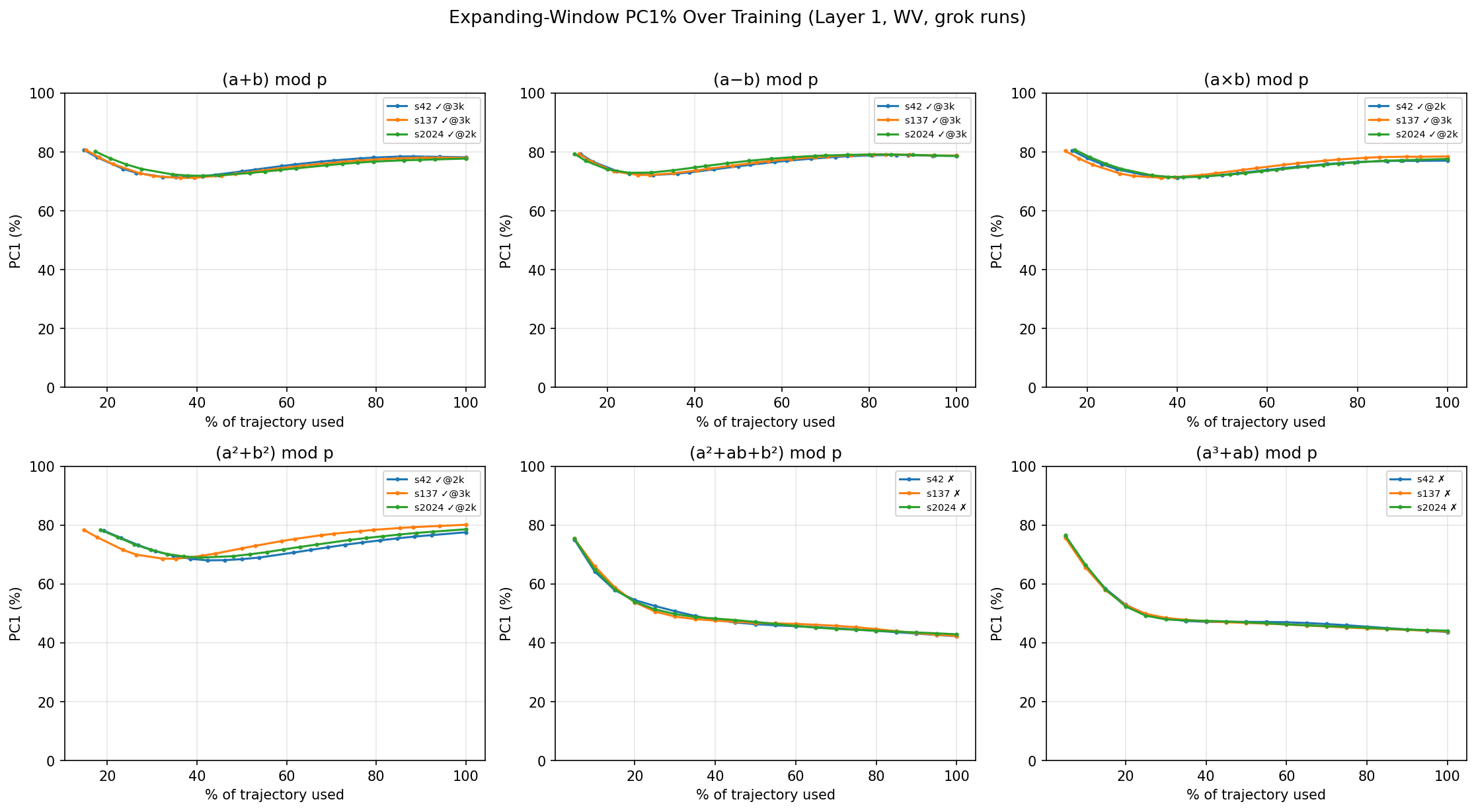}
        \caption{Temporal evolution of PC1\% during training. Concentration increases as grokking progresses.}
        \label{fig:temporal_pca}
    \end{subfigure}
    \caption{PCA concentration is genuine and increases over training. \textbf{(a)} Z-scores above random-walk null. \textbf{(b)} Expanding-window PC1\% over training.}
    \label{fig:pca_controls}
\end{figure}

No-weight-decay controls ($\lambda = 0$) also show moderately high PC1\%, but the null model comparison reveals that grokking PC1\% values are 5--20 standard deviations above the random-walk baseline (\Cref{fig:null_model}), confirming the concentration is not an artifact of trajectory smoothness.

\subsection{The Execution Manifold Exhibits Empirical Invariance}
\label{sec:integrability}

Having established that the weight trajectory lies on a low-dimensional submanifold, we ask: does this submanifold exhibit invariance under the optimization dynamics---does loss-landscape curvature deflect the trajectory out of its learned subspace, or is curvature confined to orthogonal directions?

\begin{figure}[t]
    \centering
    \begin{subfigure}[t]{0.48\textwidth}
        \centering
        \includegraphics[width=\textwidth]{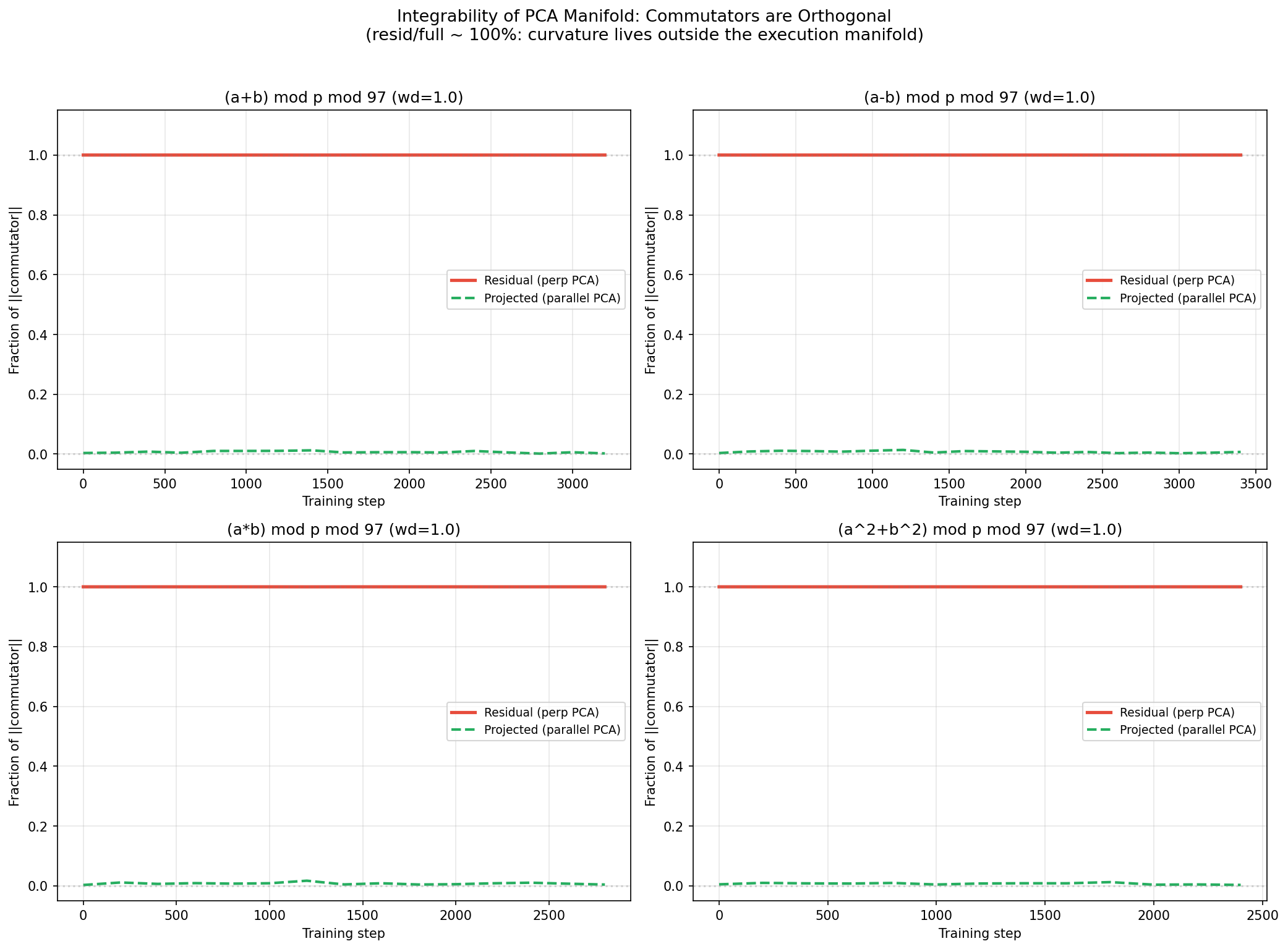}
        \caption{Invariance measure: the residual fraction $\rho \approx 1.0$ at every checkpoint, meaning commutator vectors are predominantly orthogonal to the execution manifold.}
        \label{fig:integrability_single}
    \end{subfigure}
    \hfill
    \begin{subfigure}[t]{0.48\textwidth}
        \centering
        \includegraphics[width=\textwidth]{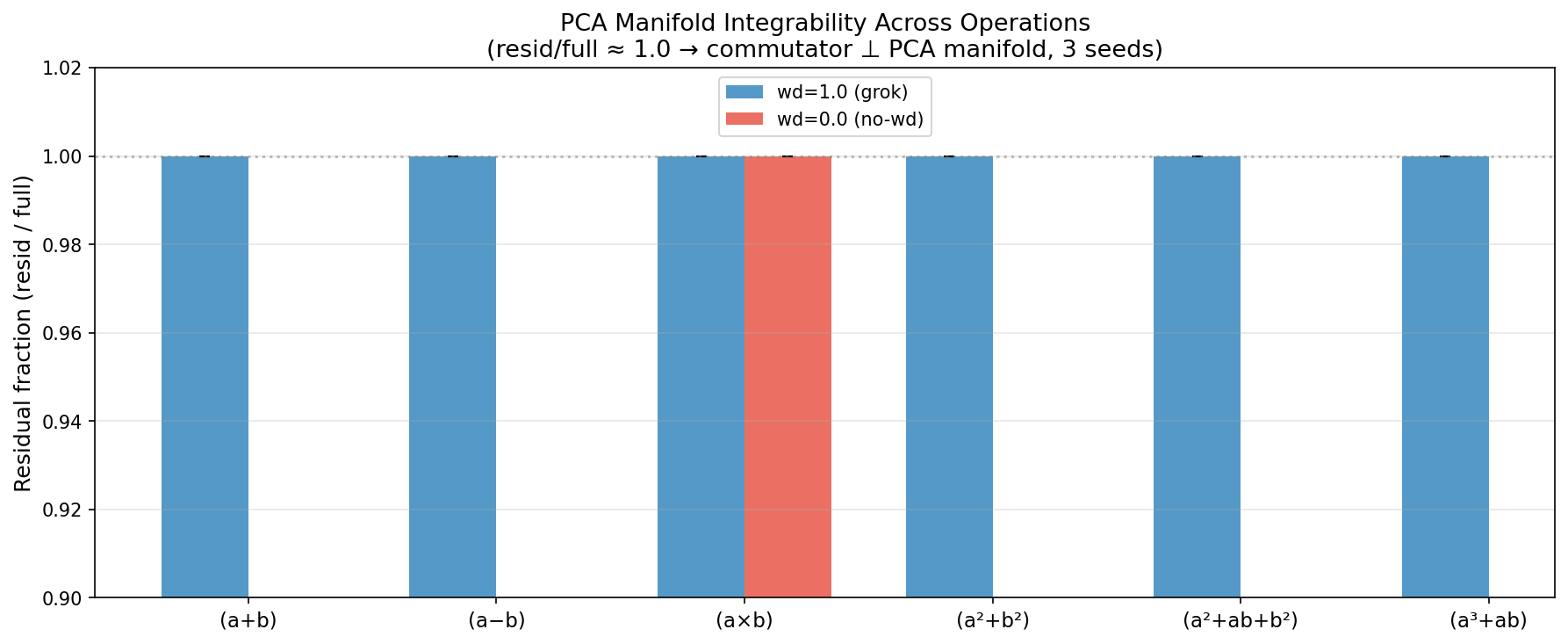}
        \caption{Multi-seed replication: $\rho \approx 1.000$ within numerical precision across all 36 conditions, confirming empirical invariance.}
        \label{fig:integrability_multiseed}
    \end{subfigure}
    \caption{The execution manifold exhibits empirical invariance under the optimization dynamics. Commutator defect vectors are predominantly orthogonal to the PCA subspace, with curvature confined to the normal bundle.}
    \label{fig:integrability}
\end{figure}

We compute commutator defects at regular checkpoints during training and project each commutator vector onto the PCA basis (\Cref{sec:projection}).
The key result: the residual fraction $\rho = \norm{\delta_\perp}/\norm{\delta}$ is $\approx 1.000$ within numerical precision across all 36 conditions (6 operations $\times$ 2 weight-decay settings $\times$ 3 seeds), as shown in \Cref{fig:integrability}.
Curvature is confined entirely to the normal bundle of the execution manifold; the submanifold is empirically invariant under the optimization flow.
We note that this near-unity value reflects the high dimensionality of parameter space ($P \approx 290$k) relative to the PCA subspace ($K = 16$--$24$); the exec/random ratio (\Cref{sec:random_control}) provides the complementary test that the small parallel component is geometrically meaningful.

This means that the dominant component of loss-landscape curvature lies \emph{outside} the directions the model actually uses for learning.
The weight trajectory evolves on an invariant submanifold: despite enormous curvature in the ambient parameter space, the curvature is confined to orthogonal directions and does not deflect the optimization trajectory out of its learned subspace.

\begin{figure}[t]
    \centering
    \begin{subfigure}[t]{0.48\textwidth}
        \centering
        \includegraphics[width=\textwidth]{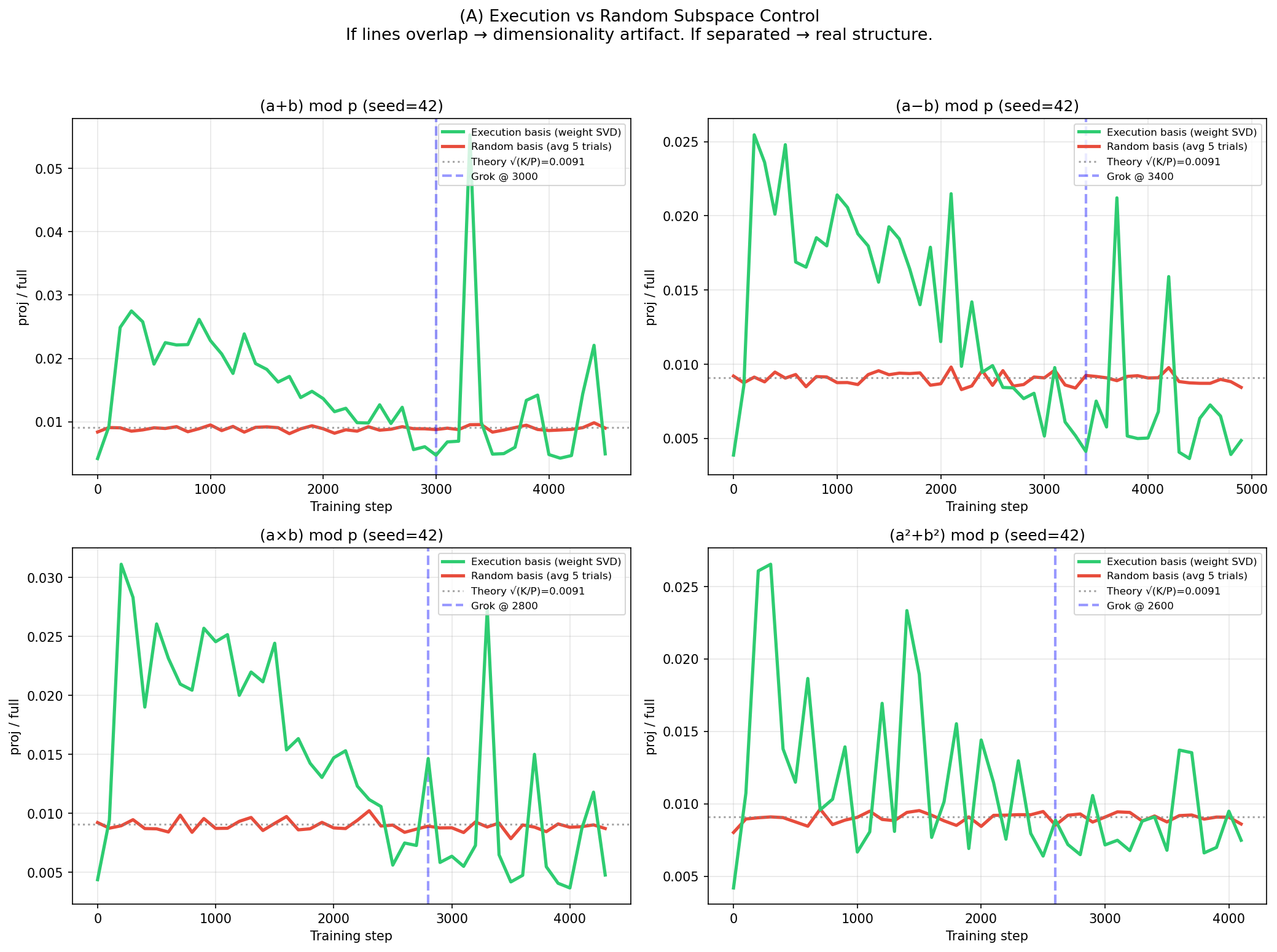}
        \caption{Projection fraction (proj/full) for execution basis (green) vs.\ random baseline (red) over training. The execution basis consistently captures more commutator energy.}
        \label{fig:exec_vs_random}
    \end{subfigure}
    \hfill
    \begin{subfigure}[t]{0.48\textwidth}
        \centering
        \includegraphics[width=\textwidth]{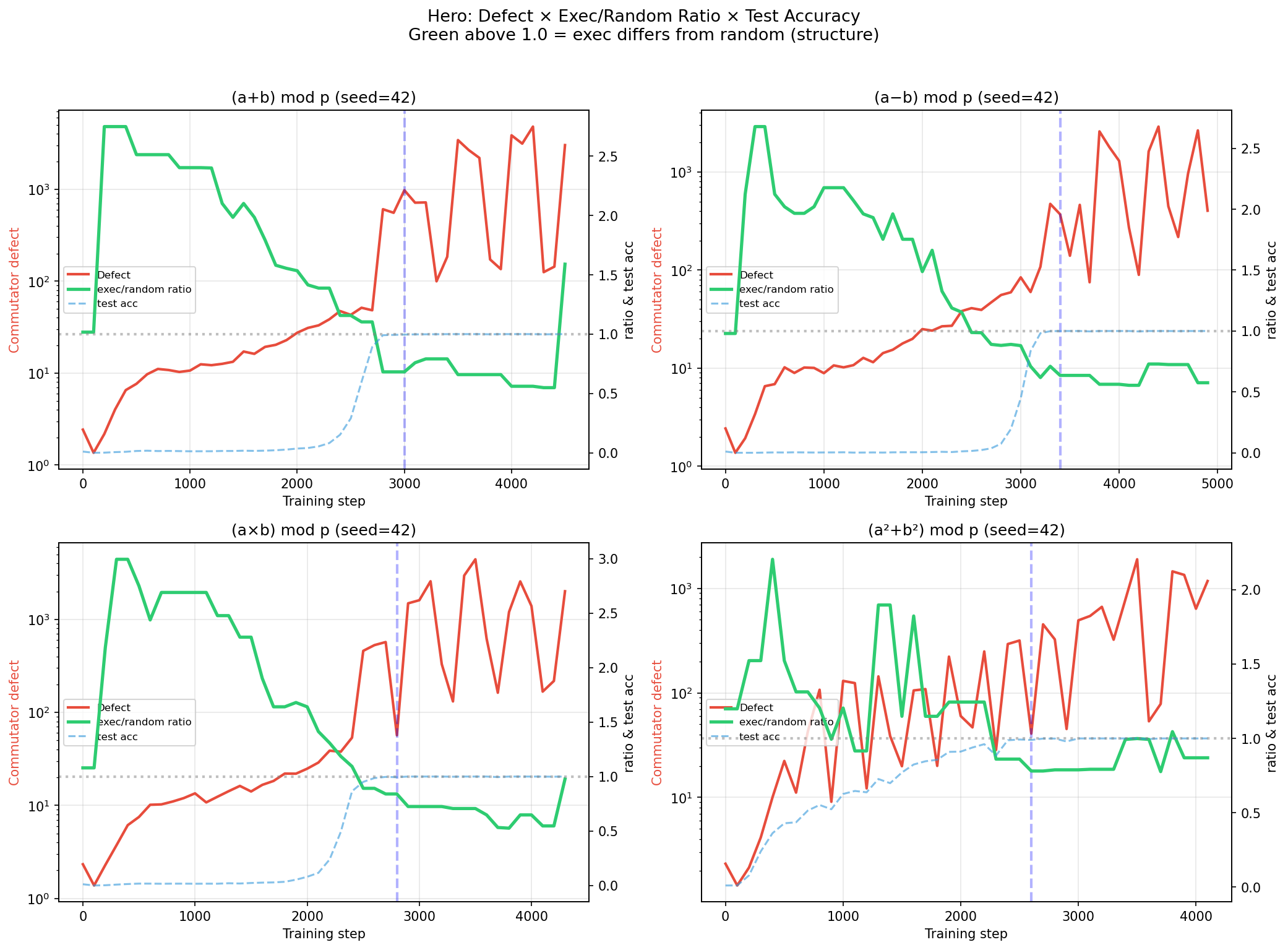}
        \caption{Combined view: commutator defect (red), exec/random ratio (green), and test accuracy (blue, dashed). The exec/random ratio is consistently above 1.0 during grokking.}
        \label{fig:hero_exec_random}
    \end{subfigure}
    \caption{Random subspace control confirms that the PCA projection is geometrically structured, not a dimensionality artifact. Exec/random ratio $\approx 1.8$--$2.9\times$ across operations.}
    \label{fig:random_control}
\end{figure}

\paragraph{Random subspace control.}
To confirm that the near-zero projection onto the PCA basis reflects genuine geometry rather than a dimensionality artifact, we compare against random $K$-dimensional subspaces (\Cref{sec:random_control}).
\Cref{fig:random_control} shows the projection fraction for the PCA (execution) basis and the random baseline over training.
Across all four grokking operations, the execution basis captures $1.8$--$2.9\times$ more commutator energy than a random subspace of equal dimension ($K = 24$), confirming that the small parallel component is geometrically structured.
We verify that this ratio is stable under variation of the PCA dimension $K$: reducing to $K = 16$ or increasing to $K = 32$ yields exec/random ratios within the same range, confirming that the result is not an artifact of the particular choice of $K$.
Crucially, both projections are very small (proj/full $< 0.05$), consistent with the invariance measure $\rho \approx 1.000$: in a space of $\sim$290k dimensions, any $K$-dimensional subspace captures negligible energy from a generic vector.
The PCA subspace nonetheless captures a structured excess above the random floor, confirming that the near-unity $\rho$ reflects genuine geometric orthogonality rather than merely high dimensionality.

\subsection{Curvature Explodes Orthogonally During Grokking}
\label{sec:curvature}

\begin{figure}[t]
    \centering
    \begin{subfigure}[t]{0.48\textwidth}
        \centering
        \includegraphics[width=\textwidth]{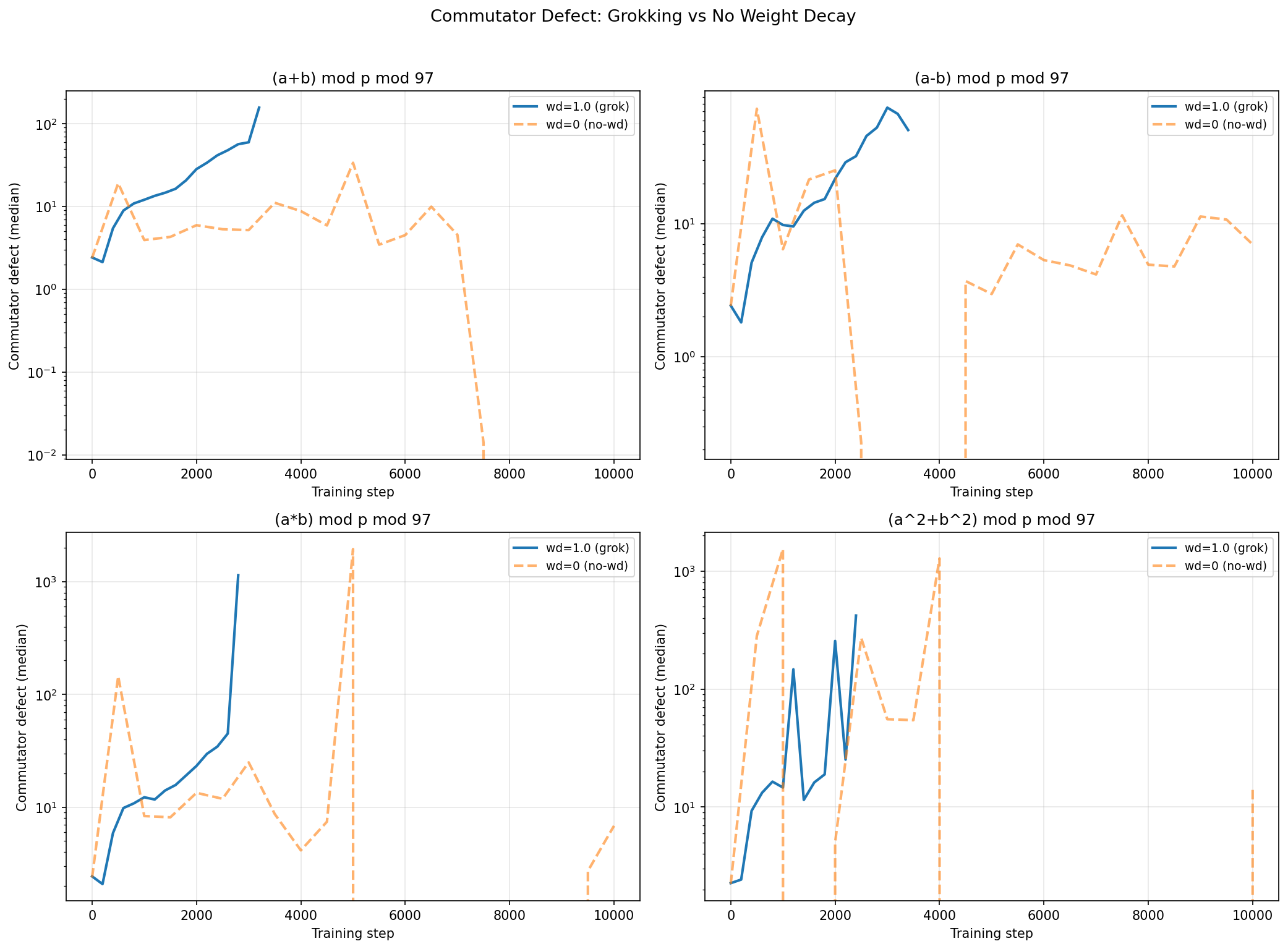}
        \caption{Grok (wd=1.0) vs.\ no-wd (wd=0.0): grokking runs develop substantially higher commutator defect.}
        \label{fig:grok_vs_nowd}
    \end{subfigure}
    \hfill
    \begin{subfigure}[t]{0.48\textwidth}
        \centering
        \includegraphics[width=\textwidth]{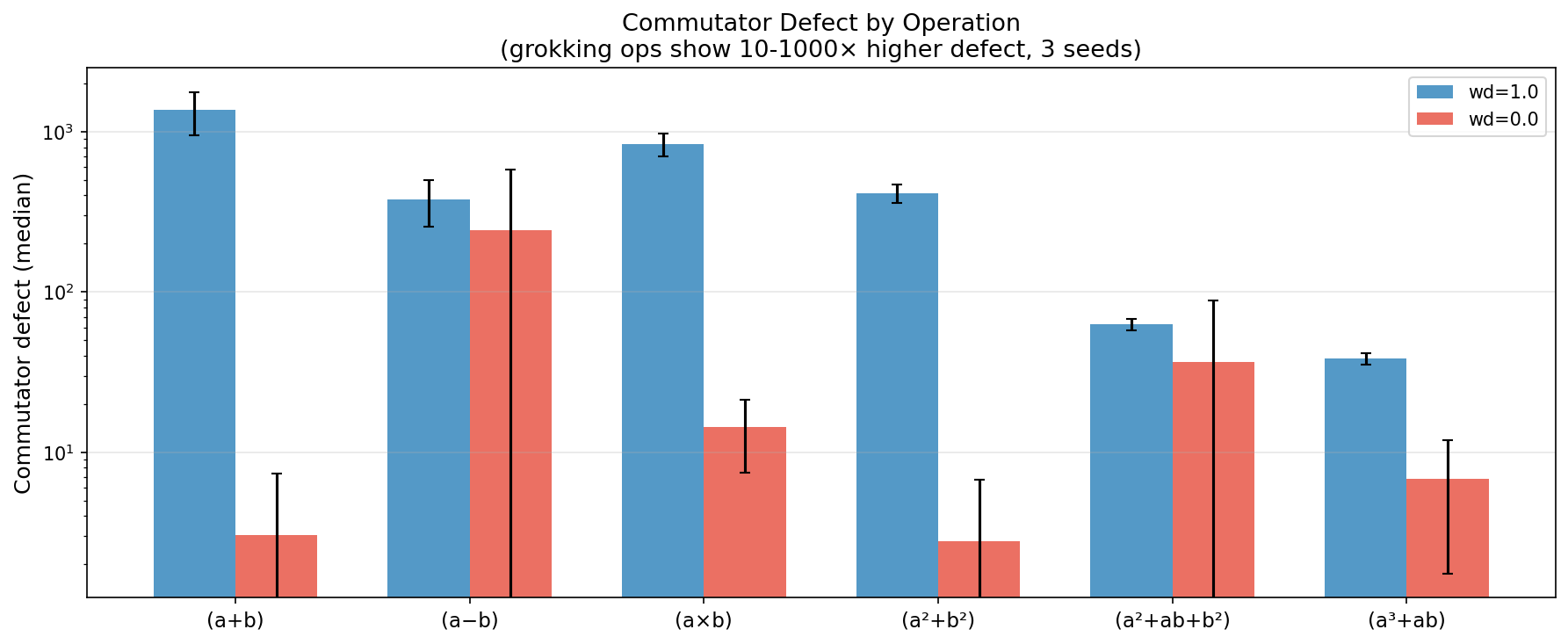}
        \caption{Commutator defect across all conditions (3 seeds). Grokking operations (wd=1.0) show 10--1000$\times$ higher defect.}
        \label{fig:defect_multiseed}
    \end{subfigure}
    \caption{Curvature explodes during grokking but remains orthogonal to the learned subspace.}
    \label{fig:defect_comparison}
\end{figure}

While the execution manifold remains empirically invariant (curvature confined to the normal bundle), the \emph{magnitude} of curvature in orthogonal directions changes substantially during grokking.
Operations that grok show 10--1000$\times$ higher commutator defect than non-grokking controls (\Cref{fig:defect_comparison}), and this curvature is concentrated predominantly outside the PCA manifold.

The converse analysis confirms that the weight trajectory does not align with curvature directions: the mean absolute cosine similarity between trajectory steps and commutator vectors is indistinguishable from the random-vector baseline ($\bar{c} \approx \sqrt{2/(\pi P)}$), meaning the trajectory actively avoids high-curvature directions.

\subsection{Curvature Growth Precedes Generalization}
\label{sec:prediction}

\begin{figure}[t]
    \centering
    \includegraphics[width=\textwidth]{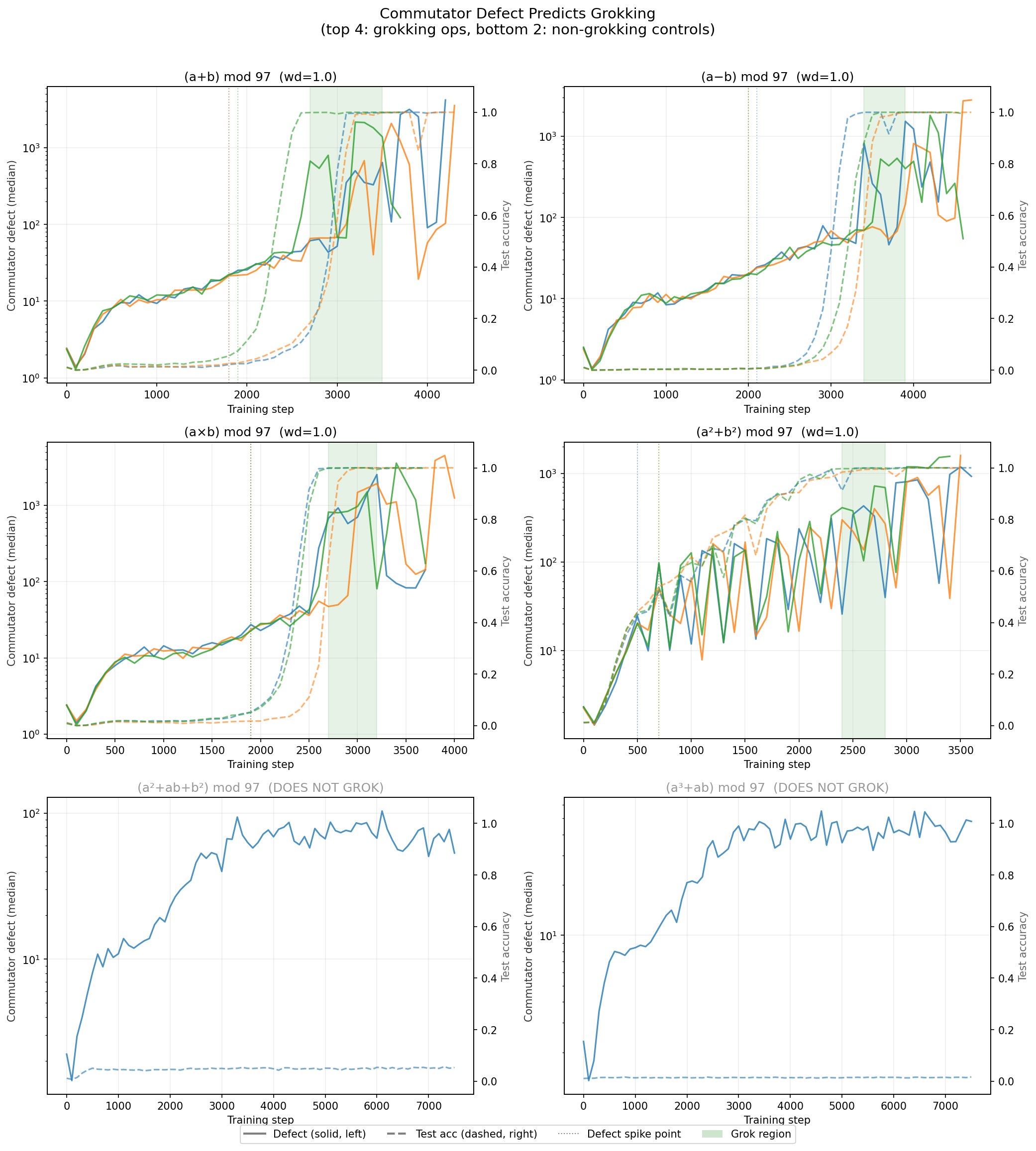}
    \caption{Temporal ordering of curvature growth and generalization. Top four panels: grokking operations (3 seeds each), showing defect (solid) rising before test accuracy (dashed). Bottom two panels: non-grokking controls show moderate defect growth ($30$--$50\times$ baseline) but no generalization. Dotted vertical lines mark defect onset; green regions mark grokking.}
    \label{fig:prediction}
\end{figure}

A key finding is that the onset of commutator defect growth consistently \emph{precedes} the generalization transition.
We define the \textbf{defect onset} as the first training step at which the commutator defect exceeds $10\times$ its early-training baseline (median of the first 3 measurements) and an absolute threshold of~20.
\Cref{fig:prediction} shows the temporal overlay of commutator defect and test accuracy for all four grokking operations and two non-grokking controls.
In every grokking run, defect begins rising before test accuracy increases.

\begin{figure}[t]
    \centering
    \begin{subfigure}[t]{0.48\textwidth}
        \centering
        \includegraphics[width=\textwidth]{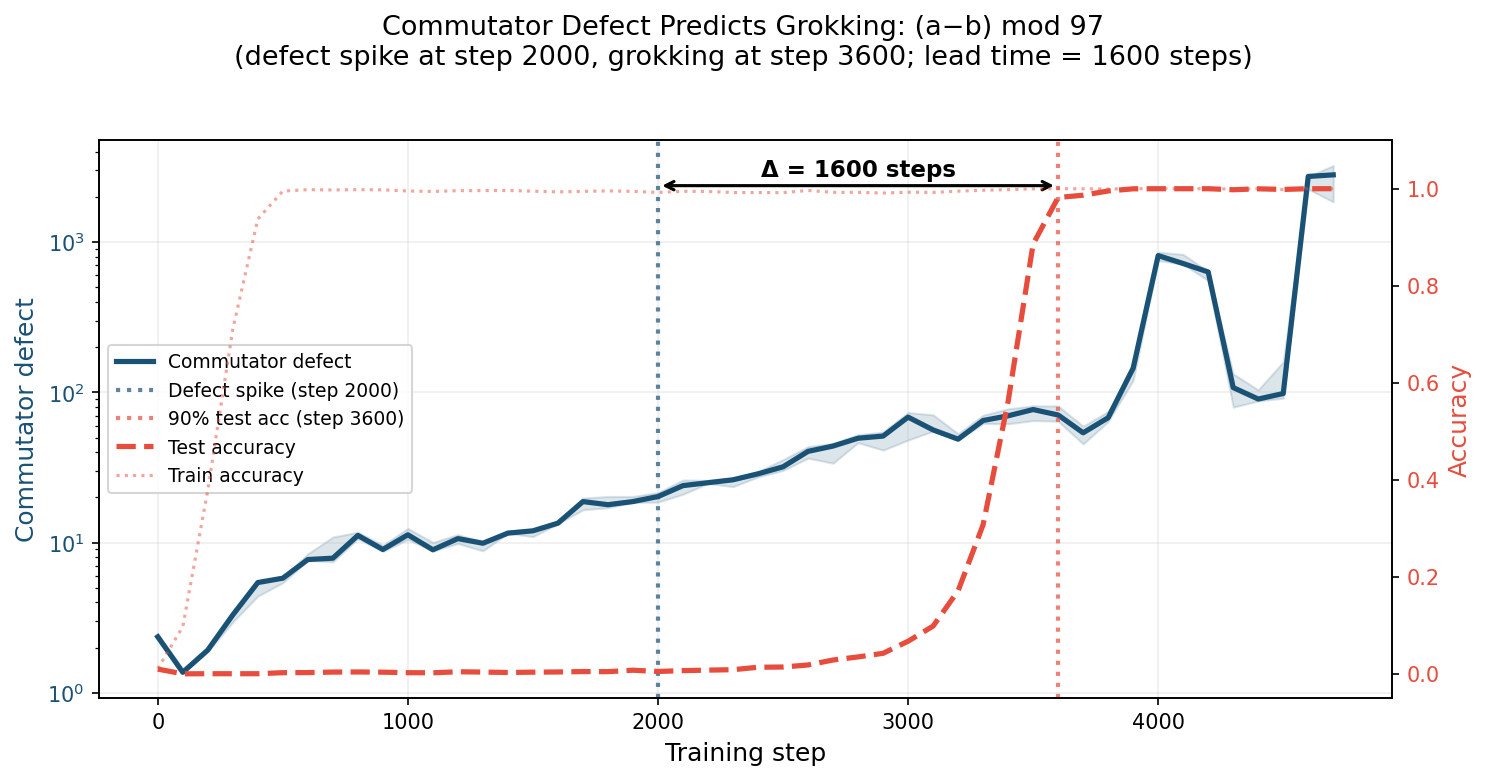}
        \caption{Hero example: $(a-b) \bmod 97$, seed 137. Defect onset at step 2000, grokking at step 3600 (lead = 1600 steps).}
        \label{fig:hero_defect_predicts}
    \end{subfigure}
    \hfill
    \begin{subfigure}[t]{0.48\textwidth}
        \centering
        \includegraphics[width=\textwidth]{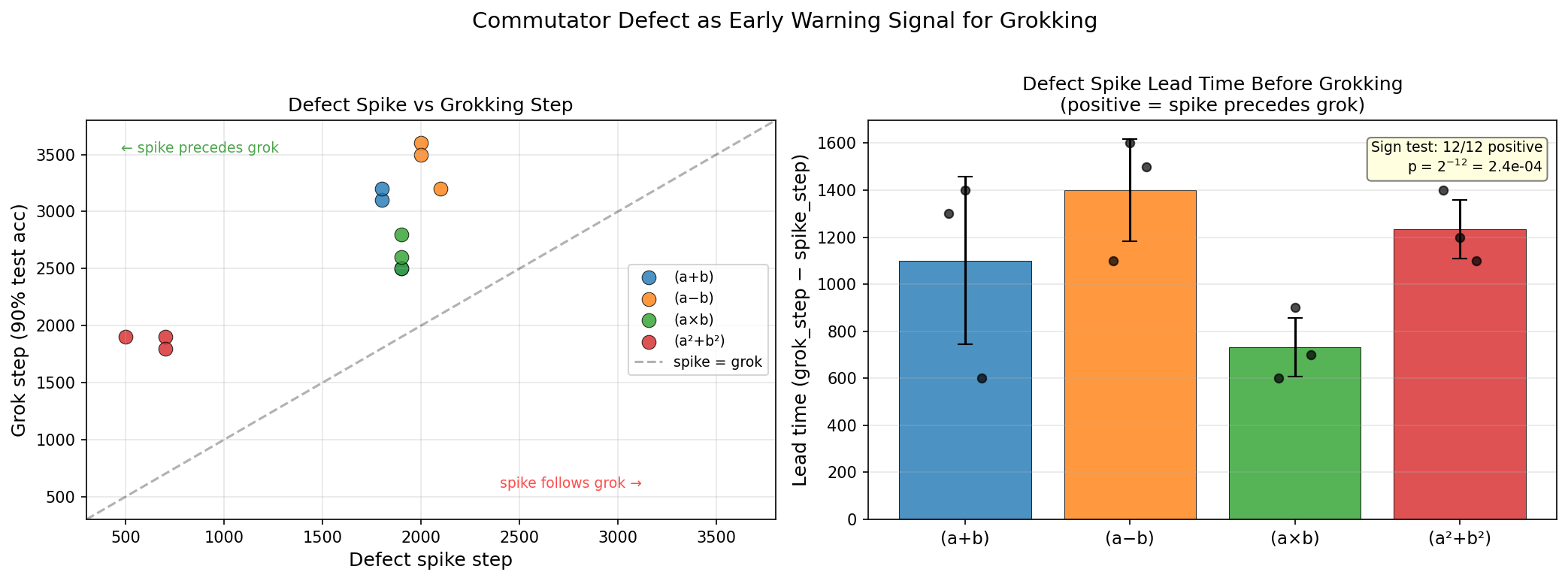}
        \caption{Lead time quantification. Left: onset step vs.\ grok step (all points above diagonal). Right: lead time by operation (sign test $p < 0.001$).}
        \label{fig:lead_time}
    \end{subfigure}
    \caption{Temporal ordering: curvature growth precedes generalization.}
    \label{fig:prediction_detail}
\end{figure}

We define the \textbf{lead time} as $\Delta t = t_\text{grok} - t_\text{onset}$, the number of training steps between defect onset and grokking, and the \textbf{lead fraction} as $\Delta t / t_\text{grok}$, the lead time normalized by the grokking timescale.
Across all 12 grokking runs (4 operations $\times$ 3 seeds), defect onset precedes the point at which test accuracy reaches 90\% by 600--1600 steps, with mean lead time of 1117 steps (\Cref{fig:lead_time}).
A one-sided sign test gives $p = 2^{-12} \approx 2.4 \times 10^{-4}$, confirming that the temporal ordering is statistically significant.

\paragraph{Necessary but not sufficient.}
Non-grokking operations ($a^2 + ab + b^2$ and $a^3 + ab$) also exhibit defect growth---reaching $30$--$50\times$ their early-training baseline---but never generalize.
Grokking operations, by contrast, reach $500$--$2000\times$ baseline, with zero overlap between the two groups in total growth magnitude (minimum grokking: $513\times$; maximum non-grokking: $47\times$).
However, this separation is only apparent retrospectively: at early training steps when grokking has not yet occurred, the defect trajectories of grokking and non-grokking operations overlap.
Defect onset is therefore best understood as a \emph{necessary precondition} for grokking---a signal that the loss landscape is reorganizing---rather than a sufficient predictor that discriminates which operations will generalize.
The causal intervention experiments (\Cref{sec:interventions}) confirm this interpretation: suppressing orthogonal gradient flow prevents grokking, establishing the mechanistic necessity of curvature growth.

\subsection{Regime Invariance}
\label{sec:regime}

To verify that our findings are not specific to a particular hyperparameter setting, we repeat the full analysis in a slow regime with qualitatively different hyperparameters (\Cref{tab:regimes}).

\begin{table}[ht]
\centering
\caption{Hyperparameter regimes and key metrics.}
\label{tab:regimes}
\begin{tabular}{@{}lcc@{}}
\toprule
Metric & Fast regime & Slow regime \\
\midrule
Learning rate & $10^{-3}$ & $5 \times 10^{-5}$ \\
Weight decay & 1.0 & 0.1 \\
Layers & 2 & 3 \\
Adam $\beta_2$ & 0.98 & 0.999 \\
\midrule
Grok step (add, mean) & $\sim$2,900 & $\sim$570,000 \\
Invariance ($\rho$) & $\approx 1.000$ & $\approx 1.000$ \\
Onset precedes grok? & 12/12 runs & 2/2 runs \\
Lead time (absolute) & $\sim$1,100 steps & $\sim$558,000 steps \\
Lead time (normalized) & $\sim$0.38 & $\sim$0.55 \\
\bottomrule
\end{tabular}
\end{table}

\begin{figure}[t]
    \centering
    \begin{subfigure}[t]{0.48\textwidth}
        \centering
        \includegraphics[width=\textwidth]{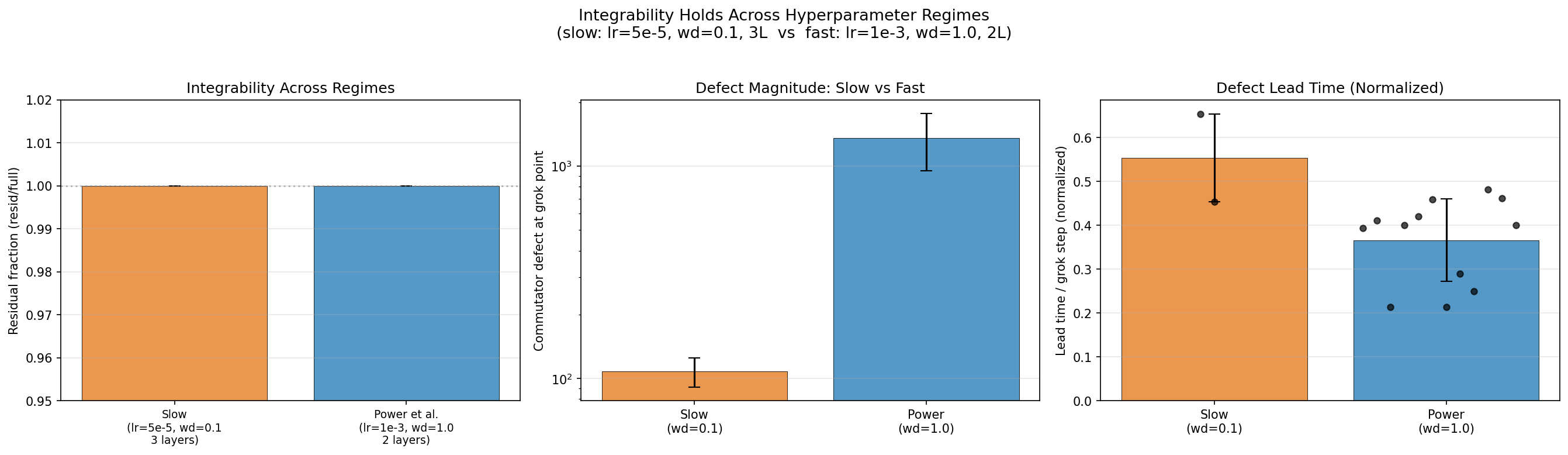}
        \caption{Regime comparison: invariance, defect, and normalized lead time are consistent across regimes.}
        \label{fig:regime_comparison}
    \end{subfigure}
    \hfill
    \begin{subfigure}[t]{0.48\textwidth}
        \centering
        \includegraphics[width=\textwidth]{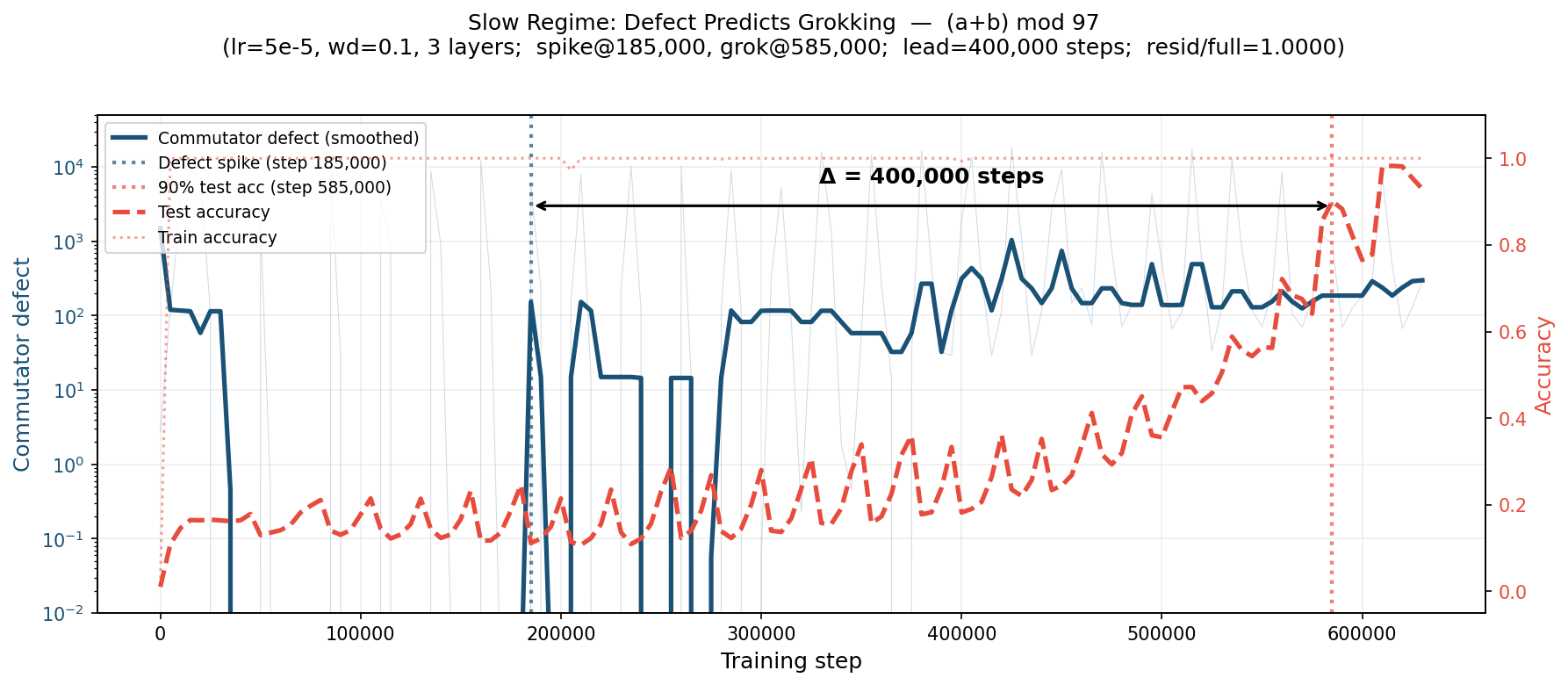}
        \caption{Slow regime hero: defect onset at $\sim$185k, grokking at $\sim$585k (lead $\approx$ 400k steps).}
        \label{fig:slow_hero}
    \end{subfigure}
    \caption{Regime invariance: qualitative findings replicate in the slow regime ($200\times$ longer training), though alignment dynamics differ quantitatively.}
    \label{fig:slow_regime}
\end{figure}

\Cref{fig:slow_regime} shows the slow-regime results.
Despite a $200\times$ difference in grokking timescale, $10\times$ difference in weight decay, and a different number of layers, the same qualitative transition is observed: the execution manifold exhibits empirical invariance ($\rho \approx 1.000$), and defect onset precedes grokking by hundreds of thousands of steps.
However, the alignment dynamics differ quantitatively between regimes.
In the fast regime, trajectory--curvature alignment is initially above the random baseline and decays toward the transition, consistent with underdamped exploration of parameter space.
In the slow regime, alignment remains at or below the random baseline throughout, consistent with overdamped motion along narrow valleys.
Both regimes nonetheless exhibit a defect-mediated generalization transition.

\subsection{Learning Rate Phase Diagram}
\label{sec:lr_sweep}

Having established regime invariance between qualitatively different configurations, we now systematically vary the learning rate alone to map the phase boundary of grokking dynamics.
We sweep $\eta \in \{3\!\times\!10^{-5},\, 10^{-4},\, 3\!\times\!10^{-4},\, 10^{-3},\, 3\!\times\!10^{-3},\, 10^{-2}\}$ (half-decade log-uniform) with fixed $\lambda = 1.0$ across all six operations and three seeds (108 runs total).

\begin{figure}[t]
    \centering
    \includegraphics[width=\textwidth]{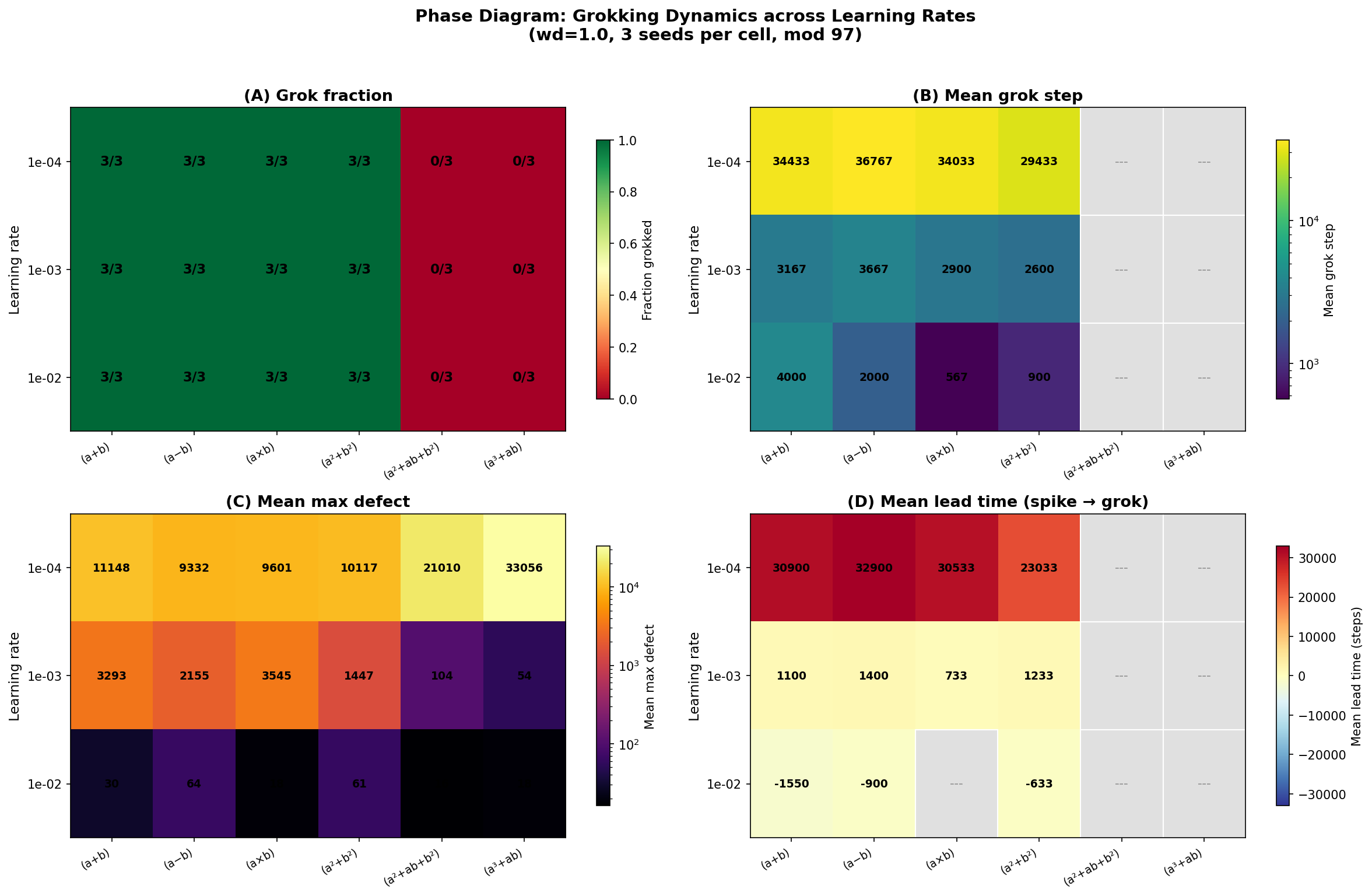}
    \caption{Phase diagram of grokking dynamics across learning rates ($\eta \in \{3\!\times\!10^{-5}\text{--}10^{-2}\}$, $\lambda = 1.0$, 3 seeds per cell, 108 runs).
    (A)~Grok fraction: the phase boundary between grokking and non-grokking operations is invariant to learning rate.
    (B)~Mean grok step (log scale): grokking speed scales with $\eta$.
    (C)~Mean max defect (log scale): curvature explosion is largest at low $\eta$.
    (D)~Mean lead time (onset step $-$ grok step): defect onset is most predictive at low $\eta$.}
    \label{fig:lr_phase}
\end{figure}

\Cref{fig:lr_phase} shows the resulting phase diagram.
The grok/no-grok boundary is \emph{invariant} to learning rate: the same four operations grok at all six rates, while the two complex operations never grok (\Cref{fig:lr_phase}A).
Grokking speed scales roughly linearly with $\eta$: mean grok steps range from $\sim$136k at $\eta = 3\!\times\!10^{-5}$ to $\sim$200 at $\eta = 10^{-2}$, spanning nearly three orders of magnitude (\Cref{fig:lr_phase}B).

The defect landscape reveals a striking asymmetry (\Cref{fig:lr_phase}C): at low learning rate, maximum defect reaches $10^4$, while at high learning rate it drops to $\sim$20--60.
This suggests that slower optimization allows curvature to accumulate more in the orthogonal bundle before the phase transition occurs.

\begin{figure}[t]
    \centering
    \includegraphics[width=\textwidth]{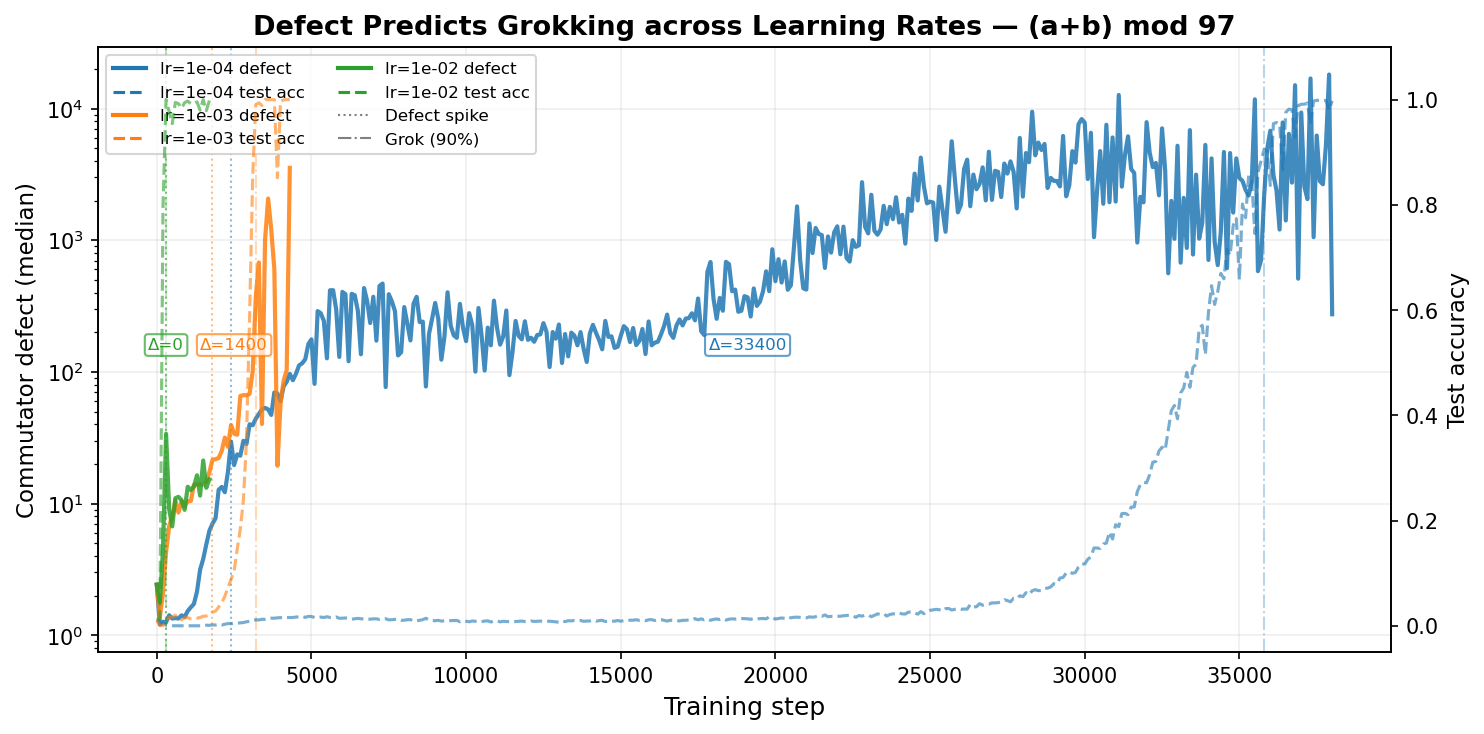}
    \caption{Defect and test accuracy trajectories for addition across six learning rates. At low $\eta$ (e.g., $10^{-4}$), defect onset precedes grokking by $\sim$30k steps (92\% of training); at high $\eta$ ($\geq 3\!\times\!10^{-3}$), grokking is nearly instantaneous and onset is concurrent.}
    \label{fig:lr_hero}
\end{figure}

\begin{figure}[t]
    \centering
    \includegraphics[width=\textwidth]{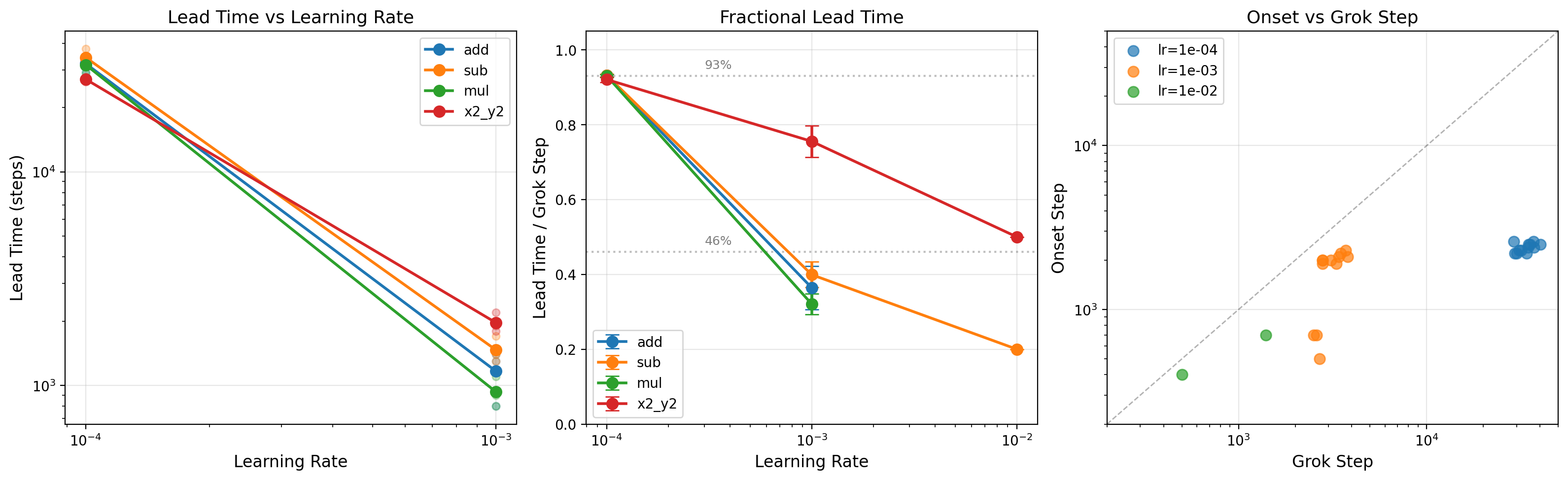}
    \caption{Lead time scaling across learning rates.
    (A)~Power-law fit: lead time $\propto$ grok step$^{\alpha}$ with $\alpha = 1.27 \pm 0.03$ ($R^2 = 0.97$, $n = 43$). Each point is one grokking run; the dashed line shows the best fit, the dotted line shows linear scaling ($\alpha = 1$).
    (B)~Lead fraction (lead / grok step) increases monotonically as $\eta$ decreases: from 24\% at $\eta = 3\!\times\!10^{-3}$ to 95\% at $\eta = 3\!\times\!10^{-5}$.
    (C)~Onset step vs.\ grok step for each operation.}
    \label{fig:lr_scaling}
\end{figure}

The predictive lead time (\Cref{fig:lr_phase}D) is largest at $\eta = 3\!\times\!10^{-5}$ ($\sim$128k steps, 95\% of training time) and decreases monotonically with learning rate: 92\% at $\eta = 10^{-4}$, 88\% at $\eta = 3\!\times\!10^{-4}$, 43\% at $\eta = 10^{-3}$, and 24\% at $\eta = 3\!\times\!10^{-3}$.
A log-log regression across all grokking runs with detectable onset yields a power-law exponent $\alpha = 1.27 \pm 0.03$ ($R^2 = 0.97$, $n = 43$; \Cref{fig:lr_scaling}), confirming that the lead time grows \emph{super-linearly} with grokking timescale.
This means the predictive window improves at slower, more realistic learning rates.
At $\eta \geq 3\!\times\!10^{-3}$, grokking occurs so rapidly ($<$1k steps) that the memorization and generalization phases overlap, and defect onset is concurrent with rather than predictive of grokking.
\Cref{fig:lr_hero} illustrates these regimes for the addition operation.

\begin{figure}[t]
    \centering
    \includegraphics[width=\textwidth]{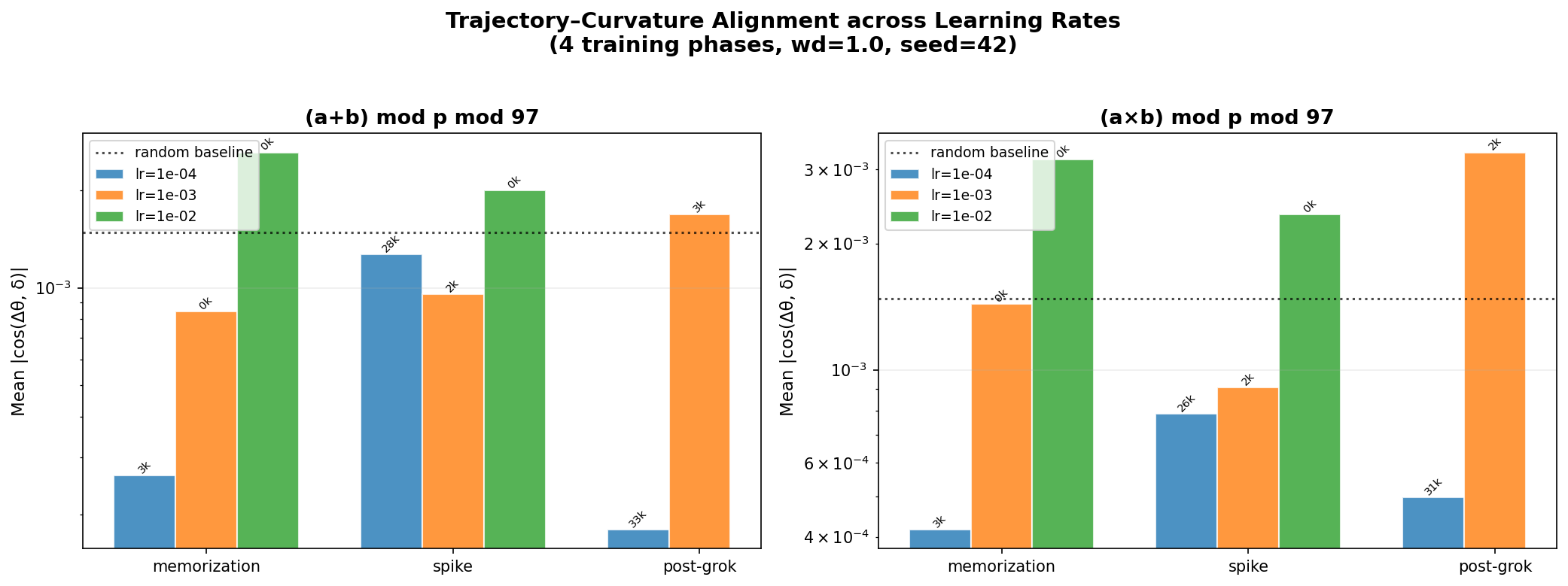}
    \caption{Trajectory--curvature alignment at three training phases across learning rates.
    At $\eta = 10^{-4}$ (blue), alignment stays below the random baseline (dotted), consistent with overdamped dynamics.
    At $\eta = 10^{-2}$ (green), alignment exceeds the baseline, consistent with underdamped exploration.
    $\eta = 10^{-3}$ (orange) shows intermediate behavior.
    Both operations exhibit the same pattern.}
    \label{fig:lr_alignment}
\end{figure}

\paragraph{LR-dependent alignment dynamics.}
To directly test whether the damping regime varies with learning rate, we measure trajectory--curvature alignment (mean $|\cos(\Delta\theta, \delta)|$, \Cref{sec:converse}) at four strategic checkpoints---memorization, defect onset, and post-grok---for each of the three learning rates on two operations (\Cref{fig:lr_alignment}).
At $\eta = 10^{-4}$, alignment remains below the random baseline ($0.18$--$0.86\times$) throughout, consistent with overdamped dynamics where the trajectory is confined to narrow valleys far from curvature directions.
At $\eta = 10^{-2}$, alignment is consistently \emph{above} the baseline ($1.4$--$1.8\times$), indicating underdamped exploration that initially samples curvature directions.
At $\eta = 10^{-3}$, the intermediate regime, alignment starts below baseline and rises toward or above it at the grokking transition.
This LR-dependent pattern replicates across both operations and provides direct evidence for the dynamical regimes discussed below.

\begin{figure}[t]
    \centering
    \includegraphics[width=0.85\textwidth]{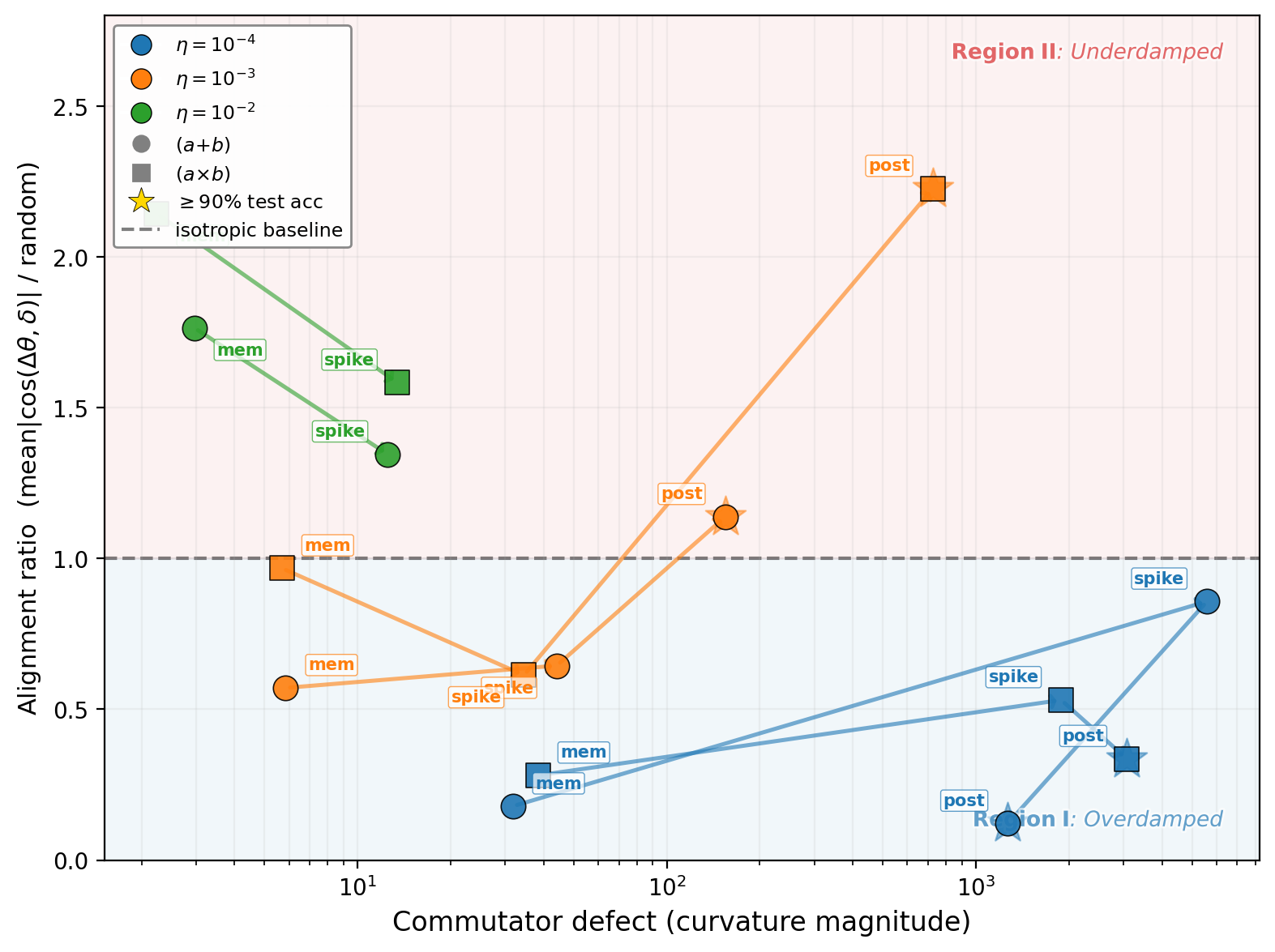}
    \caption{Phase portrait of grokking dynamics in curvature--trajectory space.
    We plot the trajectory--curvature alignment ratio (mean\,$|\!\cos(\Delta\theta,\delta)\!|$ normalized by random baseline) against the commutator defect magnitude for three learning rates ($\eta = 10^{-4}, 10^{-3}, 10^{-2}$).
    Each polyline traces the evolution from memorization (\textsc{mem}) through defect onset (\textsc{spike}) to the post-grokking regime (\textsc{post}), shown for addition (circles) and multiplication (squares); arrows indicate the direction of training.
    The horizontal dashed line indicates isotropic alignment.
    Training at high learning rates remains in an underdamped regime characterized by strong alignment and low defect, while low learning rates produce overdamped dynamics with large defect accumulation and weak alignment.
    Intermediate learning rates interpolate between these regimes.
    Stars mark checkpoints where test accuracy exceeds $90\%$.
    Grokking corresponds to escape from a metastable region of high curvature defect and reduced mobility, with regime-dependent relaxation dynamics.}
    \label{fig:alignment_vs_defect}
\end{figure}

To integrate curvature accumulation and trajectory geometry into a unified picture, we construct a reduced phase portrait using the commutator defect and the trajectory--curvature alignment as coordinates (\Cref{fig:alignment_vs_defect}).
Each training run traces a characteristic path through this space, progressing from memorization through defect onset to the post-grokking regime.

We observe three qualitatively distinct dynamical regimes controlled by the learning rate.
At high learning rates, training remains in an underdamped regime, exhibiting strong alignment with curvature directions and low defect accumulation.
At low learning rates, training becomes overdamped, with prolonged confinement to low-alignment regions and substantial defect buildup prior to grokking.
Intermediate learning rates interpolate between these behaviors, producing critically damped trajectories.

Across both addition and multiplication tasks, grokking occurs when trajectories exit a metastable region characterized by high curvature defect and suppressed mobility.
This phase portrait provides a compact geometric representation of the grokking transition and clarifies how optimization hyperparameters control the pathway to algorithmic generalization.
To our knowledge, this is the first identification of distinct overdamped, critically damped, and underdamped dynamical regimes in grokking, suggesting that the phenomenon possesses a richer phase structure than previously recognized.

\subsection{Causal Interventions on Learning Dynamics}
\label{sec:interventions}

The preceding sections establish correlational evidence for the geometric picture.
We now test three falsifiable hypotheses through targeted intervention experiments that modify the optimization trajectory while preserving the underlying architecture and dataset:
\begin{enumerate}
    \item \textbf{Necessity}: Does grokking require motion along the learned execution manifold?
    \item \textbf{Sufficiency}: Does artificially inducing transverse curvature suffice to trigger grokking?
    \item \textbf{Specificity}: Are these effects specific to the PCA directions, or would any low-dimensional constraint produce the same result?
\end{enumerate}

\subsubsection{Gradient Subspace Suppression}
\label{sec:suppression}

We first examine whether motion along the learned execution manifold is necessary for grokking.
At each training step after step 500 (post-memorization), we project the gradient onto the subspace spanned by the top principal components of the weight trajectory, with projection strength $s \in [0, 1]$:
\begin{equation}
    g \;\longrightarrow\; g_\parallel + (1 - s)\, g_\perp, \qquad g_\parallel = B\, B^\top g,\quad g_\perp = g - g_\parallel,
\end{equation}
where $B \in \R^{P \times K}$ is the PCA basis from Phase~1 training.
For comparison, we also apply random low-dimensional projections of equal rank ($K = 16$).

\begin{figure}[t]
    \centering
    \includegraphics[width=0.85\textwidth]{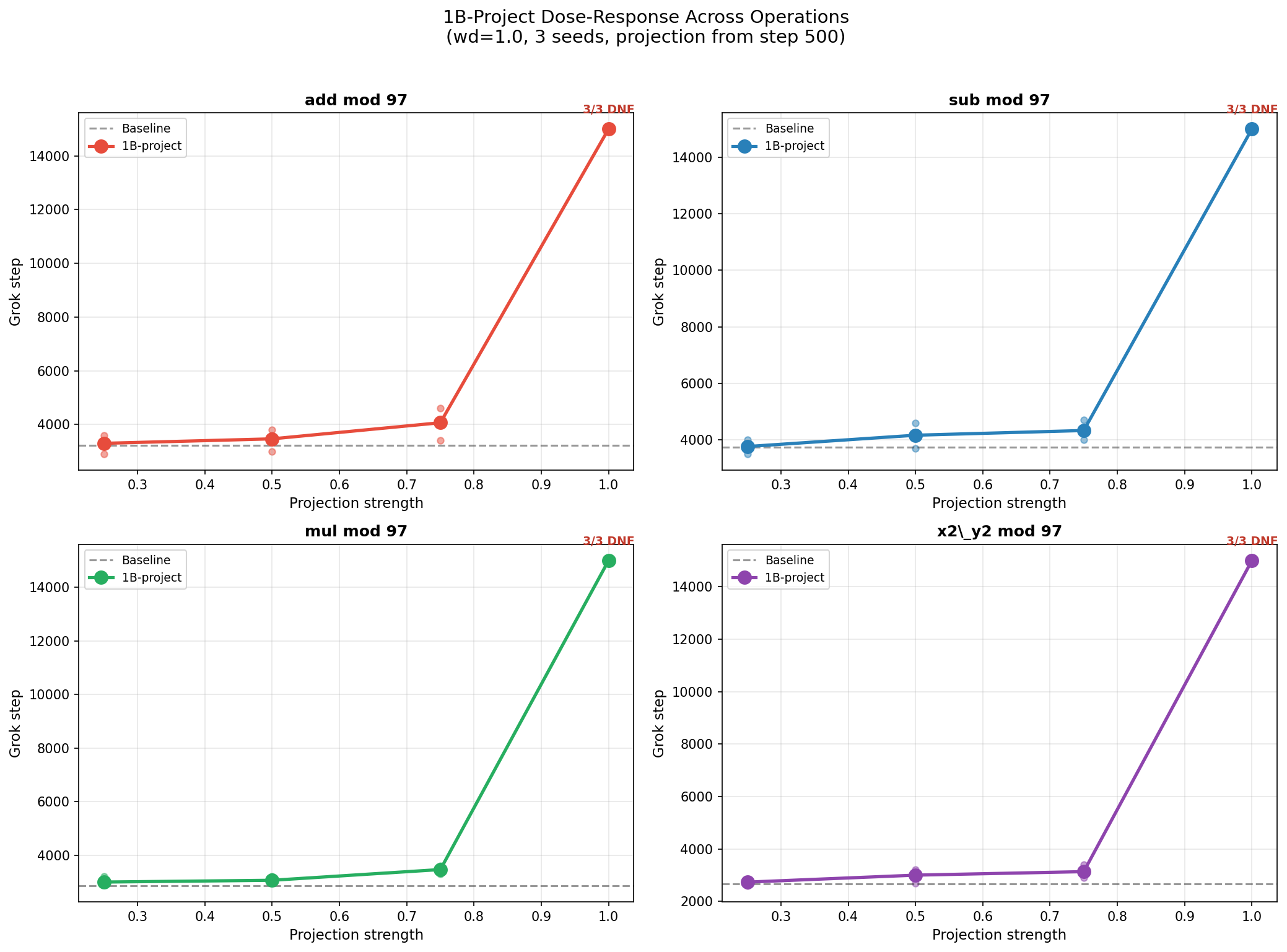}
    \caption{Dose--response curve for gradient projection across all four grokking operations.
    Each panel shows mean grok step (3 seeds) vs.\ suppression strength $s$.
    At $s = 1.0$, grokking fails universally (0/12 seeds).
    Dashed line: baseline (no intervention).
    The monotonic delay and complete suppression at full strength replicate across all operations.}
    \label{fig:multiop_dose}
\end{figure}

\begin{figure}[t]
    \centering
    \includegraphics[width=\textwidth]{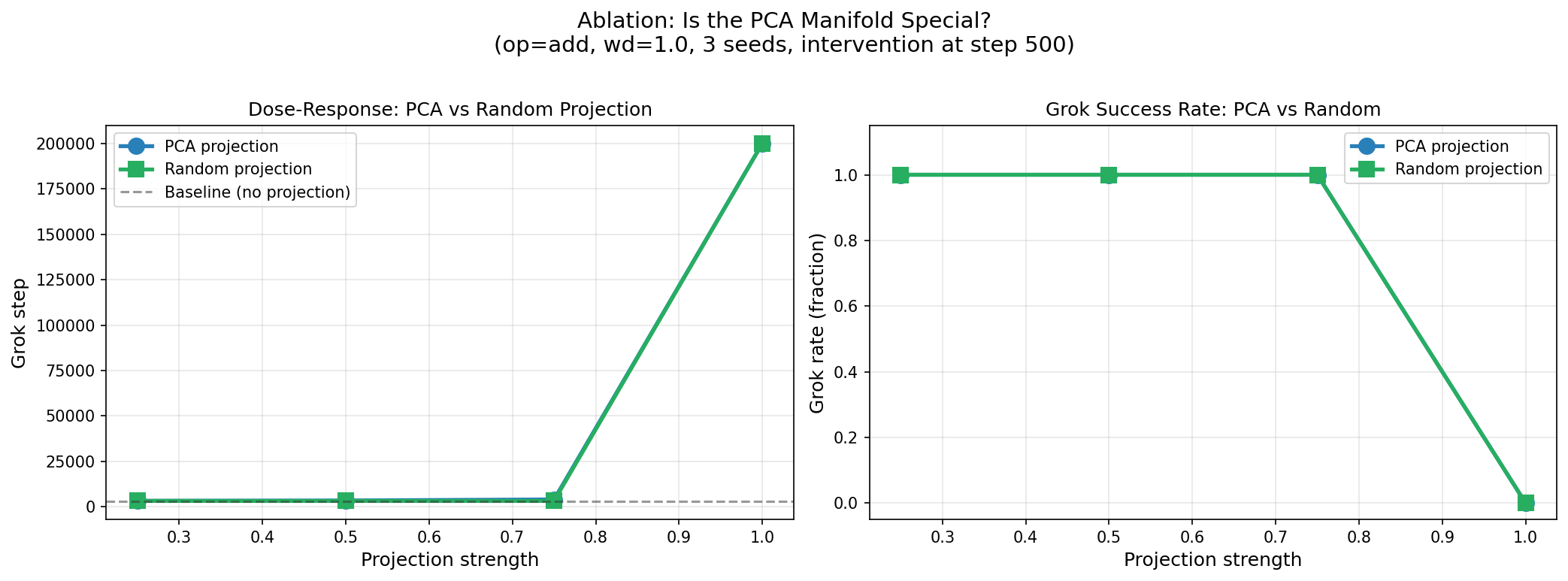}
    \caption{PCA-specific suppression control.
    Left: grok step vs.\ suppression strength for PCA projection (blue) and random projection (green); dashed line is baseline.
    Right: grok success rate.
    At intermediate strengths ($s = 0.25$--$0.75$), PCA projection monotonically delays grokking while random projection has no effect.
    At $s = 1.0$, both kill grokking (any 16-dim constraint is too restrictive).
    The dose--response separation confirms the geometric specificity of the PCA manifold.}
    \label{fig:ablation_pca_random}
\end{figure}

Partial suppression along the PCA directions ($s = 0.25$--$0.75$) systematically delays grokking, while full projection ($s = 1.0$) completely prevents generalization (0/12 seeds across four operations; \Cref{fig:multiop_dose}).
In contrast, random projections have little effect at intermediate strengths ($<$\,50-step difference from baseline; \Cref{fig:ablation_pca_random}).
At $s = 1.0$, both projections kill grokking, since confining the optimizer to \emph{any} 16-dimensional subspace of $\R^{290\text{k}}$ is too restrictive.
These results indicate that grokking requires access to specific learned directions in parameter space, rather than arbitrary low-dimensional motion.

\subsubsection{Directional Forcing and Defect Induction}
\label{sec:induction}

Next, we test whether artificially inducing curvature defects is sufficient to trigger early grokking.
Starting at step 500, we periodically inject additive weight updates aligned with the commutator direction (recomputed every 50 steps), with amplitudes $\alpha \in \{50, 100, 200, 500\}$ times the gradient step norm.
As a control, we apply kicks of equal magnitude along random orthogonal directions.

\begin{figure}[t]
    \centering
    \includegraphics[width=\textwidth]{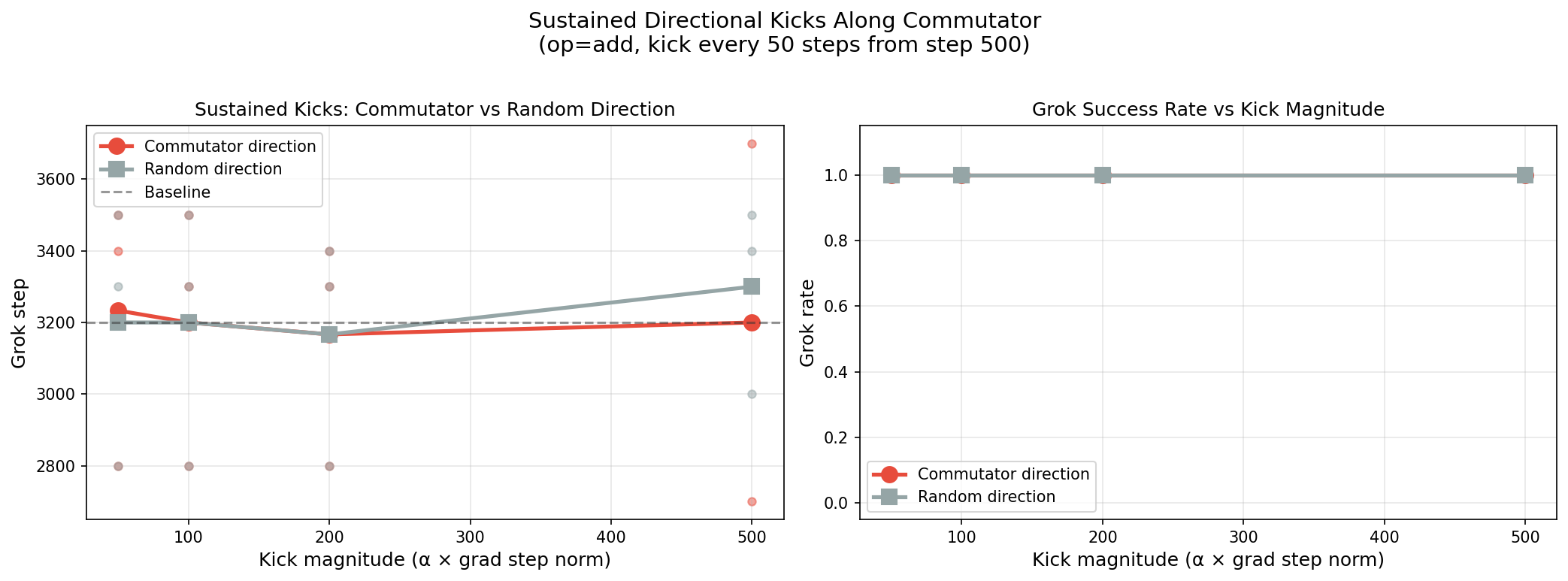}
    \caption{Sustained directional kicks along the commutator (red) vs.\ random orthogonal (gray) directions, with kick magnitudes up to $500\times$ the gradient step norm applied every 50 steps.
    Left: mean grok step (3 seeds); right: grok success rate.
    Neither direction accelerates grokking beyond baseline variability (dashed line), confirming that the orthogonal defect is not sufficient to induce the phase transition.}
    \label{fig:sustained_kicks}
\end{figure}

Across all tested amplitudes, neither commutator-aligned nor random kicks accelerate grokking relative to baseline (\Cref{fig:sustained_kicks}).
All 27/27 runs generalize at statistically indistinguishable times ($\sim$3200 steps, within seed-to-seed variability).
This negative result demonstrates that defect accumulation alone is insufficient to induce grokking, and that escape from the metastable regime requires coordinated motion along learned directions.

\subsubsection{Replication Across Operations}
\label{sec:intervention_replication}

We repeat the projection experiments across all four grokking tasks: modular addition, subtraction, multiplication, and quadratic addition ($a^2 + b^2 \bmod 97$).
The dose--response relationship between projection strength and grokking delay is consistent across all operations (\Cref{fig:multiop_dose}), with complete suppression at full strength (0/12 seeds grok at $s = 1.0$).
At $s = 0.75$, grokking is delayed by 600--800 steps (20--25\% above baseline) across all four operations.
This universality suggests that the causal role of execution-manifold directions is not task-specific, but reflects a common geometric mechanism underlying algorithmic generalization.

\subsubsection{Summary of Interventions}
\label{sec:intervention_summary}

Taken together, these experiments resolve all three hypotheses:
\emph{Necessity}---confirmed: constraining motion along the execution manifold prevents generalization with a smooth dose--response curve that replicates across four operations.
\emph{Sufficiency}---rejected: artificially increasing defect through directional forcing has no effect.
\emph{Specificity}---confirmed: PCA projection monotonically delays grokking at intermediate strengths where random projection has no effect.
This asymmetry is consistent with the commutator defect serving as a \emph{signature} of the curvature barrier between memorization and generalization solutions---a structured reorganization of the optimization trajectory---rather than a directly manipulable cause of the phase transition.
Together, these interventions rule out purely correlational explanations of our earlier findings and establish a directional causal relationship between execution-manifold geometry and generalization.

\section{Spectral Mechanism Underlying the Commutator Transition}
\label{sec:spectral}

The preceding sections establish that the SGD commutator defect rises sharply before grokking and that orthogonal gradient flow is causally necessary for generalization.
We now ask: \emph{what drives the commutator dynamics?}
To answer this, we analyze the singular value spectrum of the attention weight matrices $W_Q$ and $W_K$ at each training checkpoint.

\subsection{Spectral Gaps of the Attention Matrices}
\label{sec:spectral_gaps}

Let $\sigma_1 \ge \sigma_2 \ge \sigma_3 \ge \cdots$ denote the singular values of $W_Q$ (the analysis for $W_K$ is analogous).
We define the \emph{spectral gaps}
\begin{equation}
    g_{12} = \sigma_1 - \sigma_2, \qquad g_{23} = \sigma_2 - \sigma_3.
\end{equation}
These quantities track how separated the leading singular modes are: $g_{12} \approx 0$ indicates near-degeneracy of the top two modes, while $g_{12} \gg 0$ indicates that one mode dominates.

\subsection{A Consistent Spectral Timeline}
\label{sec:spectral_timeline}

Across all grokking runs (4 operations $\times$ 3 seeds), we observe the following consistent temporal sequence:
\begin{enumerate}
    \item The lower spectral gap $g_{23}$ gradually shrinks, indicating compression of the sub-leading spectrum.
    \item The leading modes become nearly degenerate ($\sigma_1 \approx \sigma_2$, with $g_{12}$ reaching a minimum of $0.002$--$0.02$).
    \item The SGD commutator defect $\defect$ rises sharply.
    \item The matrix commutator $\norm{[W_Q, W_K]}_F$ reaches its peak.
    \item One mode becomes dominant ($\sigma_1 \gg \sigma_2$, with $g_{12}$ increasing $15$--$25\times$ from its minimum).
    \item Test accuracy rapidly increases (grokking).
\end{enumerate}

\noindent
For the representative case of modular addition (seed~42), the ordering is:
\begin{equation}
    \underbrace{g_{23}\!\downarrow}_{\text{step 1400}}
    \;\to\;
    \underbrace{\sigma_1 \approx \sigma_2}_{\text{step 1700}}
    \;\to\;
    \underbrace{\defect\!\uparrow}_{\text{step 2000}}
    \;\to\;
    \underbrace{\norm{[W_Q,W_K]}_F\!\uparrow}_{\text{step 2700}}
    \;\to\;
    \underbrace{\sigma_1 \gg \sigma_2}_{\text{step 2800}}
    \;\to\;
    \underbrace{\text{grok}}_{\text{step 3100}}
\end{equation}

\noindent
The near-degeneracy phase ($\sigma_1 \approx \sigma_2$) corresponds to a regime in which the representation basis is \emph{unstable}: there is no preferred direction in the top singular subspace, so the optimizer dynamics become highly non-integrable, as reflected in the sharp rise of the SGD commutator defect.
Once the symmetry between the top modes breaks and a dominant direction emerges, the operators $W_Q$ and $W_K$ align into a shared eigenbasis and the matrix commutator collapses.

\subsection{Phase Portrait}
\label{sec:phase_portrait}

\begin{figure}[t]
    \centering
    \includegraphics[width=\textwidth]{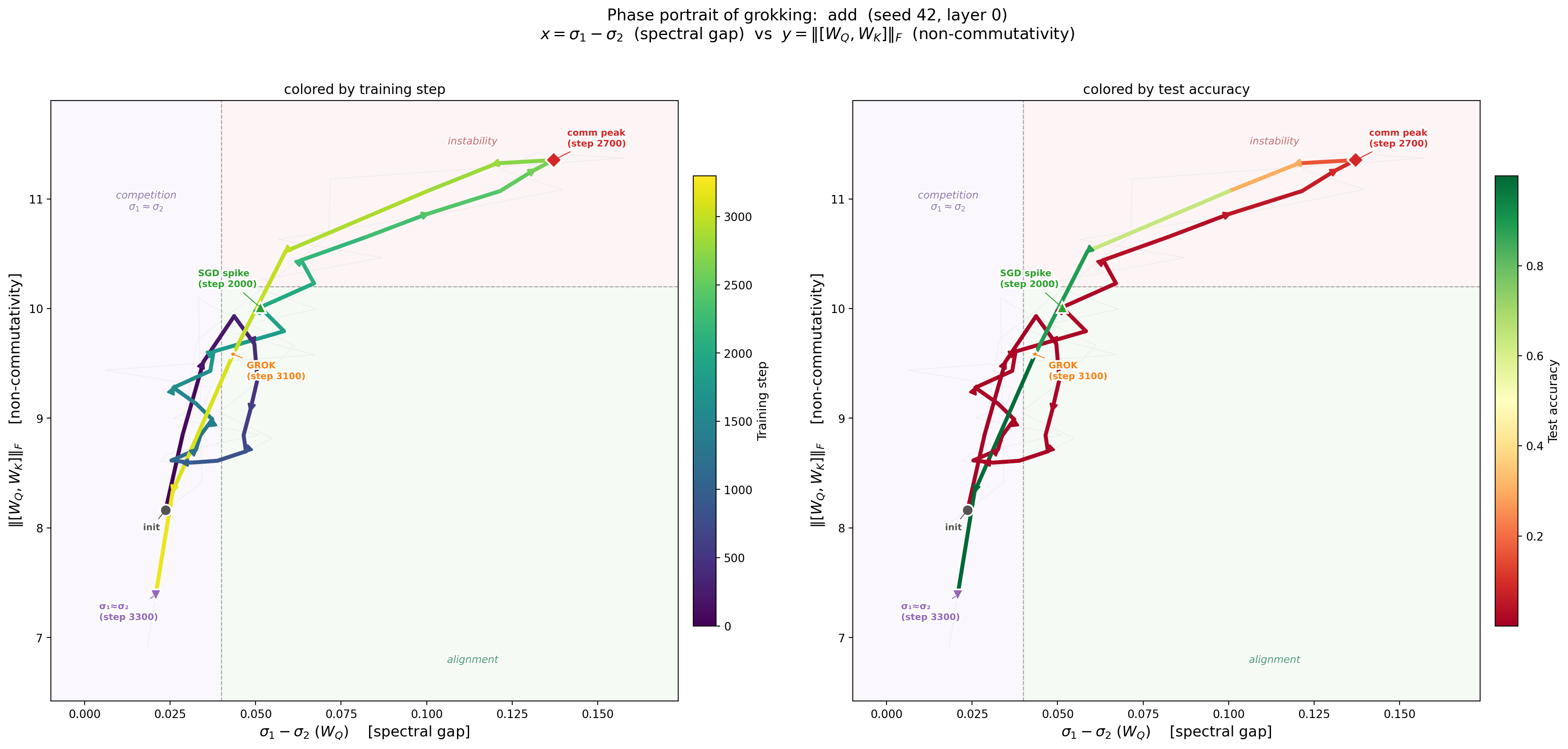}
    \caption{Phase portrait of grokking in the $(\sigma_1 - \sigma_2,\; \norm{[W_Q, W_K]}_F)$ plane for modular addition (seed~42, layer~0).
    Training follows a consistent trajectory through three regimes: a \emph{competition} phase where the top singular modes are nearly degenerate, an \emph{instability} phase where the commutator peaks as one mode begins to dominate, and an \emph{alignment} phase where the operators converge to a shared eigenbasis and grokking occurs.
    Left: colored by training step.
    Right: colored by test accuracy---the trajectory remains at chance (red) throughout competition and instability, turning green precisely as it enters the alignment region.}
    \label{fig:phase_portrait}
\end{figure}

This spectral mechanism is most clearly visible as a trajectory in the $(\sigma_1 - \sigma_2,\; \norm{[W_Q, W_K]}_F)$ phase plane (\Cref{fig:phase_portrait}).
Training follows a consistent path through three regimes:
\begin{itemize}
    \item \textbf{Competition} ($g_{12}$ small, commutator rising): the top singular modes are nearly degenerate and the representation basis is unstable.
    \item \textbf{Instability} ($g_{12}$ opening, commutator at peak): symmetry breaks as one mode begins to dominate, driving maximal non-commutativity.
    \item \textbf{Alignment} ($g_{12}$ large, commutator collapsing): $W_Q$ and $W_K$ converge to a shared basis and grokking occurs.
\end{itemize}

\begin{figure}[t]
    \centering
    \includegraphics[width=\textwidth]{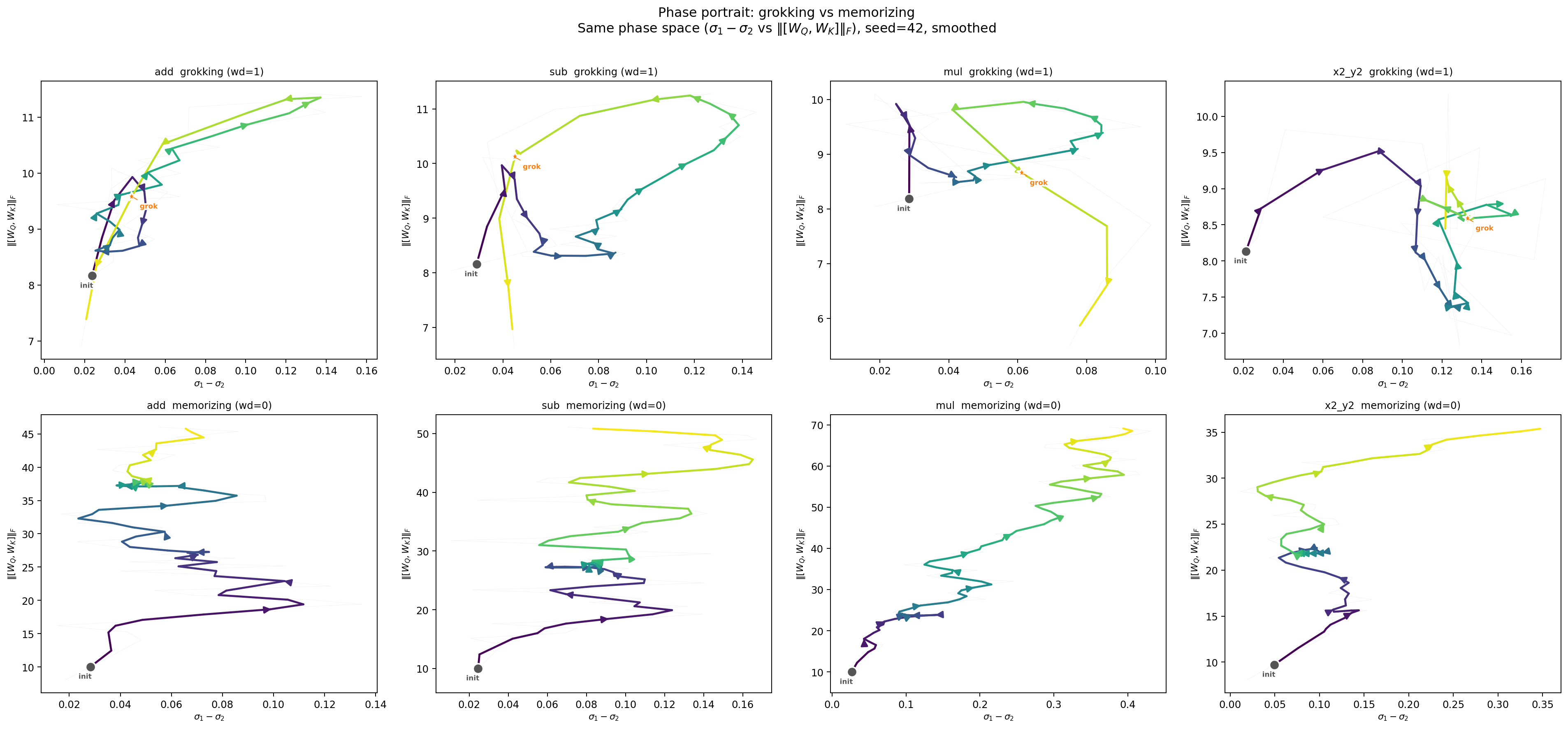}
    \caption{Phase portraits for grokking (top, $\mathrm{wd}=1$) vs.\ memorizing (bottom, $\mathrm{wd}=0$) across four operations (seed~42).
    Grokking trajectories exhibit a characteristic loop with directional flow through competition $\to$ instability $\to$ alignment.
    Memorizing trajectories are diffuse random walks in the same phase space, with much higher commutator values and no directed escape.}
    \label{fig:phase_portrait_control}
\end{figure}

\noindent
Non-grokking (memorizing) runs, by contrast, show no such loop structure: their trajectories wander diffusely in the same phase space without directional organization (\Cref{fig:phase_portrait_control}).

\begin{figure}[t]
    \centering
    \includegraphics[width=\textwidth]{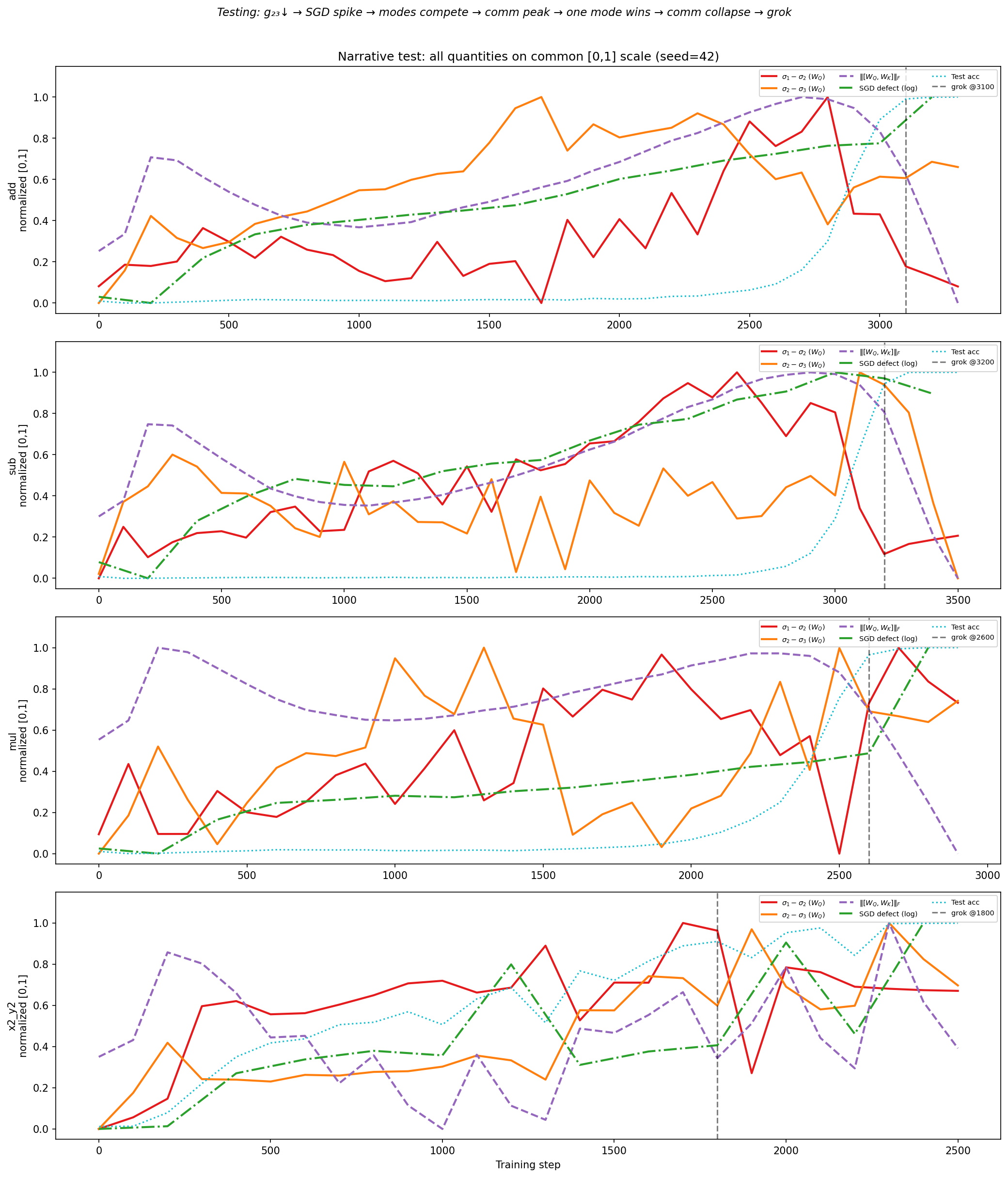}
    \caption{All quantities on a common $[0, 1]$ scale (seed~42, layer~0) for the four grokking operations.
    The spectral gap $\sigma_1 - \sigma_2$ (red) reaches its minimum \emph{before} the SGD defect spike (green, dashed-dotted), which in turn precedes the matrix commutator peak (purple, dashed).
    Test accuracy (cyan, dotted) rises last.
    Vertical dashed line: grokking step.}
    \label{fig:narrative_test}
\end{figure}

\subsection{Interpretation}
\label{sec:spectral_interpretation}

The spectral timeline provides a mechanistic explanation for the commutator dynamics that were previously only described phenomenologically.
The near-degeneracy of $\sigma_1$ and $\sigma_2$ creates an \emph{orientation instability}: when two singular values are close, the corresponding singular vectors are free to rotate under small perturbations.
This rotational freedom manifests as large commutator defects, since different mini-batch orderings push the representation basis in different directions.
Once the symmetry breaks ($\sigma_1 \gg \sigma_2$), the representation locks into a preferred orientation, the gradient ordering ambiguity resolves, and both the matrix commutator and the SGD defect decline.

This picture connects grokking to a \emph{spectral symmetry-breaking transition} in the attention weight matrices: the system passes through a transient near-degeneracy that destabilizes the representation, and grokking corresponds to the resolution of this instability as one mode dominates.

\paragraph{Relationship to trajectory PCA.}
The expanding-window PCA of update trajectories (\Cref{sec:pca}) captures optimizer dynamics---the covariance structure of successive weight changes---but does not directly reflect the spectral structure of the weight matrices themselves.
The weight-SVD analysis in this section provides a complementary and more direct view of the representation transition: it measures the eigenstructure of the actual learned operators rather than the statistics of how they are updated.

\paragraph{Connection to the intra-signal gap framework.}
The spectral timeline above can be situated within the broader \emph{intra-signal gap framework}~\citep{xu2026spectral_edge}, but a precise comparison requires distinguishing three spectral objects that appear in the literature:

\begin{itemize}
  \item[(i)] \textbf{Rolling-window Gram matrix} (Xu's object): the $W\times W$ matrix $\boldsymbol{G}(t) = \boldsymbol{X}(t)\boldsymbol{X}(t)^\top$ formed from $W$ consecutive parameter updates $\boldsymbol{\delta}_s = \boldsymbol{\theta}_{s+1} - \boldsymbol{\theta}_s$, with signal rank $k^* = \arg\max_j\, \omega_j \cdot (\sigma_j/\sigma_{j+1})$ where $\omega_j = \sigma_j / \sum_i \sigma_i$ weights the ratio by spectral mass at position $j$.
  The weighting suppresses spurious tail ratios that dominate the unweighted argmax outside the extreme aspect ratio regime ($P \gg W$).
  The gap ratio at $k^*$ is $R = \sigma_{k^*}/\sigma_{k^*+1}$.
  \item[(ii)] \textbf{Expanding-window displacement PCA} (our \Cref{sec:pca}): PCA on $\{W_t - W_0\}$ from initialization, yielding the rank-1 manifold finding (PC1 = 70--94\%).
  \item[(iii)] \textbf{Weight-matrix SVD} (this section): direct singular values $\sigma_j(W_Q)$ of the learned attention operators, with eigenvalue gap $g_{23} = \sigma_2^2 - \sigma_3^2$.
\end{itemize}

\noindent These are fundamentally different matrices with different spectra.
We verified this directly across 12 grokking runs (4~operations $\times$ 3~seeds, $\omega = 1.0$) and 12 matched control runs ($\omega = 0$, same operations): the rolling-window Gram matrix ($W = 10$) yields a gap ratio $R = 1.40 \pm 0.07$ during the pre-grokking phase for $\omega = 1.0$, while the $\omega = 0$ controls show \emph{higher} $R = 2.83 \pm 0.35$ (non-overlapping).
This inversion---memorization-only runs having more concentrated update spectra---reflects the structure of pure memorization gradients, which are highly rank-1 aligned, while grokking runs spread updates across spectral modes during circuit formation.
The signal rank stabilizes at $k^*_{\mathrm{terminal}} = 1$ in 9 of 12 grokking runs (cf.\ 10/12 in \citet{xu2026spectral_edge}), confirming dominant single-mode dynamics at convergence.

The Gram-matrix eigenvalue gap $g_{23} = \sigma_2^2 - \sigma_3^2$ is \emph{predictive}: it peaks during memorization and declines in 12 of 12 grokking runs (mean $40\times$, range $15$--$111\times$) \emph{before} grokking, while only 1 of 12 matched control runs ($\omega = 0$) shows any decline (\Cref{tab:gram_spectral}).
The decline precisely tracks the orientation instability that drives the commutator transition.

\begin{table}[t]
\centering
\caption{Rolling-window Gram matrix spectral analysis ($W = 10$, mean $\pm$ std over 3 seeds).
$g_{23}^{\mathrm{early}}$: peak $g_{23}$ before grokking;
$g_{23}^{\mathrm{grok}}$: $g_{23}$ at the grokking step;
Decline $= g_{23}^{\mathrm{early}} / g_{23}^{\mathrm{grok}}$;
$R$: gap ratio at signal rank $k^*$.}
\label{tab:gram_spectral}
\small
\begin{tabular}{llrrrrrrc}
\toprule
\textbf{Operation} & $\boldsymbol{\omega}$ & \textbf{Grok step} & $g_{23}^{\mathrm{early}}$ & $g_{23}^{\mathrm{grok}}$ & \textbf{Decline} & $R_{\mathrm{early}}$ & $k^*_{\mathrm{term}}$ & \textbf{Decl.} \\
\midrule
$a{+}b$      & 1.0 & 2500--3100  & $14.6 \pm 0.2$ & $0.56 \pm 0.4$ & $50\times$  & $1.40 \pm 0.05$ & 1 & \ding{51} \\
             & 0.0 & ---         & $33.0 \pm 9.3$ & ---            & ---         & $2.72 \pm 0.22$ & --- & --- \\
$a{\times}b$ & 1.0 & 2600--3000  & $14.7 \pm 0.1$ & $0.39 \pm 0.2$ & $49\times$  & $1.41 \pm 0.05$ & 1 & \ding{51} \\
             & 0.0 & ---         & $23.6 \pm 10.7$ & ---           & ---         & $2.85 \pm 0.11$ & --- & --- \\
$a{-}b$      & 1.0 & 3300--3700  & $14.4 \pm 0.1$ & $0.48 \pm 0.2$ & $35\times$  & $1.31 \pm 0.02$ & 1 & \ding{51} \\
             & 0.0 & ---         & $33.2 \pm 2.8$ & ---            & ---         & $2.62 \pm 0.23$ & --- & --- \\
$a^2{+}b^2$  & 1.0 & 2000--2500  & $18.9 \pm 0.5$ & $0.94 \pm 0.5$ & $28\times$  & $1.47 \pm 0.08$ & 1 & \ding{51} \\
             & 0.0 & ---         & $28.3 \pm 13.1$ & ---           & ---         & $3.12 \pm 0.64$ & --- & --- \\
\midrule
\multicolumn{2}{l}{\textit{Aggregate}} \\
All $\omega = 1$ & & & & & $40\times$ mean & $1.40 \pm 0.07$ & 1 (9/12) & 12/12 \\
All $\omega = 0$ & & & & & ---            & $2.83 \pm 0.35$ & --- & 1/12 \\
\bottomrule
\end{tabular}
\end{table}

The connection between these objects operates through the Davis--Kahan $\sin\Theta$ theorem~\citep{xu2026spectral_edge}, which is \emph{universal}: for \emph{any} symmetric matrix, the stability of a subspace under perturbation is controlled by $\|\sin\Theta(\mathcal{V},\hat{\mathcal{V}})\|_F \leq \|\boldsymbol{E}\|_F / \delta$, where $\delta$ is the eigenvalue gap.
This universality means that the qualitative prediction---gap closing destabilizes the associated subspace, gap opening stabilizes a new direction---applies regardless of whether we analyze the trajectory Gram matrix or the weight matrices directly.
What \emph{differs} between the objects are the quantitative dynamics: Xu's gap flow ODE, stability coefficient $\alpha_j$, and loss decomposition are derived specifically for the trajectory Gram matrix and do not transfer directly to weight-matrix eigenvalues.

In our setting, the weight-matrix SVD is the correct diagnostic for three reasons:
(a)~the eigenvalue gap $g_{23}$ directly controls the conditioning of the attention operator and hence the model's representational capacity;
(b)~$g_{23}$ decline \emph{precedes} grokking (lead time ${\sim}1000$ steps);
(c)~$g_{23}$ decline is reproduced in 12 of 12 grokking runs (mean $40\times$) and in only 1 of 12 matched controls.

The decline of $g_{23}$ is a gap closing event at position $k^* = 2$--$3$: as $\sigma_2$ approaches $\sigma_3$, the sub-leading subspace loses its stability, creating the orientation instability that drives the commutator transition.
The subsequent emergence of $\sigma_1 \gg \sigma_2$ is a gap opening event at position $k^* = 1$, stabilizing the dominant direction and enabling generalization.
This $k^*$ shift---from position 2--3 to position 1---is confirmed empirically: $k^*_{\mathrm{terminal}} = 1$ in 9 of 12 grokking runs using the signal-mass-weighted $k^*$ (the unweighted argmax is unstable in our aspect ratio regime where $p/W \sim 3 \times 10^4$, compared to $P/W \sim 10^7$ in \citeauthor{xu2026spectral_edge}'s setting where BBP detection threshold is vacuous).
The framework further predicts that weight decay drives the gap dynamics via suppression of low-curvature modes~\citep[Theorem~12.22]{xu2026spectral_edge}, consistent with the finding that all 12 runs with weight decay grok while only 1 of 12 matched controls without weight decay shows any decline (a late anomalous grokking event at step 41{,}700).
The $\omega = 0$ control further displays the \emph{predicted} spectral signature of the grokking-absent regime~\citep[Remark~12.22]{xu2026spectral_edge}: $g_{23}$ remains high throughout all training steps (mean early $g_{23} = 14$--$42$, never declining below $13$), confirming that without the curvature floor $\omega$ the spectral symmetry-breaking transition never initiates.

\subsection{Local Integrability and Basis-Independent Controls}
\label{sec:local_integrability}

The spectral analysis above examines the global singular value structure of individual weight matrices.
We now ask whether the commutator's alignment with weight structure is a \emph{structural} property of the optimization dynamics or an artifact of a particular choice of basis.

\begin{figure}[t]
    \centering
    \includegraphics[width=\textwidth]{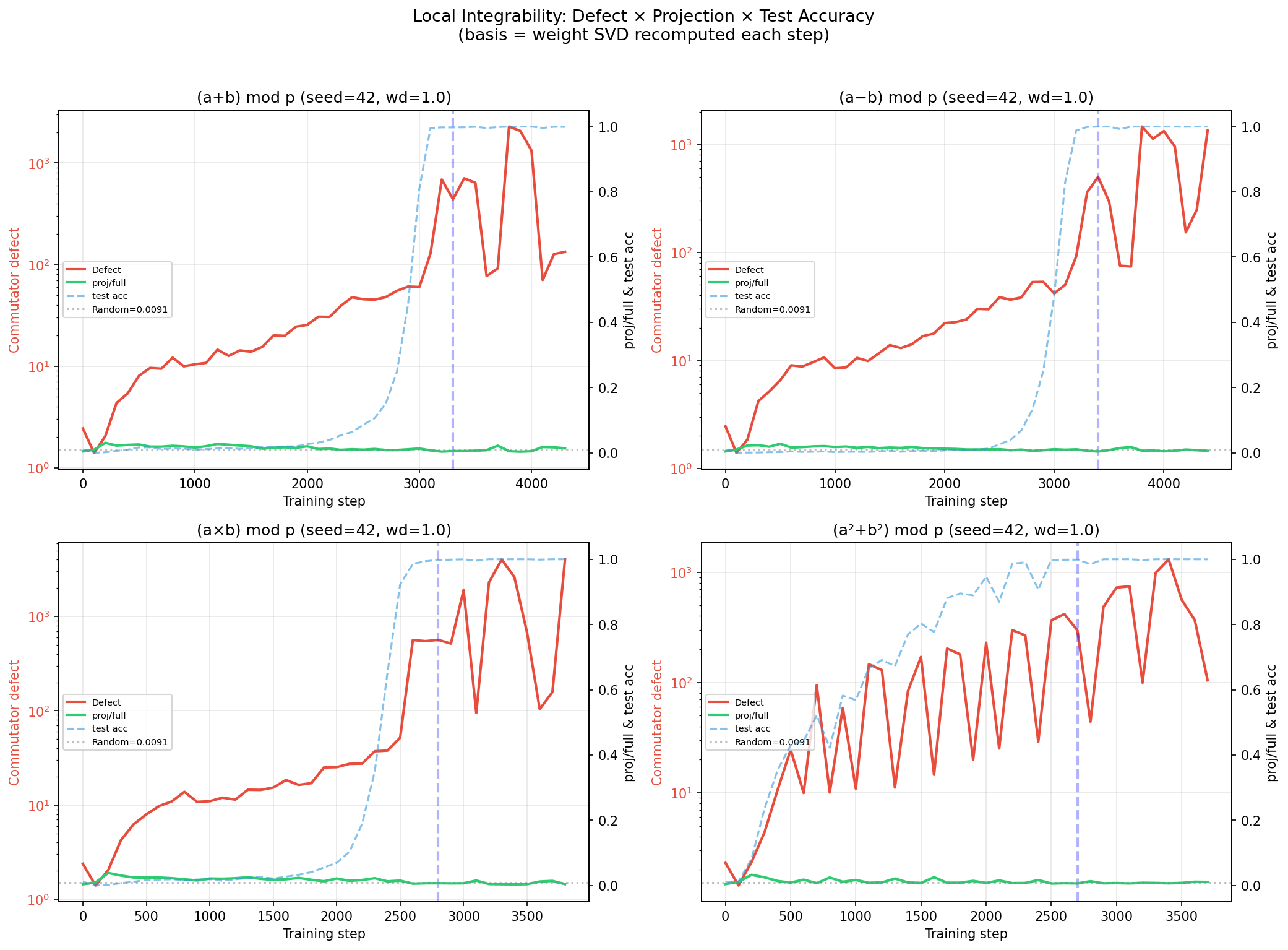}
    \caption{Local integrability: proj/full ratio $\rho_{\mathrm{local}}$ over training for grokking operations (solid) and non-grokking controls (dashed).
    Grokking operations show $\rho_{\mathrm{local}} \approx 1.0$ with a dip during memorization and recovery post-grokking; non-grokking controls remain below $0.5$.
    Random baseline (gray band) at $\sqrt{K/P} \approx 0.005$.}
    \label{fig:local_integrability}
\end{figure}

\paragraph{Local integrability.}
At each checkpoint, we construct a joint basis from the top-$k$ right singular vectors of all weight blocks ($W_Q$, $W_K$, $W_V$, $W_O$ per layer) and measure the fraction of commutator energy that projects onto this basis (proj/full ratio $\rho_{\mathrm{local}}$).
Across all 36 conditions (6 operations $\times$ 2 weight-decay settings $\times$ 3 seeds), the local integrability metric satisfies $\rho_{\mathrm{local}} \approx 1.0$: nearly all commutator energy lies within the weight-structure subspace.
The corresponding random-subspace baseline gives $\rho_{\mathrm{random}} \approx \sqrt{K/P} \approx 0.005$, yielding an execution/random ratio of $50$--$200\times$ throughout training (\Cref{fig:local_integrability}).
During the memorization phase, $\rho_{\mathrm{local}}$ dips to $0.3$--$0.5$ as commutator energy begins to leave the basis, then recovers to $>0.9$ post-grokking.
Non-grokking controls ($x^2+xy+y^2$, $x^3+xy$) show persistently lower $\rho_{\mathrm{local}} < 0.5$ with no phase-dependent recovery.

\begin{figure}[t]
    \centering
    \includegraphics[width=\textwidth]{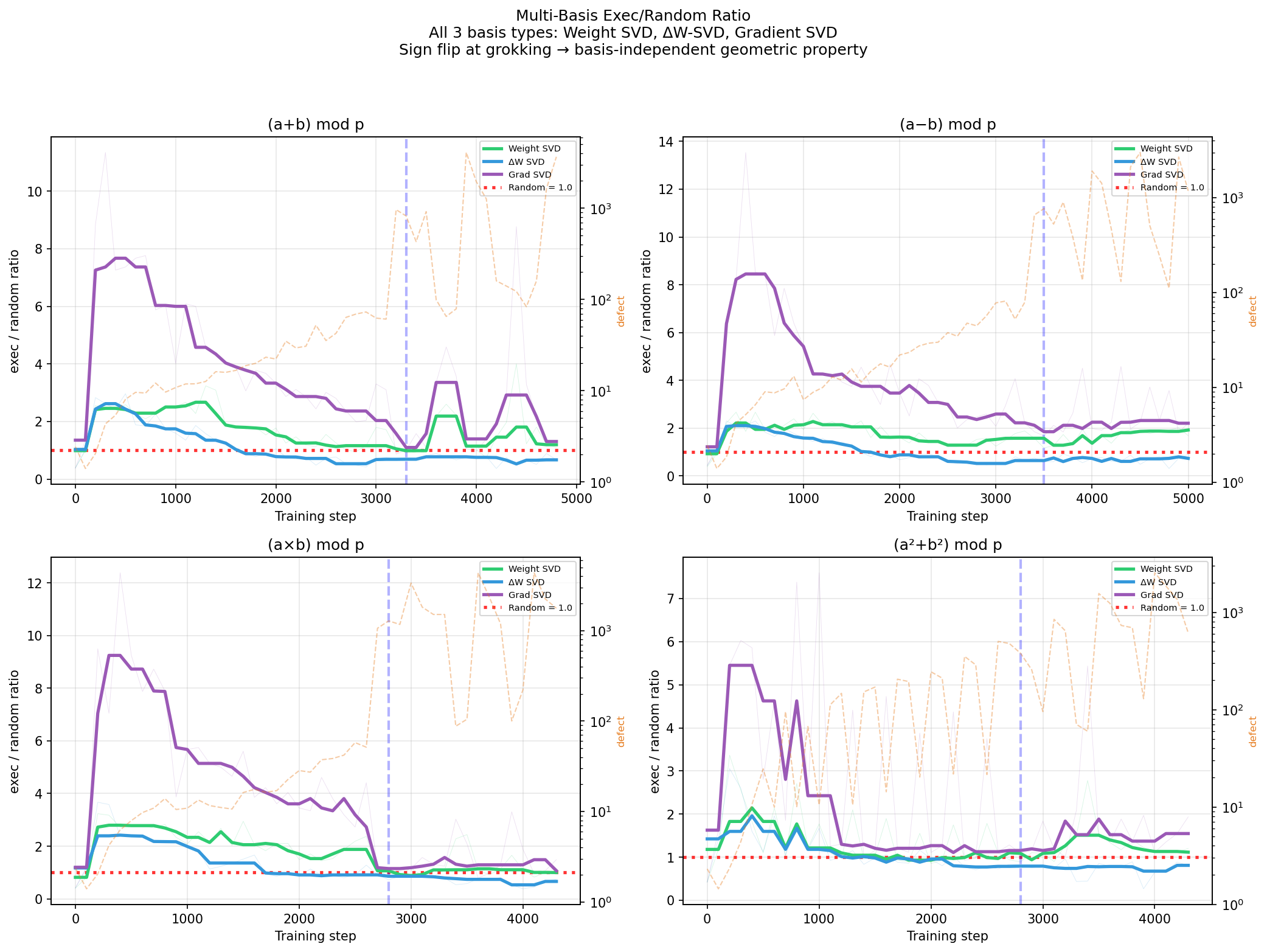}
    \caption{Basis-independent sign flip: exec/random ratio for three independent bases (weight SVD, displacement SVD, gradient SVD) across training.
    All three show ratio $>1$ during memorization (commutator aligns with weight structure) and $<1$ post-grokking (commutator exits the subspace), confirming the structural nature of the transition.}
    \label{fig:multibasis}
\end{figure}

\paragraph{Basis-independent sign flip.}
To verify that the commutator--subspace relationship is not dependent on the choice of basis, we repeat the projection analysis with three independent basis constructions:
(i)~weight SVD (top-$k$ singular vectors of $W$),
(ii)~displacement SVD (top-$k$ singular vectors of $W_t - W_0$), and
(iii)~gradient SVD (top-$k$ singular vectors of accumulated recent gradients).
All three bases exhibit a consistent \emph{sign flip}: the exec/random ratio exceeds~1 during memorization (commutator aligns with weight structure) and drops below~1 post-grokking (commutator exits the weight-structure subspace).
This pattern holds per-block across all attention weight matrices ($W_Q$, $W_K$, $W_V$, $W_O$), with MLP blocks showing a weaker but directionally consistent effect (\Cref{fig:multibasis}).
The basis independence of the sign flip confirms that the commutator's transition from within-subspace to outside-subspace alignment at grokking is a \emph{structural} geometric property of the optimization landscape, not an artifact of any particular decomposition.

\section{Discussion and Theoretical Connections}
\label{sec:discussion}

Our results support a unified geometric account of grokking: the training trajectory evolves predominantly within a low-dimensional execution subspace, transverse curvature grows in directions orthogonal to this subspace during the memorization plateau, and generalization is associated with escape from this metastable regime.
This picture---confinement, barrier growth, escape---organizes all six of our empirical contributions (rank-1 structure, empirical invariance, orthogonal curvature explosion, predictive lead time, causal asymmetry, and regime robustness) under a single dynamical narrative.
We now connect this narrative to several broader themes in learning theory and neural network optimization.

\subsection{Grokking as Metastable Escape in Curved Landscapes}
\label{sec:disc_metastable}

\emph{Thesis: grokking is best understood as escape from a metastable regime, not merely delayed learning.}

Across tasks and hyperparameter regimes, we observe that memorization confines training trajectories to regions of high curvature anisotropy, with commutator defects gradually increasing until a critical level triggers the generalization transition.
This is reminiscent of metastable escape in stochastic dynamical systems, where the defect magnitude functions as accumulated geometric tension and the learning rate controls damping.
Lower learning rates produce overdamped trajectories requiring substantial defect buildup ($\sim$30k steps at $\eta = 10^{-4}$), while higher learning rates facilitate rapid transitions ($\sim$1k steps at $\eta = 10^{-2}$; \Cref{sec:lr_sweep}).

\subsection{A Dynamical-Systems Interpretation}
\label{sec:disc_dynamical}

\emph{Thesis: grokking follows the classical dynamical-systems pattern of slow-manifold formation, transverse instability, and escape.}

Our findings admit a natural interpretation in the language of dynamical systems, proceeding through four phases:
\begin{enumerate}
    \item \textbf{Compression.} Early training collapses the weight trajectory onto a low-dimensional slow manifold (PC1 captures 68--83\% of variance; \Cref{sec:rank1}). The optimizer finds a rank-1 subspace and confines subsequent motion to it.
    \item \textbf{Transverse instability.} Curvature accumulates in the normal bundle of this manifold ($10$--$1000\times$ defect growth; \Cref{sec:curvature}), while the trajectory remains empirically invariant ($\rho \approx 1.000$; \Cref{sec:integrability}). The system becomes increasingly sensitive to off-manifold perturbations.
    \item \textbf{Critical transition.} Weight decay provides a sustained pressure toward lower-norm solutions. Combined with the accumulated transverse curvature, this drives the trajectory through a bifurcation into the generalizing basin.
    \item \textbf{Reorganization.} The trajectory settles into a new regime characterized by lower curvature anisotropy along the optimization path---the post-grokking solution.
\end{enumerate}

This parallels the classical scenario of slow-manifold collapse followed by transverse bifurcation in dissipative dynamical systems.
The spectral analysis of \Cref{sec:spectral} further sharpens this picture: the transverse instability phase corresponds to a near-degeneracy of the leading singular values of $W_Q$ and $W_K$, and the critical transition corresponds to a spectral symmetry-breaking event in which one singular mode dominates and the attention operators align into a shared eigenbasis.
The learning rate controls the damping: low $\eta$ produces overdamped dynamics with gradual defect buildup, high $\eta$ produces underdamped dynamics with rapid transitions (\Cref{sec:lr_sweep}).

\paragraph{Limitations of the dynamical-systems analogy.}
While this interpretation is supported by the temporal ordering (curvature precedes generalization) and the causal necessity of orthogonal gradient flow (\Cref{sec:interventions}), the available geometric signals---commutator defect, trajectory--curvature alignment, and invariance measure---do not combine into a \emph{predictive} composite diagnostic that discriminates grokking from non-grokking operations in advance.
The invariance measure is identically~1 across all conditions; trajectory--curvature alignment is noisy and overlapping; and defect magnitude only separates grokking from non-grokking operations retrospectively.
Moreover, the same grokking operations trained \emph{without} weight decay exhibit comparable defect acceleration without generalizing, confirming that curvature growth alone is insufficient---regularization pressure is the additional ingredient.
This raises the possibility that local curvature measurements alone may not suffice to predict grokking.
Local geometric diagnostics can detect that the loss landscape is reorganizing---transverse instability is growing, curvature is accumulating in the normal bundle---but whether this reorganization leads to generalization likely depends also on global properties of the landscape (such as the existence of a low-norm generalizing basin) and on exogenous factors like regularization strength.
Understanding how local curvature dynamics interact with global landscape structure to produce the grokking transition remains an important open question.

\subsection{Implicit Regularization and Low-Dimensional Structure}
\label{sec:disc_regularization}

\emph{Thesis: implicit regularization in grokking operates through the emergence of task-specific geometric structure, not merely norm or margin control.}

Following grokking, trajectories collapse onto low-dimensional execution manifolds (PC1 explains 68--83\% of variance; \Cref{sec:rank1}), accompanied by reduced curvature anisotropy along the trajectory.
The intervention experiments (\Cref{sec:interventions}) demonstrate that motion along these specific learned directions is necessary for grokking, while generic low-dimensional constraints are insufficient.
This suggests that implicit regularization in this setting operates through the progressive emergence of geometrically privileged subspaces.

\subsection{Scaling Behavior and Phase Diagrams}
\label{sec:disc_scaling}

\emph{Thesis: grokking exhibits a phase diagram with power-law scaling, paralleling phenomena in larger-scale systems.}

Our learning-rate sweeps across six half-decade-spaced rates reveal distinct overdamped, critically damped, and underdamped regimes (\Cref{sec:lr_sweep}), with grok time scaling approximately as $t_\text{grok} \propto \eta^{-1}$ and the curvature defect serving as an order parameter (sign test $p = 2^{-12}$; \Cref{sec:prediction}).
A power-law fit across 43 grokking runs with detectable onset yields $\alpha = 1.27 \pm 0.03$ ($R^2 = 0.97$; \Cref{fig:lr_scaling}), confirming that the defect onset lead time grows \emph{super-linearly} with grokking timescale.
The predictive window improves monotonically at slower, more realistic learning rates---from 24\% at $\eta = 3\!\times\!10^{-3}$ to 95\% at $\eta = 3\!\times\!10^{-5}$.
While our experiments operate in a small-model regime ($\sim$290k parameters), these scaling relationships parallel phenomena reported in large-scale language models, suggesting that grokking may represent a microscopic instance of optimization-driven phase transitions.

\subsection{Robustness, Flat Minima, and Quantization}
\label{sec:disc_robustness}

Recent empirical work has demonstrated that trained neural networks can tolerate substantial parameter quantization and compression with limited performance degradation.
Our results provide a geometric perspective on this robustness in the grokking regime.

Post-grokking solutions are characterized by reduced curvature anisotropy along the optimization trajectory: the commutator defect, while still nonzero, is concentrated orthogonally to the directions the optimizer traverses, and the invariance measure $\rho \approx 1.000$ (within numerical precision) indicates that curvature is confined to the normal bundle of the execution manifold.
Such regions, where curvature is confined to directions not visited by the optimizer, naturally support greater robustness to perturbations along the learned subspace.
In contrast, pre-grokking solutions reside in highly anisotropic, high-defect regions and are correspondingly more sensitive to perturbation.

This suggests that robustness to compression may be partly understood as a consequence of geometric reorganization during training, rather than solely as a byproduct of architectural or regularization choices.

\subsection{Connection to Mechanistic Interpretability}
\label{sec:disc_interpretability}

\emph{Thesis: the geometric transition during grokking corresponds to the stabilization of interpretable circuits.}

The formation of execution manifolds corresponds to the concentration of computation into low-dimensional subspaces, consistent with the emergence of interpretable circuits documented in prior work.
The collapse of curvature anisotropy during grokking indicates that these circuits become geometrically stabilized.
From this viewpoint, grokking marks the transition from distributed representations to structured, circuit-like organization---providing a potential bridge between optimization geometry and mechanistic interpretability.

\subsection{Limitations and Open Problems}
\label{sec:disc_limitations}

Our experiments are limited to relatively small Transformer models (2--3 layers, $\sim$290k parameters) and synthetic algorithmic tasks (modular arithmetic mod~97).
While these settings permit fine-grained geometric analysis, it remains unclear to what extent the observed phenomena---rank-1 manifolds, invariant execution submanifolds, predictive defect onset---generalize to large-scale language models and real-world datasets.
Preliminary experiments on Dyck languages and the SCAN compositional-generalization benchmark suggest that qualitatively similar low-dimensional confinement and transverse curvature dynamics arise beyond modular arithmetic; a systematic investigation of these settings is ongoing.

In addition, several of our diagnostic measures, including commutator defects (4 forward-backward passes per sample) and trajectory--curvature alignment, are computationally expensive and difficult to scale.
Developing efficient approximations and proxies for these geometric diagnostics remains an important direction for future work.

Finally, a complete theoretical characterization of the observed phase transitions remains open.
Deriving analytical models that capture defect accumulation, manifold formation, and damping-controlled dynamics represents a promising avenue for future research.

\subsection{Outlook}
\label{sec:disc_outlook}

Taken together, our results suggest that grokking reflects a geometric reorganization of the optimization landscape, governed by curvature, damping, and emergent low-dimensional structure.
By integrating dynamical, geometric, and causal analyses, this work provides a foundation for understanding delayed generalization as a phase transition in learning dynamics.

The commutator defect serves as a diagnostic for ongoing loss-landscape reorganization: monitoring it during training can detect geometric changes hundreds to tens of thousands of steps before they manifest in accuracy metrics.
However, defect growth alone does not discriminate which runs will ultimately generalize---non-grokking operations and unregularized training both exhibit curvature growth without generalization.
The diagnostic value lies in detecting \emph{that} the landscape is reorganizing, not in predicting \emph{whether} grokking will occur.
Developing composite diagnostics that combine curvature dynamics with regularization-sensitive signals remains an open direction.

We hope that this perspective will inform future studies of optimization, scaling, robustness, and interpretability in neural networks.

\section{Related Work}
\label{sec:related}

\paragraph{Grokking.}
\citet{power2022grokking} first observed delayed generalization in modular arithmetic.
\citet{nanda2023grokking} identified ``grokking circuits'' (Fourier-basis representations) in 1-layer models.
\citet{liu2022omnigrok} showed that grokking occurs broadly when weight decay or weight norm is controlled.
\citet{zhong2024clock} described clock and pizza representations in modular addition.
\citet{thilak2022slingshot} connected grokking to slingshot dynamics in adaptive optimizers.
\citet{lyu2024dichotomy} characterized the role of weight decay in separating memorization from generalization phases.
More recently, \citet{merrill2023tale} framed grokking as competition between sparse and dense subnetworks, \citet{varma2023explaining} explained it through circuit efficiency, \citet{davies2023unifying} connected grokking to double descent, and \citet{kumar2024grokking} characterized the lazy-to-rich training transition.
Concurrently, \citet{montanari2026phase} established sharp phase transitions for feature learning in two-layer networks under proportional asymptotics, providing a theoretical framework for understanding when gradient descent discovers low-dimensional structure.
Our work complements these representational and optimization-theoretic perspectives with a geometric and causal analysis.

\paragraph{Loss landscape geometry and scaling.}
The study of loss landscape geometry in neural networks has a rich history \citep{li2018visualizing, draxler2018essentially, garipov2018loss}.
\citet{fort2019large} studied the curvature of the loss landscape during training.
Our commutator defect is related to the Lie bracket of gradient vector fields and measures non-commutativity of the optimization flow; similar ideas appear in the study of natural gradient methods \citep{amari1998natural} and the Fisher information geometry of neural networks.
The power-law scaling we observe in grok time vs.\ learning rate resonates with the broader scaling laws literature \citep{kaplan2020scaling}, though our analysis operates at a much smaller scale.

\paragraph{Intrinsic dimensionality.}
\citet{li2018measuring} showed that neural network optimization occurs in a low-dimensional subspace.
\citet{xu2026lowdim} demonstrated that attention weight trajectories during grokking in modular arithmetic lie on a low-dimensional execution manifold, with PC1 capturing the majority of variance.
A corrected version of that work includes random-subspace baseline controls showing that the execution basis captures $2$--$10\times$ more commutator energy than a random subspace of equal dimension.
The present work extends those findings by establishing that the execution manifold exhibits empirical invariance under the optimization dynamics (curvature confined to the normal bundle), adding analogous random baseline controls, demonstrating that curvature dynamics predict the generalization transition, and testing the causal role of orthogonal gradient flow through intervention experiments.

\section{Conclusion}
\label{sec:conclusion}

We have shown that the weight-space trajectory during grokking lies on a rank-1 empirically invariant submanifold of parameter space---the execution manifold---and that loss-landscape curvature is confined to the normal bundle of this submanifold.
Curvature growth in the normal bundle consistently precedes generalization by 600--1600 steps, establishing a robust temporal ordering; however, non-grokking operations also exhibit moderate curvature growth without generalizing, so onset is a necessary precondition rather than a sufficient predictor.
These findings hold across six modular arithmetic operations, three random seeds, two weight-decay settings, a $100\times$ learning rate sweep, and two qualitatively different hyperparameter regimes ($200\times$ range in training timescale).
Causal intervention experiments close the loop: suppressing orthogonal gradient flow prevents grokking with a monotonic dose--response across four operations, while artificially boosting curvature defects has no effect, establishing that normal-bundle curvature growth is mechanistically necessary for the generalization transition.
Weight-SVD analysis reveals the mechanism underlying these dynamics: grokking is preceded by a transient near-degeneracy of the leading singular values of the attention matrices ($\sigma_1 \approx \sigma_2$), during which the representation basis is unstable and the commutator defect rises sharply.
Generalization coincides with the breaking of this spectral symmetry as one mode dominates and the operators align.
Local integrability analysis confirms that the commutator energy is almost entirely contained within the weight-structure subspace ($\rho_{\mathrm{local}} \approx 1.0$), and this alignment is basis-independent: three independent decompositions all exhibit a sign flip at grokking, confirming the structural nature of the transition.
The geometric picture---empirically invariant execution manifold, orthogonal curvature confinement, spectral symmetry-breaking, basis-independent integrability, and causal confirmation via interventions---provides a new lens for understanding the grokking phenomenon and suggests that monitoring gradient non-commutativity during training may serve as a diagnostic for ongoing loss-landscape reorganization, even when generalization is not yet observable in accuracy metrics.

\paragraph{Reproducibility.}
All code and figures are available at \url{https://github.com/skydancerosel/grokking-integrability}.
Total compute for full reproduction is approximately 9 hours on a single Apple M-series machine.


\bibliographystyle{plainnat}

\begin{thebibliography}{20}
\providecommand{\natexlab}[1]{#1}
\providecommand{\url}[1]{\texttt{#1}}
\expandafter\ifx\csname urlstyle\endcsname\relax
  \providecommand{\doi}[1]{doi: #1}\else
  \providecommand{\doi}{doi: \begingroup \urlstyle{rm}\Url}\fi

\bibitem[Amari(1998)]{amari1998natural}
Shun-ichi Amari.
\newblock Natural gradient works efficiently in learning.
\newblock \emph{Neural Computation}, 10\penalty0 (2):\penalty0 251--276, 1998.

\bibitem[Davies et~al.(2023)Davies, Langosco, and Krueger]{davies2023unifying}
Xander Davies, Lauro Langosco, and David Krueger.
\newblock Unifying grokking and double descent.
\newblock \emph{arXiv preprint arXiv:2303.06173}, 2023.

\bibitem[Draxler et~al.(2018)Draxler, Veschgini, Salmhofer, and
  Hamprecht]{draxler2018essentially}
Felix Draxler, Kambis Veschgini, Manfred Salmhofer, and Fred~A Hamprecht.
\newblock Essentially no barriers in neural network energy landscape.
\newblock In \emph{International Conference on Machine Learning}, pages
  1309--1318, 2018.

\bibitem[Fort and Jastrzebski(2019)]{fort2019large}
Stanislav Fort and Stanislaw Jastrzebski.
\newblock Large scale structure of neural network loss landscapes.
\newblock In \emph{Advances in Neural Information Processing Systems},
  volume~32, 2019.

\bibitem[Garipov et~al.(2018)Garipov, Izmailov, Podoprikhin, Vetrov, and
  Wilson]{garipov2018loss}
Timur Garipov, Pavel Izmailov, Dmitrii Podoprikhin, Dmitry Vetrov, and
  Andrew~Gordon Wilson.
\newblock Loss surfaces, mode connectivity, and fast ensembling of {DNN}s.
\newblock In \emph{Advances in Neural Information Processing Systems},
  volume~31, 2018.

\bibitem[Kaplan et~al.(2020)Kaplan, McCandlish, Henighan, Brown, Chess, Child,
  Gray, Radford, Wu, and Amodei]{kaplan2020scaling}
Jared Kaplan, Sam McCandlish, Tom Henighan, Tom~B Brown, Benjamin Chess, Rewon
  Child, Scott Gray, Alec Radford, Jeffrey Wu, and Dario Amodei.
\newblock Scaling laws for neural language models.
\newblock \emph{arXiv preprint arXiv:2001.08361}, 2020.

\bibitem[Kumar et~al.(2024)Kumar, Bordelon, Gershman, and
  Pehlevan]{kumar2024grokking}
Tanishq Kumar, Blake Bordelon, Samuel~J Gershman, and Cengiz Pehlevan.
\newblock Grokking as the transition from lazy to rich training dynamics.
\newblock \emph{arXiv preprint arXiv:2310.06110}, 2024.

\bibitem[Li et~al.(2018{\natexlab{a}})Li, Farkhoor, Liu, and
  Yosinski]{li2018measuring}
Chunyuan Li, Heerad Farkhoor, Rosanne Liu, and Jason Yosinski.
\newblock Measuring the intrinsic dimension of objective landscapes.
\newblock In \emph{International Conference on Learning Representations},
  2018{\natexlab{a}}.

\bibitem[Li et~al.(2018{\natexlab{b}})Li, Xu, Taylor, Studer, and
  Goldstein]{li2018visualizing}
Hao Li, Zheng Xu, Gavin Taylor, Christoph Studer, and Tom Goldstein.
\newblock Visualizing the loss landscape of neural nets.
\newblock In \emph{Advances in Neural Information Processing Systems},
  volume~31, 2018{\natexlab{b}}.

\bibitem[Liu et~al.(2022)Liu, Kitouni, Nolte, Michaud, Tegmark, and
  Williams]{liu2022omnigrok}
Ziming Liu, Ouail Kitouni, Niklas Nolte, Eric~J Michaud, Max Tegmark, and Mike
  Williams.
\newblock Omnigrok: Grokking beyond algorithmic data.
\newblock \emph{arXiv preprint arXiv:2210.01117}, 2022.

\bibitem[Lyu et~al.(2024)Lyu, Jin, Li, Du, Lee, and Hu]{lyu2024dichotomy}
Kaifeng Lyu, Jikai Jin, Zhiyuan Li, Simon~S Du, Jason~D Lee, and Wei Hu.
\newblock Dichotomy of early and late phase implicit biases can provably induce
  grokking.
\newblock \emph{arXiv preprint arXiv:2311.18817}, 2024.

\bibitem[Merrill et~al.(2023)Merrill, Tsilivis, and Shukla]{merrill2023tale}
William Merrill, Nikolaos Tsilivis, and Aman Shukla.
\newblock A tale of two circuits: Grokking as competition of sparse and dense
  subnetworks.
\newblock \emph{arXiv preprint arXiv:2303.11873}, 2023.

\bibitem[Montanari and Wang(2026)]{montanari2026phase}
Andrea Montanari and Zihao Wang.
\newblock Phase transitions for feature learning in neural networks.
\newblock \emph{arXiv preprint arXiv:2602.01434}, 2026.
\newblock URL \url{https://arxiv.org/abs/2602.01434}.

\bibitem[Nanda et~al.(2023)Nanda, Chan, Liberum, Smith, and
  Steinhardt]{nanda2023grokking}
Neel Nanda, Lawrence Chan, Tom Liberum, Jess Smith, and Jacob Steinhardt.
\newblock Progress measures for grokking via mechanistic interpretability.
\newblock \emph{arXiv preprint arXiv:2301.05217}, 2023.

\bibitem[Power et~al.(2022)Power, Burda, Edwards, Babuschkin, and
  Misra]{power2022grokking}
Alethea Power, Yuri Burda, Harri Edwards, Igor Babuschkin, and Vedant Misra.
\newblock Grokking: Generalization beyond overfitting on small algorithmic
  datasets.
\newblock In \emph{ICLR 2022 Workshop on MATH-AI}, 2022.
\newblock URL \url{https://arxiv.org/abs/2201.02177}.

\bibitem[Thilak et~al.(2022)Thilak, Littwin, Zhai, Saremi, Paiss, and
  Susskind]{thilak2022slingshot}
Vimal Thilak, Etai Littwin, Shuangfei Zhai, Omid Saremi, Roni Paiss, and Joshua
  Susskind.
\newblock The slingshot mechanism: An empirical study of adaptive optimizers
  and the grokking phenomenon.
\newblock \emph{arXiv preprint arXiv:2206.04817}, 2022.

\bibitem[Varma et~al.(2023)Varma, Shah, Kenton, Kram{\'a}r, and
  Nanda]{varma2023explaining}
Vikrant Varma, Rohin Shah, Zachary Kenton, J{\'a}nos Kram{\'a}r, and Neel
  Nanda.
\newblock Explaining grokking through circuit efficiency.
\newblock \emph{arXiv preprint arXiv:2309.02390}, 2023.

\bibitem[Xu(2026{\natexlab{a}})]{xu2026lowdim}
Yongzhong Xu.
\newblock Low-dimensional execution manifolds in transformer learning dynamics:
  Evidence from modular arithmetic tasks.
\newblock \emph{arXiv preprint arXiv:2602.10496}, 2026{\natexlab{a}}.
\newblock URL \url{https://arxiv.org/abs/2602.10496}.

\bibitem[Xu(2026{\natexlab{b}})]{xu2026spectral_edge}
Yongzhong Xu.
\newblock The spectral edge thesis: Intra-signal gap dynamics in transformer
  training.
\newblock \emph{arXiv preprint arXiv:2603.28964}, 2026{\natexlab{b}}.
\newblock URL \url{https://arxiv.org/abs/2603.28964}.

\bibitem[Zhong et~al.(2024)Zhong, Liu, Tegmark, and Andreas]{zhong2024clock}
Ziqian Zhong, Ziming Liu, Max Tegmark, and Jacob Andreas.
\newblock The clock and the pizza: Two stories in mechanistic explanation of
  neural networks.
\newblock \emph{Advances in Neural Information Processing Systems}, 36, 2024.

\end{thebibliography}

\appendix
\section{Additional Figures}
\label{app:figures}

\begin{figure}[ht]
    \centering
    \includegraphics[width=0.7\textwidth]{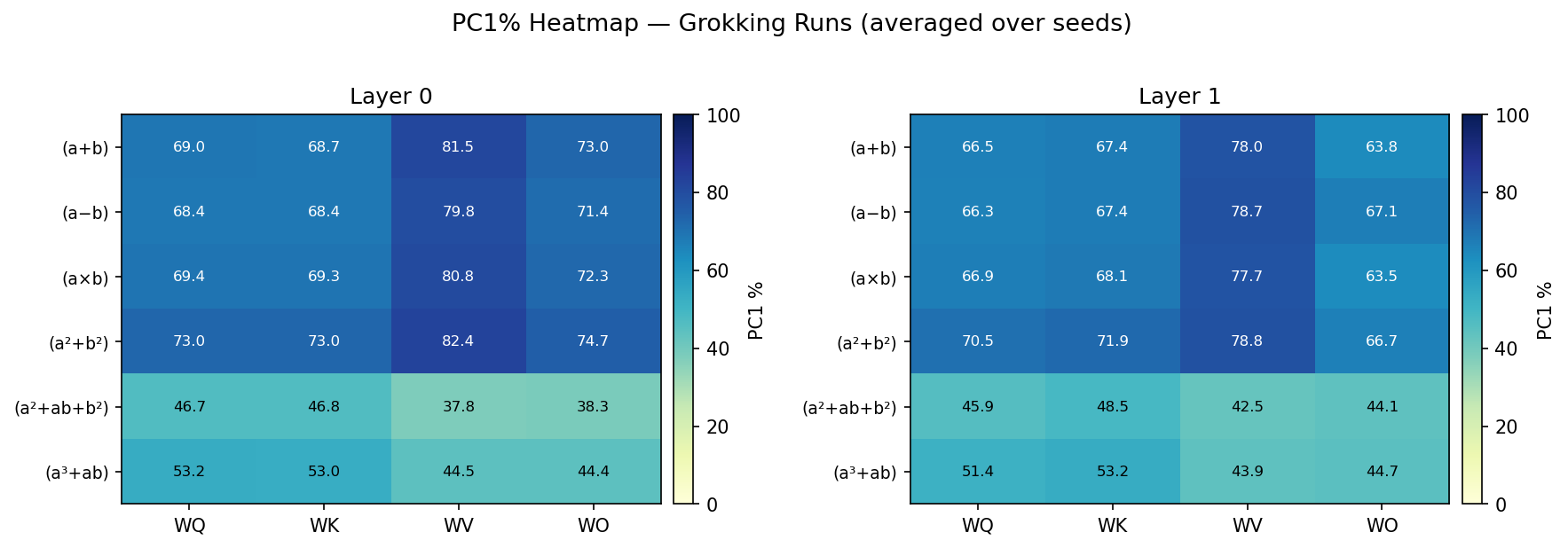}
    \caption{PC1\% heatmap by operation and weight matrix (last layer, wd=1.0). All weight matrices show high PC1\% for grokking operations.}
    \label{fig:heatmap}
\end{figure}

\begin{figure}[ht]
    \centering
    \includegraphics[width=0.7\textwidth]{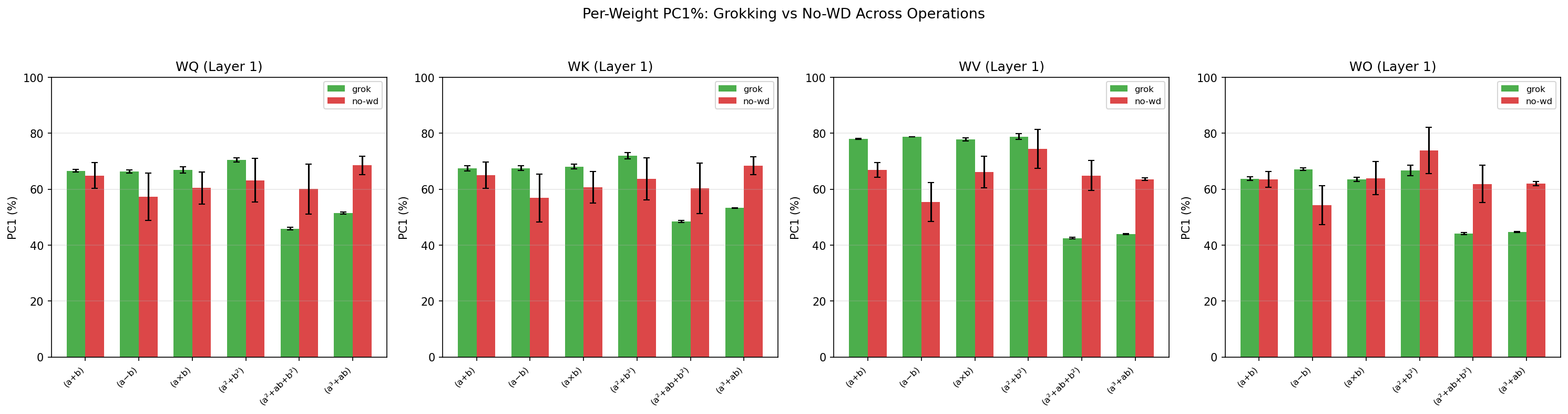}
    \caption{Per-weight-matrix PC1\% comparison across operations (grok vs.\ no-wd).}
    \label{fig:per_weight}
\end{figure}

\begin{figure}[ht]
    \centering
    \begin{subfigure}[t]{0.48\textwidth}
        \centering
        \includegraphics[width=\textwidth]{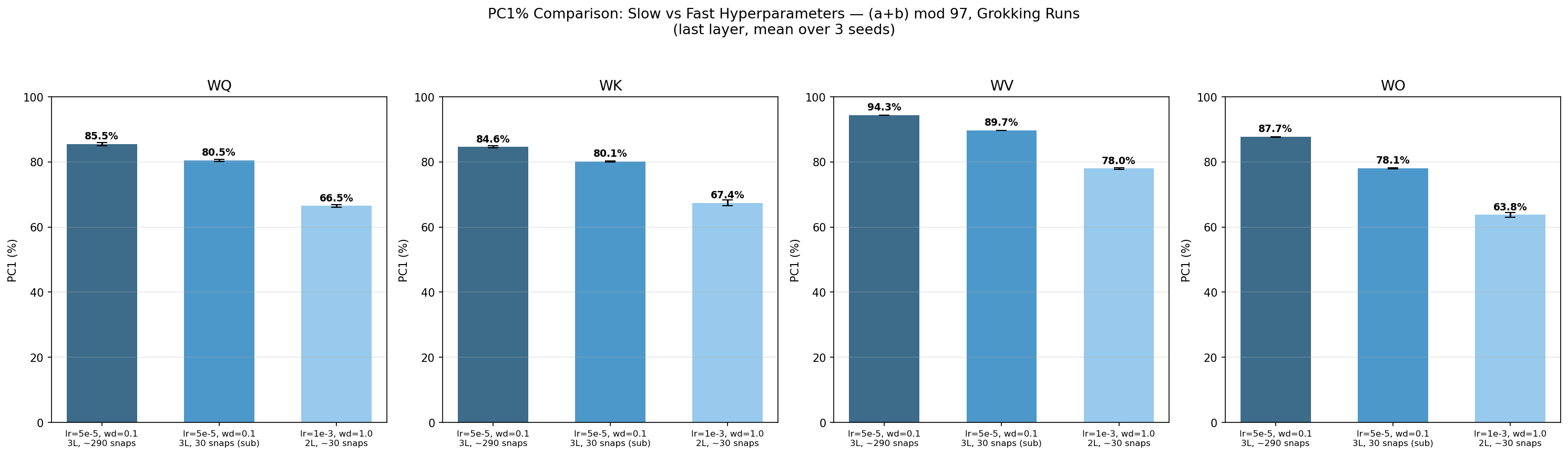}
        \caption{Slow vs.\ fast regime PC1\%. The slow regime shows lower PC1\%, but still well above the null model.}
        \label{fig:regime_pca}
    \end{subfigure}
    \hfill
    \begin{subfigure}[t]{0.48\textwidth}
        \centering
        \includegraphics[width=\textwidth]{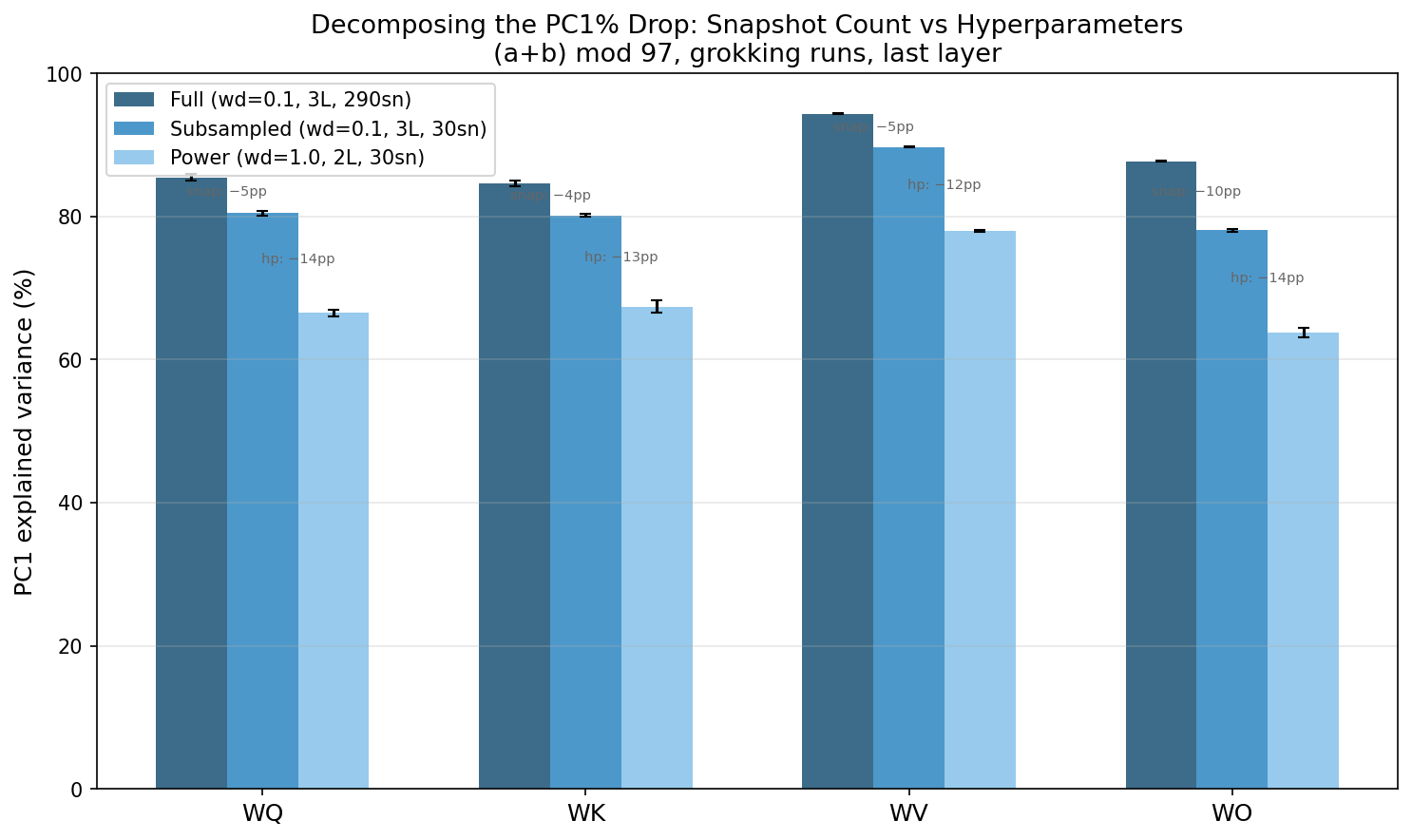}
        \caption{Decomposition of PC1\% drop between regimes: which hyperparameter drives the difference.}
        \label{fig:decomposition}
    \end{subfigure}
    \caption{Regime comparison for PCA concentration.}
    \label{fig:regime_pca_detail}
\end{figure}

\begin{figure}[ht]
    \centering
    \begin{subfigure}[t]{0.48\textwidth}
        \centering
        \includegraphics[width=\textwidth]{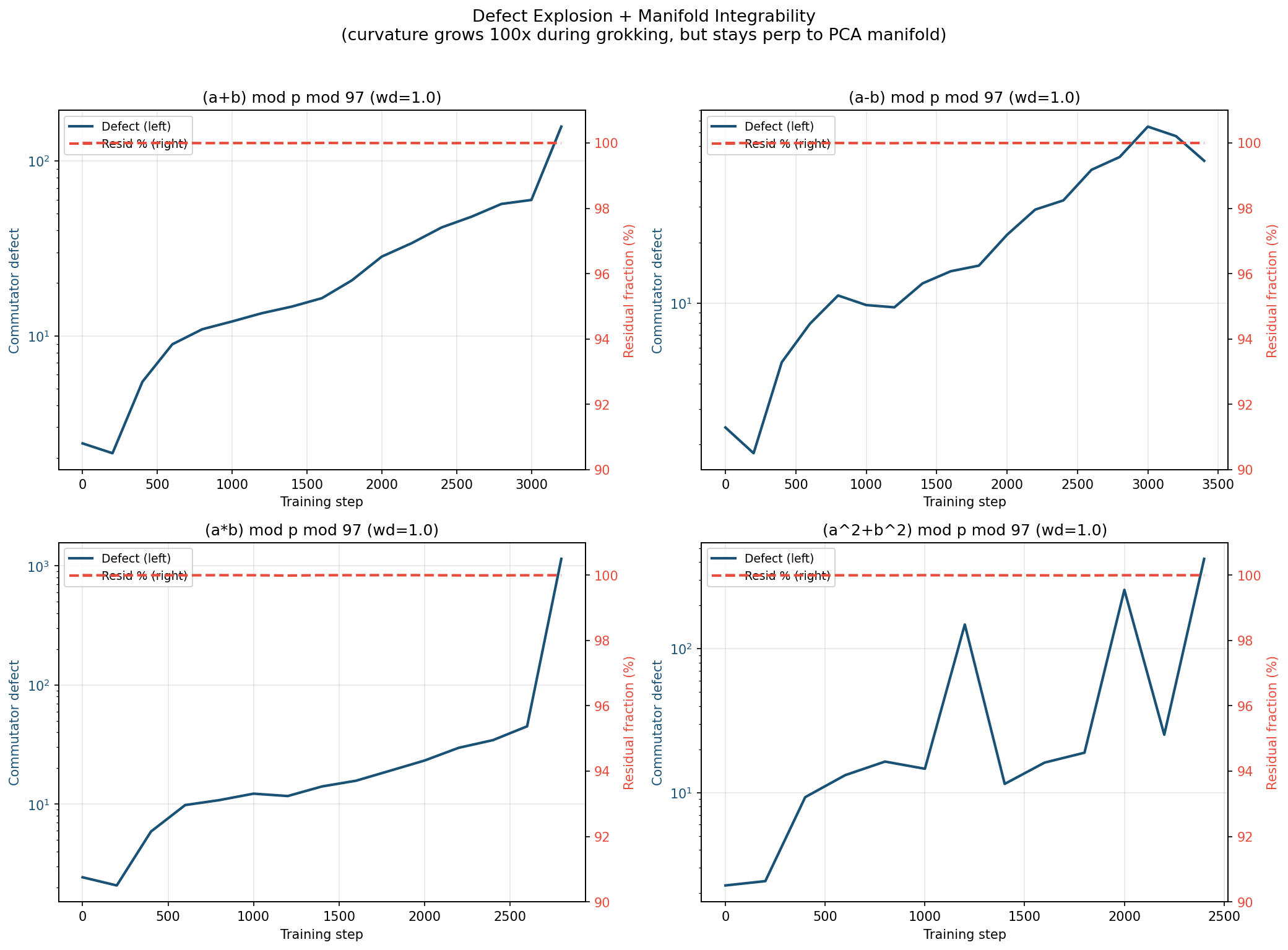}
        \caption{Combined view: defect magnitude and invariance measure over training.}
        \label{fig:combined}
    \end{subfigure}
    \hfill
    \begin{subfigure}[t]{0.48\textwidth}
        \centering
        \includegraphics[width=\textwidth]{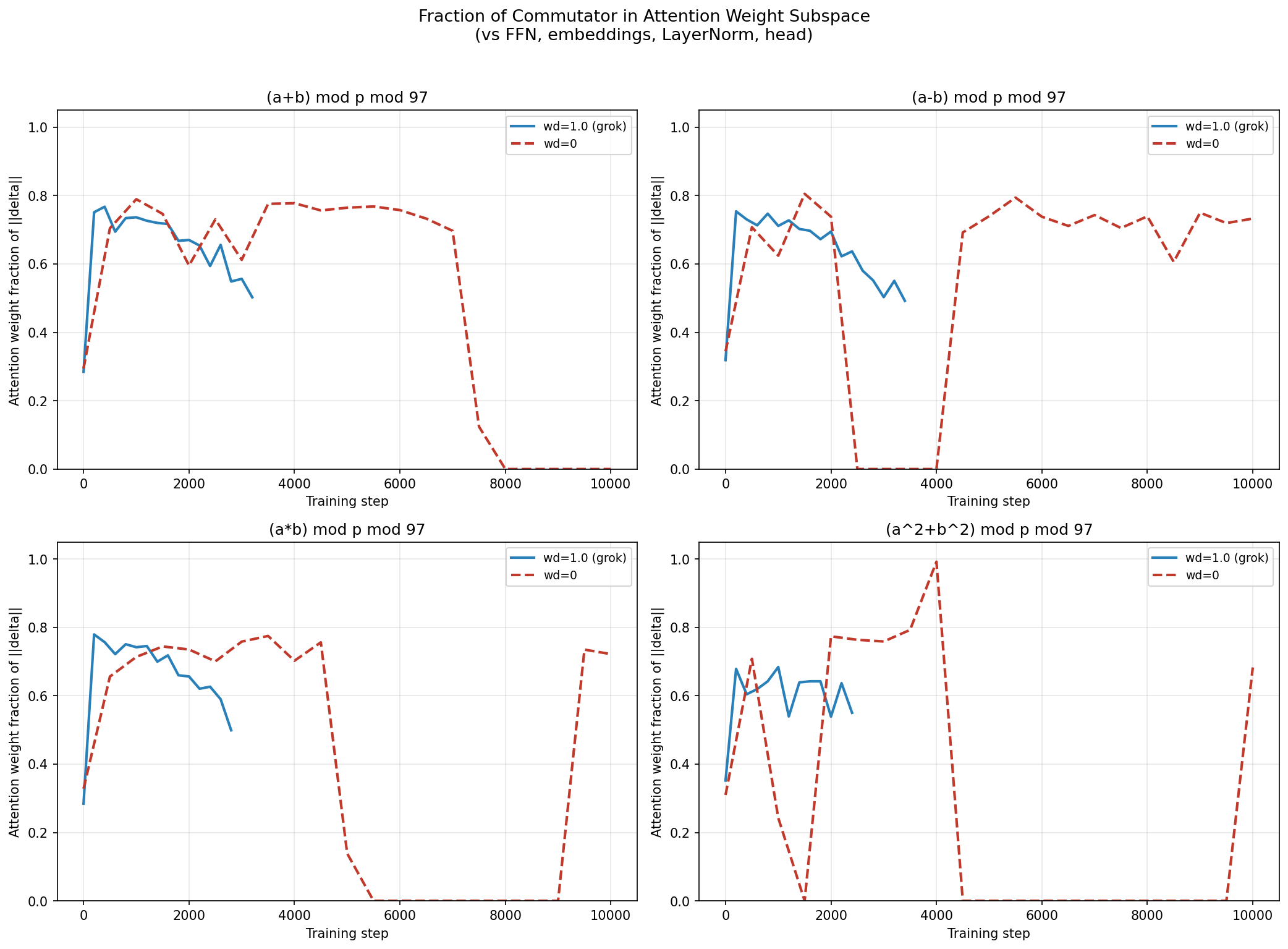}
        \caption{Attention weight fraction of commutator defect.}
        \label{fig:attn_fraction}
    \end{subfigure}
    \caption{Commutator analysis details.}
    \label{fig:comm_details}
\end{figure}

\begin{figure}[ht]
    \centering
    \begin{subfigure}[t]{0.48\textwidth}
        \centering
        \includegraphics[width=\textwidth]{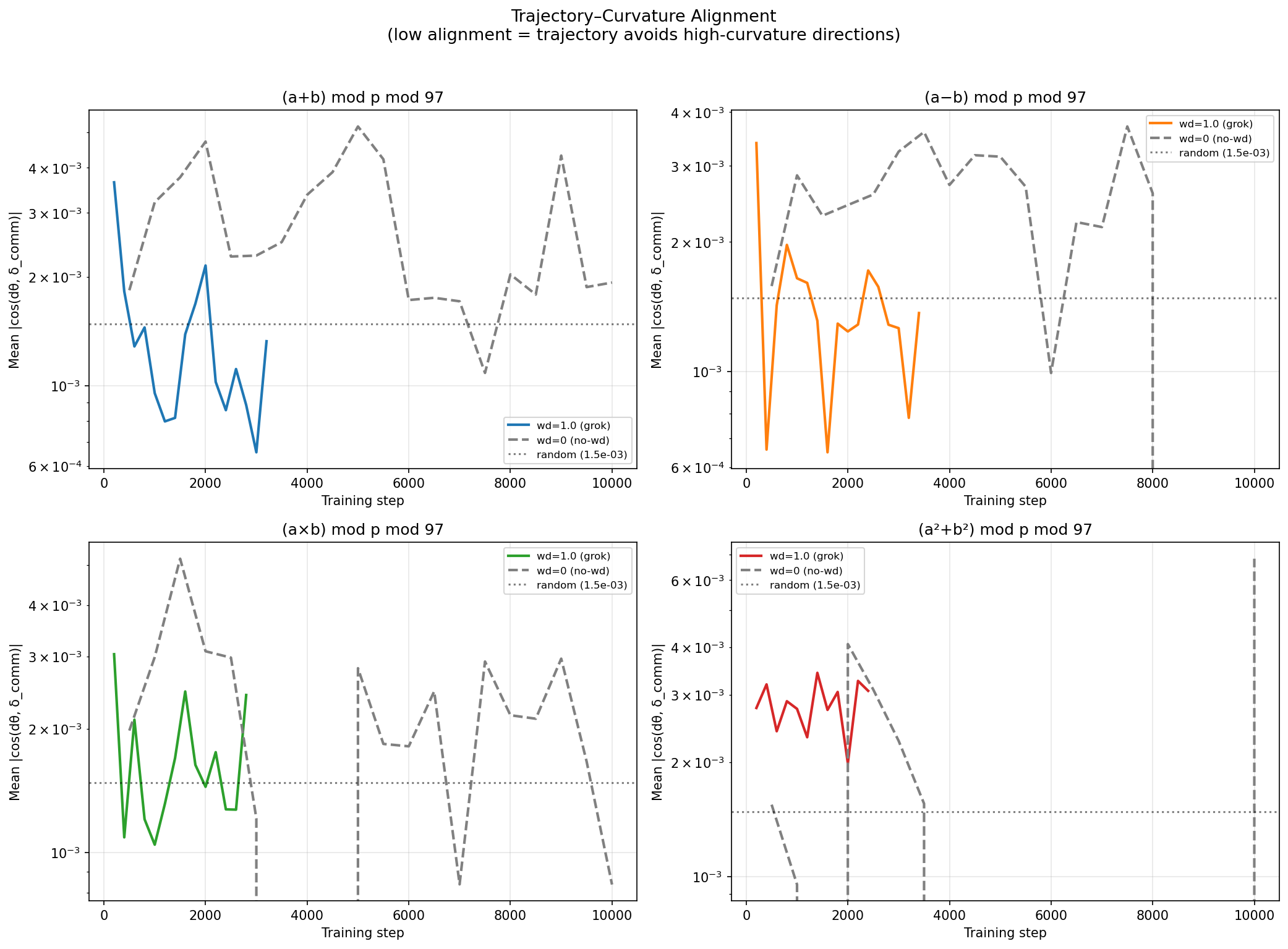}
        \caption{Trajectory-curvature alignment over training.}
        \label{fig:alignment}
    \end{subfigure}
    \hfill
    \begin{subfigure}[t]{0.48\textwidth}
        \centering
        \includegraphics[width=\textwidth]{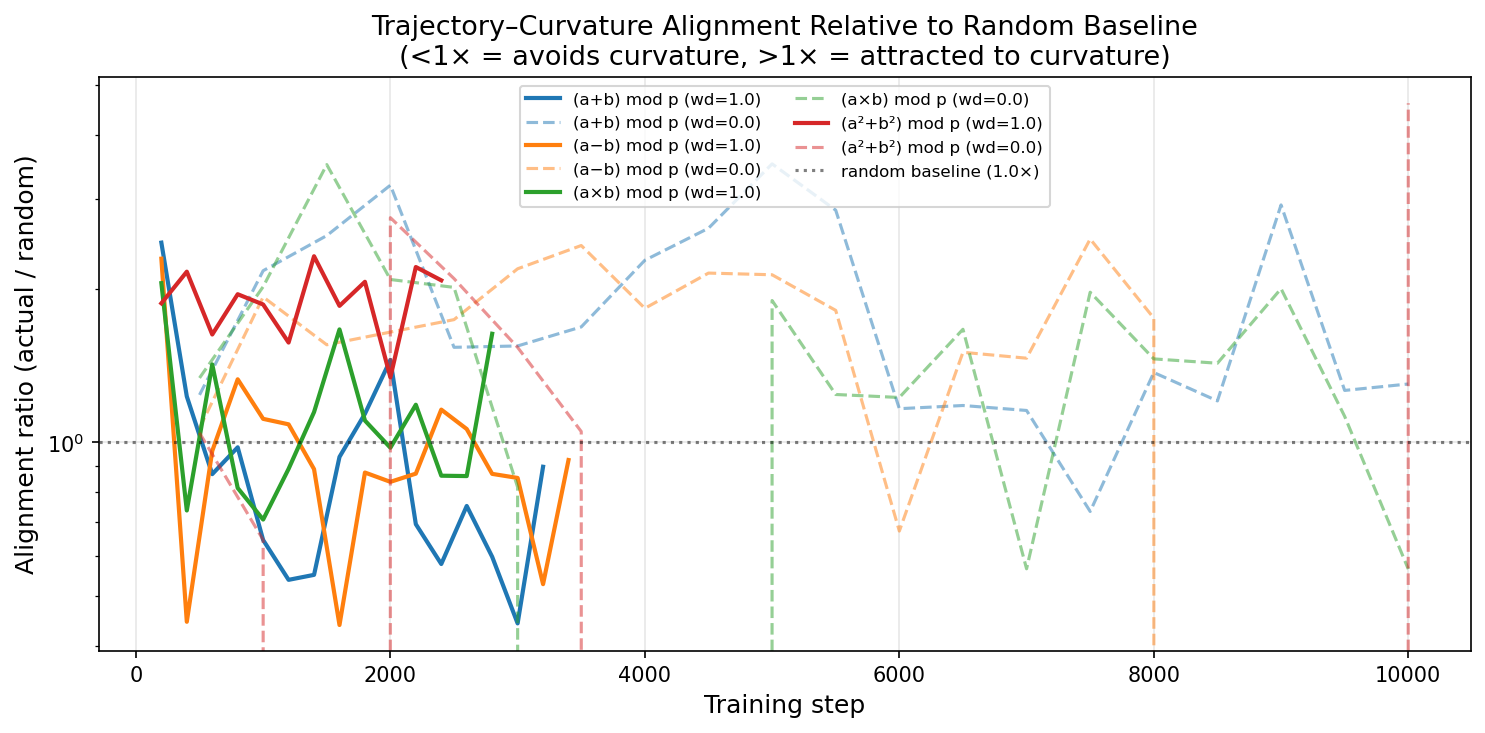}
        \caption{Alignment ratio vs.\ random baseline: the trajectory does not prefer curvature directions.}
        \label{fig:alignment_ratio}
    \end{subfigure}
    \caption{Converse analysis: the weight trajectory avoids high-curvature directions.}
    \label{fig:converse}
\end{figure}

\begin{figure}[ht]
    \centering
    \includegraphics[width=0.7\textwidth]{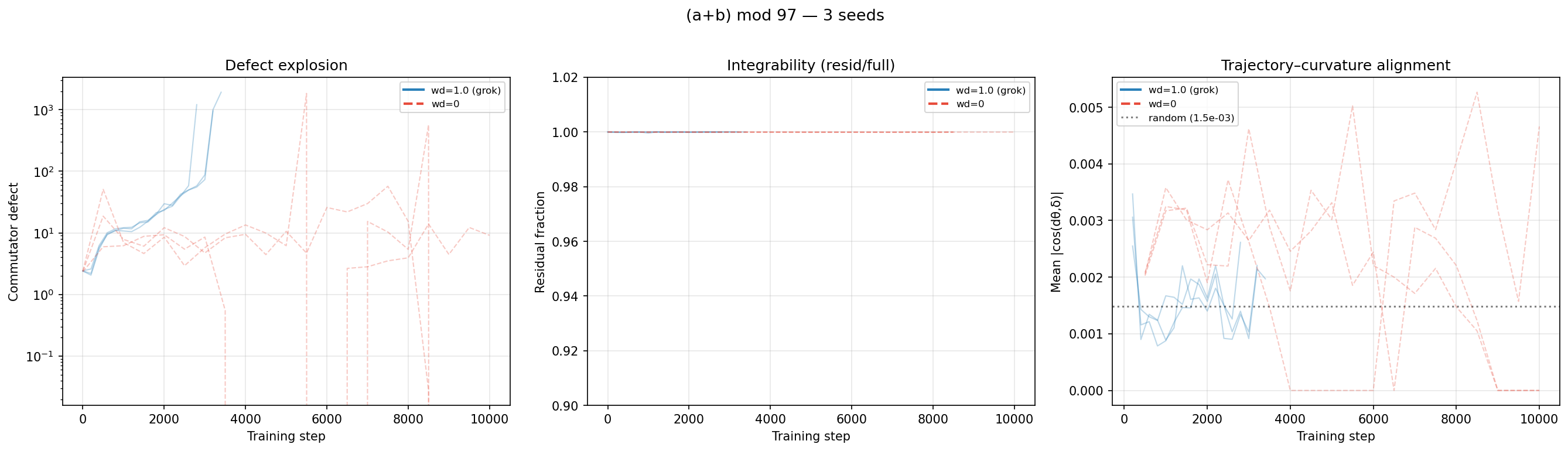}
    \caption{Temporal traces for $(a+b) \bmod 97$ with 3 seed overlays, showing consistency of the invariance and defect patterns across seeds.}
    \label{fig:temporal_seeds}
\end{figure}

\end{document}